\documentclass[10.5pt]{article}
\usepackage[fleqn]{amsmath} 
\usepackage{amsthm,amsmath,amssymb}  
\usepackage{mathrsfs} 
\usepackage{geometry}
\usepackage{cite}
\usepackage[colorlinks,linkcolor=blue,citecolor=blue]{hyperref} 
\usepackage{booktabs}
\usepackage{threeparttable}
\usepackage{multirow}
\usepackage[font=small,labelfont=bf,labelsep=none]{caption}
\usepackage{graphics}
\usepackage{epsfig}
\usepackage{graphicx} 
\usepackage{epstopdf}
\usepackage{mathrsfs}
\usepackage{float}
\usepackage{subfigure}
\usepackage{indentfirst}
\usepackage[ruled,linesnumbered]{algorithm2e}
\usepackage{setspace}
\usepackage{color}
\usepackage{lineno}  
\usepackage{threeparttable}  
\usepackage{enumerate}
\usepackage[dvipsnames]{xcolor}
\usepackage{tikz}  
\usepackage{makecell}  
\usepackage{titlesec}  
\titleformat{\section}{\bf\large}{\thesection.\,}{0.24em}{}
\titlespacing{\section}{0cm}{*2}{*2}
\titleformat{\subsection}{\bf}{\thesubsection.\enspace}{0.24em}{}
\titlespacing{\subsection}{0cm}{*1.5}{*1.5}

\let\oldequation\equation
\let\oldendequation\endequation
\renewenvironment{equation}{\linenomathNonumbers\oldequation}{\oldendequation\endlinenomath}

\usetikzlibrary{arrows,arrows.meta,shapes,chains,positioning, fit, calc,decorations.markings}

\definecolor{c1}{HTML}{FDB515}
\definecolor{c2}{HTML}{FDE8E9}
\definecolor{c3}{HTML}{79CCEE}
\definecolor{c4}{HTML}{4E9595}
\definecolor{c5}{HTML}{D6B0B0}
\definecolor{c6}{HTML}{8BABD4}

\numberwithin{equation}{section}
\theoremstyle{definition}

\newtheorem{example}{Example}
\newtheorem{definition}{Definition}

\newtheorem{remark}{Remark}
\makeatletter
\renewcommand\normalsize{%
	\abovedisplayskip 3\p@ \@plus2\p@ \@minus5\p@
	\abovedisplayshortskip \z@ \@plus3\p@
	\belowdisplayshortskip 3\p@ \@plus3\p@ \@minus3\p@
	\belowdisplayskip \abovedisplayskip
	\let\@listi\@listI}
\makeatother

\title{GFDC: A Granule Fusion Density-Based Clustering with \\ Evidential Reasoning}
\author
{Mingjie Cai $^{a,b}$  \hspace{0.3cm} Zhishan Wu$^{a}$ \hspace{0.3cm} Qingguo Li$^{a}$ \hspace{0.3cm} Feng Xu $^{a}$
	\thanks{Corresponding author:  fengxuphd@163.com (F. Xu)
		\newline\mbox{}\hspace{0.55cm}
		E-mail address:  cmjlong@163.com (M. Cai); zhishanwu@hnu.edu.cn (Z. Wu);  liqingguoli@aliyun.com (Q. Li);
		jie\_jpu@163.com (J. Zhou)
		}\hspace{0.3cm}
	Jie Zhou$^{c}$\\
	\small {$^{a}$ School of Mathematics, Hunan University}\\
	\small {Changsha, Hunan, 410082, P.R. China}\\
	\small {$^{b}$ Shenzhen Research Institute of Hunan University}\\
	\small {Shenzhen, Guangdong, 518000, P.R. China}\\
	\small {$^{c}$ School of Computer Science and Software Engineering, Shenzhen University}\\
	\small {Shenzhen, Guangdong, 518060, P.R. China}}
\date{}

\begin{document}

\maketitle 
\vspace{0cm}
\begin{center}
\begin{quote}
{{\bf Abstract:} Currently, density-based clustering algorithms are widely applied because they can detect clusters with arbitrary shapes. However, they perform poorly in measuring global density, determining reasonable cluster centers or structures, assigning samples accurately and handling data with large density differences among clusters. To overcome their drawbacks, this paper proposes a granule fusion density-based clustering with evidential reasoning (GFDC). Both local and global densities of samples are measured by a sparse degree metric first. Then information granules are generated in high-density and low-density regions, assisting in processing clusters with significant density differences. Further, three novel granule fusion strategies are utilized to combine granules into stable cluster structures, helping to detect clusters with arbitrary shapes. Finally, by an assignment method developed from Dempster-Shafer theory, unstable samples are assigned. After using GFDC, a reasonable clustering result and some identified outliers can be obtained. The experimental results on extensive datasets demonstrate the effectiveness of GFDC.
				
}
\end{quote}
\end{center}

\section{Introduction\label{section: Introduction}}	

Clustering is an important method of grouping samples in data and each group is called a cluster. If samples belong to the same cluster, they have a high degree of similarity to each other, and vice versa. As an unsupervised machine learning technique, clustering can mine the inherent structural information of data without any prior knowledge. Hence, clustering algorithms are developing at a faster pace with a better knowledge framework and more widespread application scenarios. In general, existing clustering algorithms can be classified into the following categories: partitioning clustering (e.g., $k$-means \cite{k-means}, $k$-means++ \cite{k-means++}) , hierarchical clustering  (e.g., BIRCH \cite{BIRCH}), graph-based clustering (e.g., SC \cite{SC2007}), density-based clustering (e.g., DBSCAN \cite{DBSCAN}, DPC \cite{DPC}), grid-based clustering (e.g., STING \cite{STING}),  model-based clustering (e.g., EM \cite{EM}), etc. A common problem with the above categories of clustering methods except for density-based clustering algorithms is that they cannot handle data with arbitrary shapes of clusters, whereas most density-based clustering algorithms have the advantage of handling more of these data. A novel and effective density-based clustering algorithm (DPC) was proposed by Rodriguez and Laio in 2014 \cite{DPC}, which has been widely used and studied. The main idea of DPC is that the density of cluster centers is higher than that of their neighbors, and the distance between cluster centers and other denser samples is relatively greater. Therefore, the algorithm defines formulae to measure the density and distance of samples, and then selects cluster centers and identifies outliers by a decision graph, which is a scatter plot consisting of density and distance. After determining the cluster centers, each remaining sample is assigned to the cluster where the nearest sample with a greater density than it is located. 

DPC is very powerful in clustering but there are still some problems to be solved. Firstly, the main idea of DPC is ingenious, however, the parameter cutoff distance in the density formula is difficult to determine, and an improper parameter can affect the generation of a decision graph and the selection of cluster centers. In order to address the difficulty of determining cutoff distance, some scholars improve the formula of density measure with the idea of k-nearest neighbor \cite{DPC-KNN,FKNN-DPC,ADPC-KNN,DPC-DBFN,CFDPC,LDP-SC,DLORE-DP,Hou2020,IDPC}. Furthermore, other scholars question the validity of density metrics, so they introduce the concept of belief in the Dempster-Shafer theory \cite{Dempster1967} to improve the density measure formula \cite{BPEC,CBP-EKNN,BPDNEC}. From the experimental results of the above algorithms, it turns out that the difficulty in determining the parameter is indeed reduced by replacing the cutoff distance with the k-nearest neighbor. Secondly, although DPC can select cluster centers so as to determine the number of clusters, in fact, the cluster centers cannot always be easily determined by a decision graph. Because it may happen that, in the upper right region of the decision graph, there is only one sample or a large number of samples overlapping each other. What's more, the step of selecting cluster centers makes the algorithm impossible to achieve full process automation. To solve this problem, the authors of DPC propose a hint that cluster centers can be chosen from the samples with the highest product of density and distance. ADPC-KNN \cite{ADPC-KNN} regards the samples with a distance greater than the cutoff distance as initial cluster centers. To measure the likelihood of each sample being a cluster center, a scoring formula with density and distance is exploited in DPC-DBFN \cite{DPC-DBFN}, and the samples with the highest score are considered as cluster centers. These algorithms \cite{ADPC-KNN,DPC-DBFN,CFDPC,LDP-SC} can determine cluster centers simply and achieve full process automation. Thirdly, the assignment strategy of DPC is simple and brilliant, but it produces unreasonable results when cluster centers are chosen improperly, and the errors are propagated rapidly. As a result, many assignment strategies are proposed \cite{FKNN-DPC,ADPC-KNN,BPEC,PPC,CBP-EKNN,DPC-DBFN,CFDPC,LDP-SC,Hou2020,IDPC,BPDNEC,3W-DPET}. Some of them are not based on selected cluster centers instead  initial clusters or other stable structures \cite{IDPC,DPC-DBFN,CBP-EKNN}, which is more beneficial to avoid budding errors and propagating errors rapidly.

Information granules originated in the literature \cite{Zadeh1979} proposed by Zadeh, in which the concept of fuzzy granules and information granularity was first introduced and discussed. Information granules are the basic elements of granular computing \cite{Zadeh1997}. The process of generating information granules is called information granulation. Through information granulation, data can be expressed by information granules according to a certain relationship of data. In other words, due to information granulation, elements that have similar characteristics in data can be aggregated and the purpose of simplifying data while keeping the data structure unchanged can be achieved. Because of such good properties, information granulation and information granules are favored by more and more scholars. Nowadays, an increasing number of studies demonstrate that excellent results can be achieved through the inclusion of information granules into machine learning. For example, GSVM (granular support vector machine) decomposes a linearly non-separable problem into multiple linearly separable problems by generating information granules and building an SVM in each granule \cite{GSVM}. Ding proposes a new fuzzy support vector machine based on information granulation, which granulates a dataset with FCM (fuzzy $c$-means) to get multiple granules and builds a classifier only on the mixed granules because mixed granules contain more useful boundary information \cite{FSVM-FIG}. Recently, Xia introduces the concept of a granular ball, which is a granule granulated by $k$-means and represented by a hypersphere structure, and presents a granular classifier framework base on granular balls \cite {granular_ball1}. In addition, he also uses granular balls to reduce the computation time of distances among centers and accelerate $k$-means \cite{granular_ball2}. As can be seen, information granulation not only reduces the difficulty of problem processing, but also discovers the most informative samples in data. Moreover, different samples  may be suitable for different processing methods, so it is reasonable to aggregate samples with the same character by information granulation and take into account processing strategies separately in terms of information granules. Therefore, it can be said that granulation is beneficial to improving the efficiency and quality of data mining as well as making classification, clustering and more machine learning algorithms work better.


Evidential reasoning originated from a framework for analyzing arguments \cite{ER1958}, which was a quantitative method used to make decisions based on acquirable evidence \cite{ER2002}. Dempster-Shafer theory, proposed in the 1960s \cite{Dempster1967}, can be considered as an approach of evidential reasoning. Dempster-Shafer theory represents the concepts of uncertainty and unknown by introducing belief function and plausibility function, and combines pieces of evidence into conclusions by implementing a rule of combination. Therefore, it is particularly good at dealing with the representation and fusion of uncertain information. Due to its advantages, Dempster-Shafer theory has become an important technological approach and effective expression tool for machine learning nowadays. And it has been widely applied in the fields of information fusion \cite{Li2017}, classification \cite{Lian2015}, clustering, ensemble learning \cite{Wang2020}, image segmentation \cite{Lian2018}, medical diagnosis \cite{DS-apply2022}, etc. In 2004, Denoeux and Masson proposed EVCLUS \cite{EVCLUS}, the first method combining Dempster-Shafer theory and clustering. What's more, they first introduced the concept of credal partition, by which the uncertainty of data can be well judged and represented. Remarkably, the credal partition is a general model of partition and it can generalize crisp partition, fuzzy partition, rough partition, etc. After that, many effective evidential clustering algorithms are researched and proposed \cite{ECM,EKNNclus,ECMdd,Zhang2021,BPEC,CBP-EKNN,BPDNEC}, not least of which is the improvement of density-based algorithms using Dempster-Shafer theory.

In fact, the improved algorithms of DPC overcome the drawbacks in DPC to a certain extent, but there are still some problems to be solved. In the first place, despite the difficulty in determining the parameter is reduced by some improved algorithms, only the local density is measured. It results that these algorithms are still confused about data where multiple density peaks exist in the same cluster and tend to misjudge the cluster centers. In the second place, those strategies for choosing centers proposed by the above algorithms only solve the problem that how to choose cluster centers but not how to choose good cluster centers. What’s more, the fact that some datasets do not have so-called cluster centers, so it is meaningless for them to find cluster centers in a geometric sense. Instead, detecting initial clusters that can reflect the true shapes and distributions of the clusters may be more meaningful. In the third place, DPC does not always perform well on datasets that have clusters with arbitrary shapes or with large differences in density. 

Motivated by the above discussion, a granule fusion density-based clustering with evidential reasoning (abbreviated as GFDC) is proposed in this paper, which can process data with arbitrary cluster shapes and large density differences among clusters. In GFDC, both local and global densities of samples are measured, and then samples are granulated considering their densities. Next, three effective fusion strategies of information granules are designed for generating stable initial clusters, after which the unstable samples outside the initial clusters are assigned by an improved evidential assignment method. Finally, a reasonable clustering result and identified outliers are obtained. In detail, the main contributions of GFDC are as follows:
\begin{enumerate}[1)]
	\setlength{\itemsep}{-1mm}
\item By integrating the notion of optimal information granularity and k-nearest neighbor, this paper proposes a novel sparse degree metric to measure both local and global densities of samples.
\item The proposed algorithm granulates samples based on their sparse degrees, generating granules not only in high-density regions but also in low-density regions with clear density gradients, which allows the algorithm to cope with data where there are significant differences in density among clusters.
\item Based on intersection relationship, density transmission and distance, three fusion strategies are designed in this paper to combine information granules into structurally stable granule-clusters, granule-flocks and initial clusters, which enables the proposed algorithm to detect clusters with arbitrary shapes.
\item An improved evidential assignment method is proposed in this paper for assigning the remaining unstable samples based on stable initial clusters, which makes the assignment results more reasonable than that based on selected cluster centers and also can identify outliers.
\end{enumerate}

The rest of this paper is organized as follows: In \autoref{section: preliminaries}, the basic concepts of granular computing, the Dempster-Shafer theory and credal partition are briefly recalled. The details of the proposed algorithm GFDC are introduced in \autoref{section: method}. \autoref{section: Experimental results} presents the experimental results of GFDC on extensive synthetic and real-world datasets. The conclusions of the paper are given in \autoref{section: conclusion}.

\section{Preliminaries\label{section: preliminaries}}
The purpose of this section is to briefly review some necessary background information for readers, including the connotation of information granules \cite{Zadeh1997,granule} (\autoref{subsection: Granular computing}), some basis of the Dempster-Shafer theory \cite{Dempster1967,Shafer1976,Smets1994,Smets1990} (\autoref{subsection: Theory of evidence}) and the concept of credal partition \cite{EVCLUS} (\autoref{subsection: Credal partition}).

\subsection{Information granules\label{subsection: Granular computing}}

\emph{Information granules} are collections of elements (samples) gathered together according to indistinguishability, similarity, proximity or functionality \cite{Zadeh1997}. The process of constructing information granules is called \emph{information granulation}, which involves the abstract representation of samples and the extraction and summarization of information for data. In particular, binary relations are the frequently used information granulation strategies, and they include equivalence relations, neighborhood relations, tolerance relations and dominance relations.

\begin{definition} \cite{granule}
\emph{Given a universe $U$ and a family of binary relations $\boldsymbol R$, the \textbf{granular structure} induced by a binary relation $P \in \boldsymbol R$ can be represented as a vector $K(P) = \big(N_{P}(x_1),N_{P}(x_2), \dots,N_{P}(x_n)\big)$, where $x_i \in U$ and $N_{P}(x_i)$ is the \textbf{information granule} generated by a sample $x_i$ concerning $P$.}
\end{definition}

An information granule $N_{P}(x_i)$ is a set of samples containing at least $x_i$ in $U$. A collection of information granules is a granular structure. It is worth noting that a granule structure always satisfies $\bigcup_{x_i \in U}N_{P}(x_i) = U$, in other words, a granule structure can form a cover of the universe. In particular, a granule structure induced by a equivalence relation satisfies $N_{P}(x_i)\cap N_{P}(x_j)=\emptyset$, where $i \neq j$, which means that it can even form a partition of the universe.

\emph{Information granularity} of an information granule is used to measure the uncertainty of the actual structure of the granule and represents the discernibility ability of information in the granule. A smaller information granularity is associated with a stronger discernibility ability. However, the fact is that it is not better to have a smaller information granularity, but to determine the optimal information granularity according to the actual situation. Since information granulation is a process of abstract expression and extraction of information with clear objectives, corresponding to the degree and perspective of abstract expression as well as the depth and breadth of extracted information, the information granularity must be determined by the practical requirements when performing information granulation.

\subsection{Dempster-Shafer theory \label{subsection: Theory of evidence}}
\emph{Dempster-Shafer theory}, also known as the evidence theory or theory of belief functions \cite{Dempster1967,Shafer1976} is an important method of uncertain reasoning.

Given a finite and unordered set $\Omega=\{\omega_1, \omega_2, \dots, \omega_c\}$ , which represents the possible values of a variable $x$, called the \emph{frame of discernment}. Given a mass function is defined as a mapping from $2^{\Omega}$ to $[0,1]$ such that
\begin{equation}
\sum_{A \subseteq \Omega} m^{\Omega}(A)=1 \label{basic belief assignment},
\end{equation}
where $2^{\Omega}$ is the power set of $\Omega$, with $2^{\Omega}=\{\emptyset, \omega_1, \omega_2, \dots, \{\omega_1, \omega_2\}, \dots, \Omega\}$. The mass function is called the \emph{basic belief assignment} (BBA) \cite{Smets1994} and the value of $m^{\Omega}(A)$ are called the \emph{basic belief masses}, which indicates a degree of belief assigned to the hypothesis that $x$ is associated with $A$. The subsets $A$ of ${\Omega}$ with $m^{\Omega}(A)>0$ are called \emph{focal sets} of $m^{\Omega}$. It is said to be normal that a BBA such that $m^{\Omega}(\emptyset)=0$, but this condition may be relaxed if the \emph{open-world assumption} state that the set $\Omega$ might be incomplete can be accepted \cite{Smets1990}. That is to say, $m^{\Omega}(\emptyset)$ denotes a degree of belief assigned to the hypothesis that $x$ might not lie in $\Omega$.

Assume that there are two BBAs $m_1$ and $m_2$, which represent distinct items of evidence, to be combined. The standard way of the combination of $m_1$ and $m_2$ is to utilize the conjunctive sum operation \textcircled{$\cap$} defined as
\begin{equation}
(m_{1} \textcircled{$\cap$} m_{2})(A) \triangleq \sum_{B \cap C=A} m_{1}(B) m_{2}(C), \label{conjunctive sum}
\end{equation}
for all $A \subseteq \Omega$. The \emph{degree of conflict} between $m_1$ and $m_2$ is defined as
\begin{equation}
\mathcal{K} \triangleq(m_{1} \textcircled{$\cap$} m_{2})(\emptyset)=\sum_{B \cap C=\emptyset} m_{1}(B) m_{2}(C), \label{degree of conflict}
\end{equation}
which denotes a degree of inconsistency between two information sources. If necessary, the normality condition $m^{\Omega}(\emptyset)$ can be recovered by dividing each mass $(m_{1} \textcircled{$\cap$} m_{2})(A)$ by $1-\mathcal{K}$. Based on the conjunctive sum operation \textcircled{$\cap$} and the degree of conflict, the \emph{Dempster's rule of combination} is widely used to combine two BBAs:
\begin{equation}
(m_{1} \textcircled{$+$} m_{2})(A) \triangleq \frac{(m_{1} \textcircled{$\cap$} m_{2})(A)}{1-\mathcal{K}} = \frac{1}{1-\mathcal{K}} \sum_{B \cap C=A} m_{1}(B) m_{2}(C), \label{Dempster's rule of combination}
\end{equation}
for all $A \subseteq \Omega $ and $ A \ne \emptyset$. It is worth noting that, both combination rules have two excellent properties for combination operators, commutative and associative.

\subsection{Credal partition\label{subsection: Credal partition}}
Consider a set $U=\{x_1,x_2,\dots,x_n\}$ of $n$ samples and suppose that a set $\Omega=\{Cl_1, Cl_2, \dots, Cl_C\}$ of $C$ clusters form a partition of $U$. For the problem that which cluster each sample belongs to, if complete knowledge is available, then a \emph{crisp partition} can be acquired (as shown in \autoref{table: crisp parition}), which is represented by binary variables $u_{ip}$. Sample $x_i$ belongs to cluster $Cl_p$ if $u_{ip}=1$ and sample $x_i$ does not belong to cluster $Cl_p$ if $u_{ip}=0$. Assume that only partial knowledge is available, a \emph{fuzzy partition} might be acquired (as shown in \autoref{table: fuzzy partition}), which is represented by fuzzy membership $f_{ip}$. The probability that sample $x_i$ belongs to cluster $Cl_p$ is 0.8 if $f_{ip}=0.8$. Furthermore, based on partial knowledge, a \emph{credal partition} might be acquired (as shown in \autoref{table: credal partition}), which is represented by BBAs on the set $\Omega$, as the following example illustrates.

\begin{table}[t]\footnotesize
\setlength{\abovecaptionskip}{0cm} 
\setlength{\belowcaptionskip}{-0.2cm}
    \begin{minipage}[t]{0.49\linewidth}
		\caption{An example of crisp partition.}
		\label{table: crisp parition}
            \tabcolsep0.3in
			\begin{tabular}{cccc}
				\toprule  
			      & $Cl_1$ & $Cl_2$ & $Cl_3$ \\
                \midrule  
				$u_{1p}$ & 1 & 0 & 0 \\
                $u_{2p}$ & 0 & 0 & 1 \\
                $u_{3p}$ & 0 & 1 & 0 \\
                \bottomrule  
			\end{tabular}
        \end{minipage}
	\hfill
	\begin{minipage}[t]{0.49\linewidth}
		\caption{An example of fuzzy partition.}
		\label{table: fuzzy partition}
            \tabcolsep0.25in
			\begin{tabular}{cccc}
				\toprule  
			      & $Cl_1$ & $Cl_2$ & $Cl_3$ \\
                \midrule  
				$f_{1p}$ & 0.875 & 0.08 & 0.045 \\
                $f_{2p}$ & 0.183 & 0.208 & 0.609 \\
                $f_{3p}$ & 0.03 & 0.51 & 0.46 \\
                \bottomrule  
			\end{tabular}
	\end{minipage}
\end{table}

\begin{table}[t]\footnotesize
\setlength{\abovecaptionskip}{0cm} 
\setlength{\belowcaptionskip}{0cm}
\caption{An example of credal partition on $\Omega=\{Cl_1,Cl_2,Cl_3\}$.}
\label{table: credal partition}
\tabcolsep0.185in
	\begin{tabular}{ccccccccc}
		\toprule  
	      & $\varnothing$ & $\{Cl_1\}$ & $\{Cl_2\}$ & $\{Cl_3\}$ & $\{Cl_1,Cl_2\}$ & $\{Cl_1,Cl_3\}$ & $\{Cl_2,Cl_3\}$ & $\Omega$ \\
        \midrule  
        $m_1^{\Omega}(\cdot)$ & 0 & 0.8 & 0.02 & 0.01 & 0.1 & 0.05 & 0.02 & 0 \\
        $m_2^{\Omega}(\cdot)$ & 0 & 0.1 & 0.1 & 0.5 & 0.05 & 0.05 &0.1 & 0.1 \\
        $m_3^{\Omega}(\cdot)$ & 0 & 0.01 & 0.35 & 0.3 & 0.02 & 0.02 & 0.3 & 0\\
        $m_4^{\Omega}(\cdot)$ & 0 & 1 & 0 & 0 & 0 & 0 & 0 & 0\\
        $m_5^{\Omega}(\cdot)$ & 1 & 0 & 0 & 0 & 0 & 0 & 0 & 0\\
        $m_6^{\Omega}(\cdot)$ & 0 & 0 & 0 & 0 & 0 & 0 & 0 & 1\\
        \bottomrule  
	\end{tabular}
\end{table}

\begin{example}
	Given an example of credal partition on $U=\{x_1, x_2, x_3, x_4, x_5, x_6\}$ of six samples and $\Omega=\{Cl_1, Cl_2, Cl_3\}$ of three clusters showing in \autoref{table: credal partition}, whose element in \autoref{table: credal partition} is the value of BBA for each sample. The situation of each sample is illustrated: The cases of samples $x_1$, $x_2$ and $x_3$ are the most common situation in a credal partition based on only partial knowledge. The credal partition not only gives the degree of belief that samples belong to single-clusters $Cl_1$, $Cl_2$ and $Cl_3$, but also gives the degree of belief that samples belong to meta-clusters $\{Cl_1, Cl_2\}$, $\{Cl_1, Cl_3\}$, $\{Cl_2, Cl_3\}$ and $\Omega$. The cluster of sample $x_4$ is known certainly from $m_4^{\Omega}(Cl_1)=1$, whereas the cluster of sample $x_6$ is absolutely unknown from $m_6^{\Omega}(\Omega)=1$. Ultimately, it can be known that sample $x_5$ does not lie in $\Omega$, which is indicated by $m_5^{\Omega}(\emptyset)=1$.
\end{example}

	As shown above, a \emph{credal partition} of $n$ samples $U=\{x_1, x_2, \dots, x_n\}$ is defined as the $n$-tuple BBAs $M=\{m_1^{\Omega}, m_2^{\Omega}, \dots, m_n^{\Omega}\}$ \cite{EVCLUS}. Interestingly, the \emph{credal partition} can be regarded as a general model of partition, and the \emph{crisp partition} and the \emph{fuzzy partition} are two particular cases of the \emph{credal partition}:
\begin{itemize}
	\setlength{\itemsep}{-1mm}

\item When the \emph{focal set} of all BBAs $m^{\Omega}$ are singletons of $\Omega$ and the whole \emph{basic belief mass} are allocated to a unique singleton of $\Omega$, then \emph{credal partition} $M$ degenerates into \emph{crisp partition};
\item When the \emph{focal set} of all BBAs $m^{\Omega}$ are singletons of $\Omega$ and all BBAs $m^{\Omega}$ are equivalent to probability functions, then \emph{credal partition} $M$ degenerates into \emph{fuzzy partition}.
\end{itemize}

\section{Density-based Clustering Algorithm with Granulation, Fusion and Evidential assignment \label{section: method}}

\begin{figure}[H]\footnotesize
	\centering
	\setlength{\abovecaptionskip}{0pt}  
	\includegraphics[width=15cm]{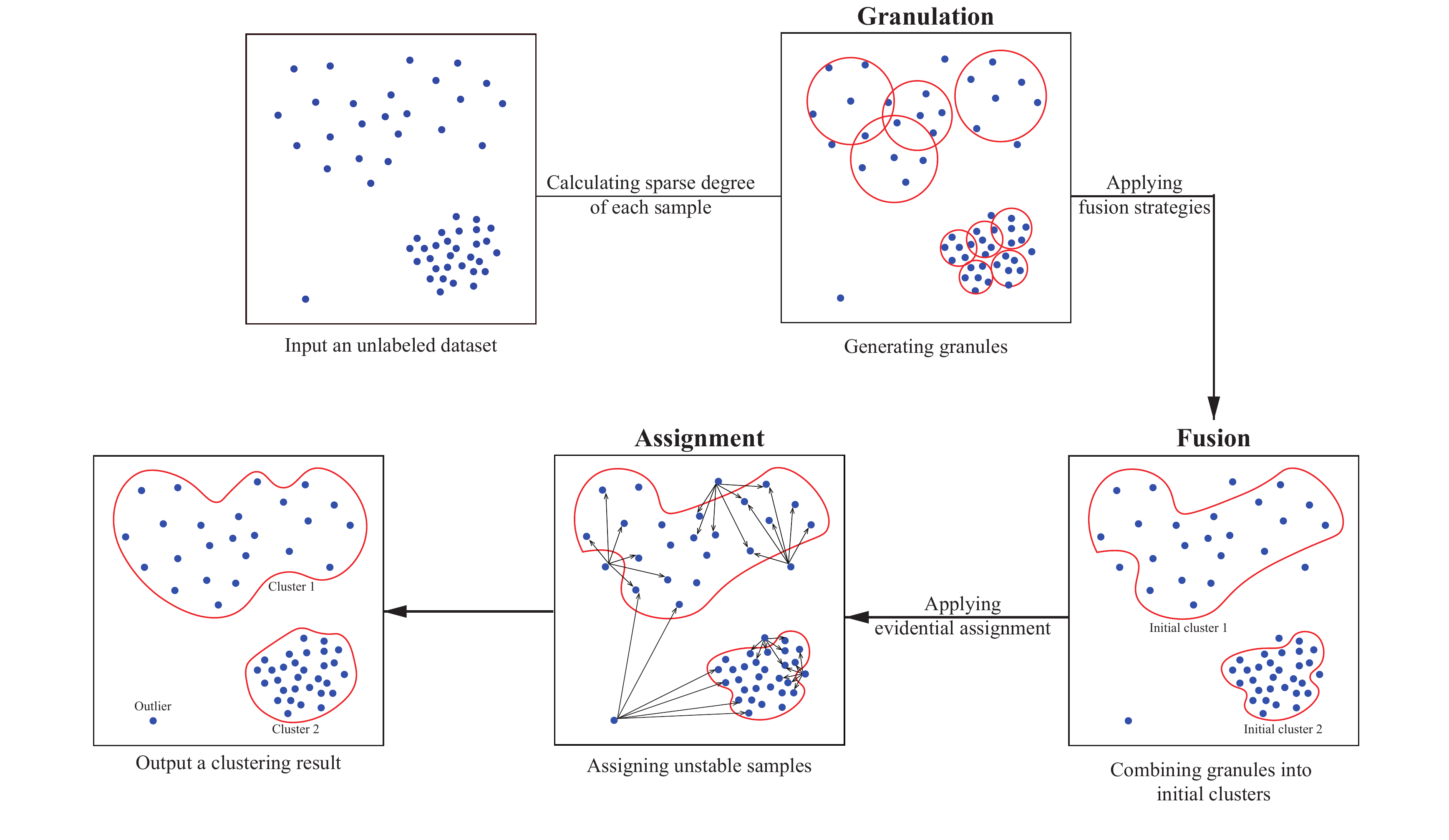}
	\caption{The framework of the proposed algorithm.}
	\label{fig: framework}
\end{figure}

This section aims at expounding the principle, process, and detail of the proposed clustering algorithm named GFDC. As illustrated by the framework of the proposed algorithm in \autoref{fig: framework}, GFDC consists of three main steps: \textbf{(1) Granulation:} calculating the sparse degrees of samples and aggregating the samples into granules based on the sparse degrees; \textbf{(2) Fusion:} combining granules into stable granule-related structures and initial clusters; \textbf{(3) Assignment:} assigning remaining unstable samples. In the first step, a sparse degree formula for measuring the densities of samples is designed by finding the adaptive neighborhood radius and combining the idea of k-nearest neighbor. It is remarkably different from other density measure formulas, because the sparse degree can provide both local and global density information of a sample. After that, multiple granules are generated based on the sparse degrees of samples, and the center of each granule is the peak density sample in the granule. In the second step, three fusion strategies are designed and used sequentially to combine granules into different forms of stable granule-related structures and obtain initial clusters. What's more, these fusion strategies are based on intersection relationship, density transmission and distance respectively. In the third step, an improved assignment method on the basis of Dempster-Shafer theory is applied to assign the remaining unstable samples and identify outliers. The details of each step of GFDC are discussed in \autoref{subsection: Sample's sparse degree} to \autoref{subsection: Unstable sample's evidential assignment}. At the end of this section, the pseudo-code of the proposed algorithm is demonstrated and the time complexity of GFDC is analyzed in \autoref{subsection: Analysis of algorithm}.

\subsection{Sparse degrees of samples\label{subsection: Sample's sparse degree}}

Without loss of generality, assume that the dataset $U$ consists of $n$ samples $\{x_1,x_2,\dots,x_n\}$ and $w$ attributes, with the attribute values of a sample $x_i$ represented as $\{x_{i1},x_{i2},\dots,x_{iw}\}$.

Considering the shortcomings of the density measure formulae in DPC and its variations discussed above, a formula that can measure both local and global densities of samples with respect to the whole dataset is designed, by combining the adaptive neighborhood radius and the radius of $k$th nearest neighbor.

At first, choosing different neighborhood radii, each sample can form the corresponding neighborhood:
\begin{equation}
\delta (x_i) = \{x_j \mid x_j \in U, d(x_i, x_j) \le \delta\}, \notag
\end{equation}
where $d(x_i, x_j) = d_{ij}=||x_i-x_j||_2$ denotes the Euclidean distance between samples $x_i$ and $x_j$. The neighborhood radius $\delta$ is from 0 to $d(x_i^{\prime},x_j^{\prime})$, where $x_i^{\prime}$ and $x_j^{\prime}$ have a maximum distance in $U$. In fact, the meaningful neighborhood radius of a sample $x_i$ should be the distance between $x_i$ and any other sample in $U$:
\begin{equation}
\delta (x_i, d_{ij}) = \{x_z \mid x_z \in U, d(x_i, x_z) \le d_{ij}\}. \label{neighborhood}
\end{equation}
In order to measure the density in the neighborhood of a sample, an understandable formula can be used to express the relative density of a sample in its neighborhood with a radius:
\begin{equation}
\rho^{*}(x_i,d_{ij})=\frac{|\delta (x_i, d_{ij})|}{{(d_{ij})}^{w}}, \label{relative density}
\end{equation}
where $|\cdot|$ denotes the cardinality of a set. The numerator of \autoref{relative density} indicates the number of samples in the neighborhood of sample $x_i$ with $d_{ij}$ as the neighborhood radius, and the denominator is the $w$th power of neighborhood radius and the intent of the power is to eliminate differences in dimensional space.

Actually, different neighborhood radii can be regarded as corresponding to different information granularities. It is necessary to find a suitable information granularity to reflect the global density of samples in $U$. We consider that the neighborhood radius, which makes the relative density of a sample highest, is the optimal information granularity, and it can reflect the global density of the sample:
\begin{equation}
r^{*}(x_i)=\underset{d_{ij}}{\operatorname{arg~max}} \frac{|\delta (x_i, d_{ij})|}{{(d_{ij})}^{w}}, \label{neighborhood radius}
\end{equation}
for all $x_{j} \in U$. Further, according to the neighborhood $\delta (x_i, r^{*}(x_i)) = \{x_z \mid x_z \in U, d(x_i, x_z) \le r^{*}(x_i)\}$, we define a circle, sphere or hypersphere, where sample $x_i$ is the center and $r^{*}(x_i)$ is the radius, as the \textbf{density-limit-granule}. It can be known that a density-limit-granule has the highest relative density among all information granules centered on $x_i$, in other words, a density-limit-granule with $r^{*}(x_i)$ can reflect the density limitation of $x_i$. For the density-limit-granules of different samples, the higher the relative density of a density-limit-granule, the more samples have in the density-limit-granule or the smaller neighborhood radius has. Therefore, without considering the number of samples tentatively, if the neighborhood radius $r^{*}(x_i)$ of the density-limit-granule of sample $x_i$ is smaller, it indicates that the relative density of the density-limit-granule of $x_i$ is likely to be higher, namely, $x_i$ is likely to be a sample with a higher density. 

\begin{figure}[ht]\footnotesize
	\centering
	\vspace{0pt}  
	\setlength{\abovecaptionskip}{5pt}  
	\subfigcapskip=0pt  
	\subfigure[]{
		\includegraphics[width=5cm]{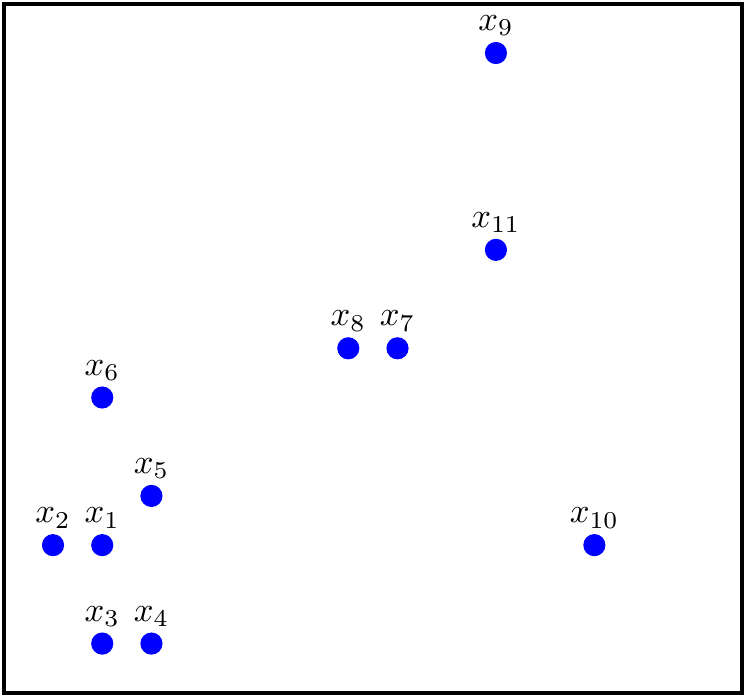}
		\label{fig: original}
	}
	\hskip 0pt
	\subfigure[]{
		\includegraphics[width=5cm]{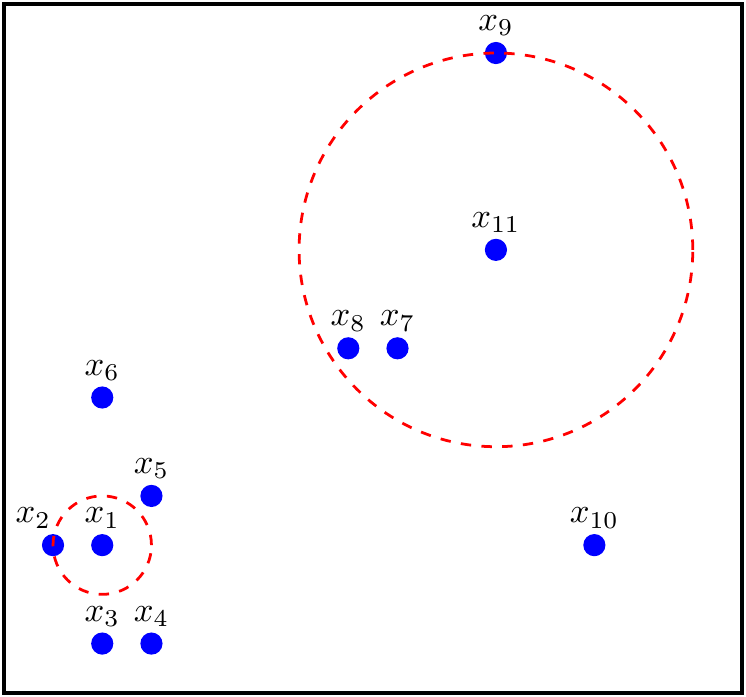}
		\label{fig: sd1}
	}
	\hskip 0pt
	\subfigure[]{
		\includegraphics[width=5cm]{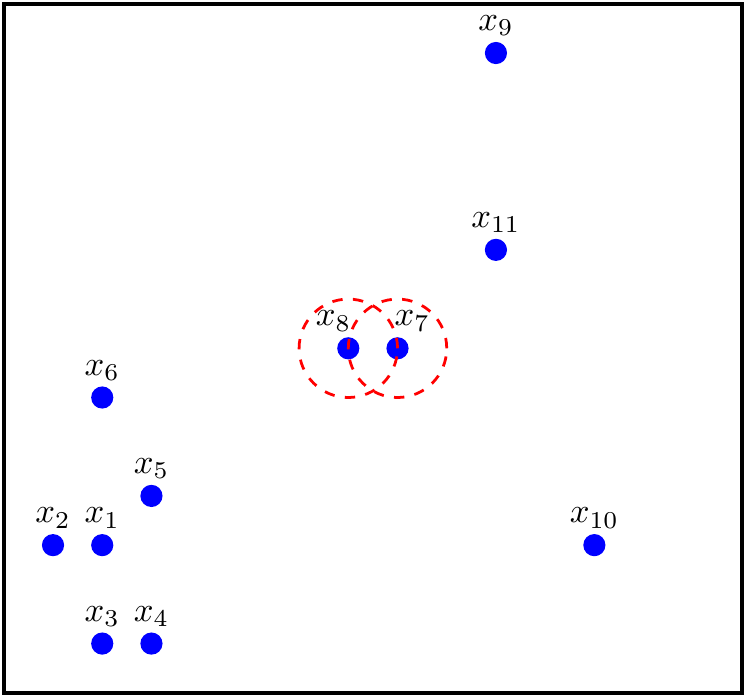}
		\label{fig: sd2}
	}
	\caption{Illustration for the density-limit-granules of samples.}
	\label{fig: Illustration for the limit cricle of samples}
\end{figure}

This observation is illustrated by a simple example from \autoref{fig: Illustration for the limit cricle of samples}. As shown in \autoref{fig: original}, it can be found that sample $x_1$ is denser, whereas sample $x_{11}$ is sparser. After calculating the neighborhood radius of the density-limit-granule of $x_1$ and $x_{11}$ respectively, the density-limit-granules are plotted in \autoref{fig: sd1}. It can be visualized that sample $x_1$ has a density-limit-granule with a smaller neighborhood radius, i.e., sample $x_1$ is more likely to be a high-density sample. Sample $x_{11}$ has a density-limit-granule with a larger neighborhood radius, i.e., sample $x_{11}$ is less likely to be a low-density sample. This means that, to some extent, $r^{*}(x_i)$ may be able to measure the density of a sample. Moreover, $r^{*}(x_i)$ measures the global density because it is obtained by taking into account the relationships between a sample and all other samples.

However, as shown in \autoref{fig: sd2}, both samples $x_7$ and $x_8$ are sparser in terms of the whole dataset, but they are too close to each other such that the neighborhood radii of their density-limit-granules both are the distance between them, i.e., correspondingly $r^{*}(x_7)$ and $r^{*}(x_8)$ are small. That is to say, according to the above discussion about the relationships between the neighborhood radius of the density-limit-granule and the density, $x_7$ and $x_8$ are most likely to be mistaken as high-density samples. It can be believed that this problem stems from the lack of considering the number of samples within the density-limit-granule according to the above discussion. Nevertheless, since the number of samples within the density-limit-granule is correlated with the neighborhood radius, to avoid the impact of correlation, another density measurement is introduced. The radius corresponding to the $k$th nearest neighbor, which takes into account the number of samples and is able to measure the local density, is combined with the neighborhood radius of the density-limit-granule in this paper. Further, a new formula is proposed to measure both local and global densities of samples as below.
\begin{definition}[\textbf{sparse degree}]
\emph{The sparse degree of sample $x_i$, denoted by $sd(x_i)$, is defined by
\begin{equation}
sd(x_i)=r^{*}(x_i)+d\big(x_i,x_i^{(k)}\big), \label{sparse degree}
\end{equation}
where $x_i^{(k)}$ indicates the $k$th nearest neighbor of sample $x_i$ and $d\big(x_i,x_i^{(k)}\big)$ means the distance between $x_i$ and its $k$th nearest neighbor. Generally, the value of $k$ is taken to be $\lceil log_2(n) \rceil$, where $\lceil \cdot \rceil$ denotes rounding up for real numbers.}
\end{definition}

Obviously, both $d\big(x_i,x_i^{(k)}\big)$ and $r^{*}(x_i)$ change in the same direction, which means that, given the value of $k$, the denser the sample, the smaller $d\big(x_i,x_i^{(k)}\big)$ is, and the sparser the sample, the larger $d\big(x_i,x_i^{(k)}\big)$ is. Therefore, it is reasonable to combine $d\big(x_i,x_i^{(k)}\big)$ and $r^{*}(x_i)$ to form a new and effective density measure.

\subsection{Granulation of samples\label{subsection: Granulation}}
\begin{definition}[\textbf{granule}]
\emph{Given a positive integer of $k$, according to the sparse degree, a set of samples can be called a \textbf{granule} $G$ in this paper, if two requirements are satisfied: $\exists x_i \in U$, such that\\
(1) $G=\{x_j \mid x_j \in U, d(x_i,x_j) \le d\big(x_i,x_i^{(k-1)}\big)\}$ and $|G|=k$;\\
(2) $sd(x_i)=\underset{x_j \in G}{\operatorname{min}}~sd(x_j)$.}
\end{definition}

Thus, there are $k$ samples in a granule $G$ and the center of $G$ is with the lowest sparse degree within $G$. Furthermore, all granules and samples outside any granule can constitute a cover of the dataset, so the above process of finding granules can be known as \textbf{granulation}. From the definition of information granulation, it is known that a non-single-sample granule or single-sample can be called an information granule, but it is clear that only non-single-sample granules can reflect the structure and distribution of the original dataset. Therefore, the granules mentioned in the later refer to non-single-sample granules, and they are noted as $\{G_1,G_2,\dots,G_g\}$ (a solid line circle is used in the illustration of this paper to represent a granule in the two-dimensional plane). A simple example of granulation is given below for the convenience of understanding. 

\begin{figure}[ht]\footnotesize
\centering
\setlength{\abovecaptionskip}{5pt}  
\subfigcapskip=0pt  
\subfigure[]{
\includegraphics[width=5cm]{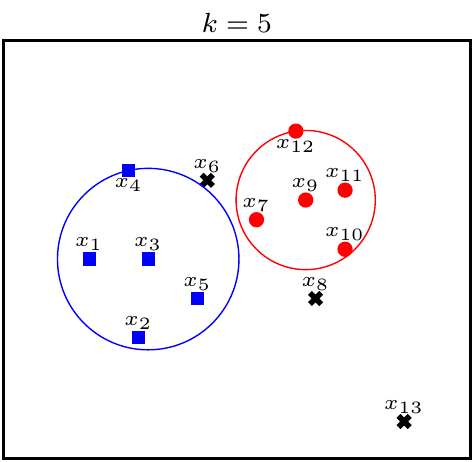}
\label{fig: k5}
}
\hskip 0pt
\subfigure[]{
\includegraphics[width=5cm]{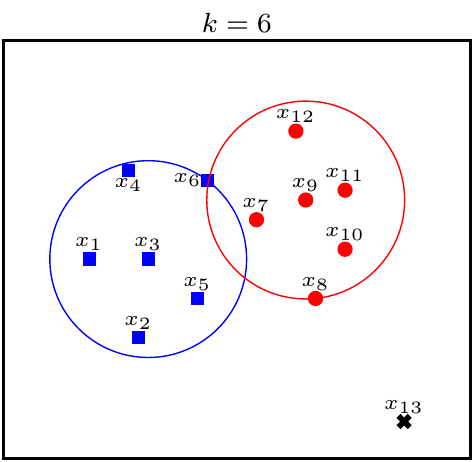}
\label{fig: k6}
}
\hskip 0pt
\subfigure[]{
\includegraphics[width=5cm]{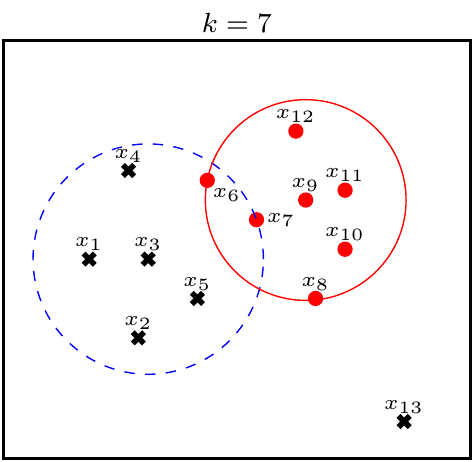}
\label{fig: k7}
}
\caption{Illustration of granulation.}
\label{fig: Illustration of granulation}
\end{figure}

\begin{example}
	For the dataset consisting of 13 samples shown in \autoref{fig: Illustration of granulation}, $k$ is taken to be 5 and the sparse degrees of samples are calculated firstly. And then with \autoref{table: the sparse degree of samples in example}, it can be found that only samples $x_3$ and $x_9$ can generate granules because the 4-nearest neighbors of the other samples contain samples with lower sparse degrees than centers. For instance, the 4-nearest neighbors of sample $x_7$ is samples $x_9$, $x_6$, $x_{10}$ and $x_{11}$, but the sparse degrees of $x_9$, $x_{10}$ and $x_{11}$ are lower than that of $x_7$, so $x_7$ can not generate a granule. As shown in \autoref{fig: k5} and \autoref{fig: k6}, it is similar to the above when $k$ takes the value of 6. The sparse degrees of the 5-nearest neighbors of sample $x_3$ are higher than that of $x_3$ and the sparse degrees of the 5-nearest neighbors of sample $x_9$ are higher than that of $x_9$. While $k$ is taken to be 7, the 6-nearest neighbors of sample $x_3$ contain samples $x_1$, $x_5$, $x_2$, $x_4$, $x_6$, and $x_7$, where the sparse degrees of samples $x_5$ and $x_7$ are lower than that of sample $x_3$, so it is impossible to generate a granule with $x_3$ as the center. At last, only one granule with sample $x_9$ as the center can be generated, which is shown in \autoref{fig: k7}. \label{example: granulation}
\end{example}

\begin{table}[ht]\footnotesize
\caption{The sparse degrees of samples in \autoref{example: granulation}.}
\label{table: the sparse degree of samples in example}
\tabcolsep0.15in
\begin{tabular}{ccccccc}
\toprule
\textbf{} & \multicolumn{2}{c}{$k=5$}                          & \multicolumn{2}{c}{$k=6$}                                   & \multicolumn{2}{c}{$k=7$}                                           \\
\cmidrule(lr){2-3} \cmidrule(lr){4-5} \cmidrule(lr){6-7}
\textbf{} & 4-nearest neighbors                  & $sd(x_i)$ & 5-nearest neighbors                           & $sd(x_i)$ & 6-nearest neighbors                                   & $sd(x_i)$ \\
\midrule
$x_1$         & $x_{3}$,$x_{2}$,$x_{4}$,$x_{5}$     & 2.0422    & $x_{3}$,$x_{2}$,$x_{4}$,$x_{5}$,$x_{6}$      & 2.3464    & $x_{3}$,$x_{2}$,$x_{4}$,$x_{5}$,$x_{6}$,$x_{7}$      & 2.8804    \\
$x_2$         & $x_{5}$,$x_{3}$,$x_{1}$,$x_{7}$     & 2.5092    & $x_{5}$,$x_{3}$,$x_{1}$,$x_{7}$,$x_{4}$      & 2.5527    & $x_{5}$,$x_{3}$,$x_{1}$,$x_{7}$,$x_{4}$,$x_{6}$      & 2.6501    \\
$x_3$         & $x_{1}$,$x_{5}$,$x_{2}$,$x_{4}$     & 1.6403    & $x_{1}$,$x_{5}$,$x_{2}$,$x_{4}$,$x_{6}$      & 1.8108    & $x_{1}$,$x_{5}$,$x_{2}$,$x_{4}$,$x_{6}$,$x_{7}$      & 2.3491    \\
$x_4$         & $x_{6}$,$x_{3}$,$x_{1}$,$x_{7}$     & 2.4614    & $x_{6}$,$x_{3}$,$x_{1}$,$x_{7}$,$x_{5}$      & 2.6878    & $x_{6}$,$x_{3}$,$x_{1}$,$x_{7}$,$x_{5}$,$x_{2}$      & 2.7313    \\
$x_5$         & $x_{3}$,$x_{2}$,$x_{7}$,$x_{1}$     & 1.9211    & $x_{3}$,$x_{2}$,$x_{7}$,$x_{1}$,$x_{8}$      & 1.9253    & $x_{3}$,$x_{2}$,$x_{7}$,$x_{1}$,$x_{8}$,$x_{6}$      & 2.1976    \\
$x_6$         & $x_{7}$,$x_{4}$,$x_{3}$,$x_{9}$     & 2.0591    & $x_{7}$,$x_{4}$,$x_{3}$,$x_{9}$,$x_{12}$    & 2.2337    & $x_{7}$,$x_{4}$,$x_{3}$,$x_{9}$,$x_{12}$,$x_{5}$    & 2.4331    \\
$x_7$         & $x_{9}$,$x_{6}$,$x_{10}$,$x_{11}$  & 1.9849    & $x_{9}$,$x_{6}$,$x_{10}$,$x_{11}$,$x_{12}$ & 2.0000    & $x_{9}$,$x_{6}$,$x_{10}$,$x_{11}$,$x_{12}$,$x_{8}$ & 2.0000    \\
$x_8$         & $x_{10}$,$x_{7}$,$x_{9}$,$x_{11}$  & 1.7831    & $x_{10}$,$x_{7}$,$x_{9}$,$x_{11}$,$x_{5}$   & 2.1234    & $x_{10}$,$x_{7}$,$x_{9}$,$x_{11}$,$x_{5}$,$x_{13}$ & 2.2110    \\
$x_9$         & $x_{11}$,$x_{7}$,$x_{10}$,$x_{12}$ & 1.4173    & $x_{11}$,$x_{7}$,$x_{10}$,$x_{12}$,$x_{8}$  & 1.4321    & $x_{11}$,$x_{7}$,$x_{10}$,$x_{12}$,$x_{8}$,$x_{6}$  & 1.8989    \\
$x_{10}$        & $x_{8}$,$x_{11}$,$x_{9}$,$x_{7}$   & 1.9403    & $x_{8}$, $x_{11}$,$x_{9}$,$x_{7}$,$x_{12}$  & 2.2056    & $x_{8}$, $x_{11}$,$x_{9}$,$x_{7}$,$x_{12}$,$x_{6}$  & 2.2215    \\
$x_{11}$        & $x_{9}$,$x_{10}$,$x_{12}$,$x_{7}$  & 1.5525    & $x_{9}$,$x_{10}$,$x_{12}$,$x_{7}$,$x_{8}$   & 1.8159    & $x_{9}$,$x_{10}$,$x_{12}$,$x_{7}$,$x_{8}$,$x_{6}$   & 2.2724    \\
$x_{12}$        & $x_{9}$, $x_{11}$,$x_{7}$,$x_{6}$   & 2.0810    & $x_{9}$,$x_{11}$,$x_{7}$,$x_{6}$,$x_{10}$   & 2.4927    & $x_{9}$, $x_{11}$,$x_{7}$,$x_{6}$,$x_{10}$,$x_{8}$   & 2.5274    \\
$x_{13}$        & $x_{8}$,$x_{10}$,$x_{11}$,$x_{5}$  & 5.6249    & $x_{8}$,$x_{10}$,$x_{11}$,$x_{5}$,$x_{9}$   & 5.7029    & $x_{8}$,$x_{10}$,$x_{11}$,$x_{5}$,$x_{9}$,$x_{7}$   & 5.9933    \\
\bottomrule
\end{tabular}
\end{table}

\begin{figure}[htbp]\footnotesize
	\centering
	\vspace{-25pt}  
	\subfigcapskip=-25pt  
	\subfigure[]{
		\includegraphics[width=8cm]{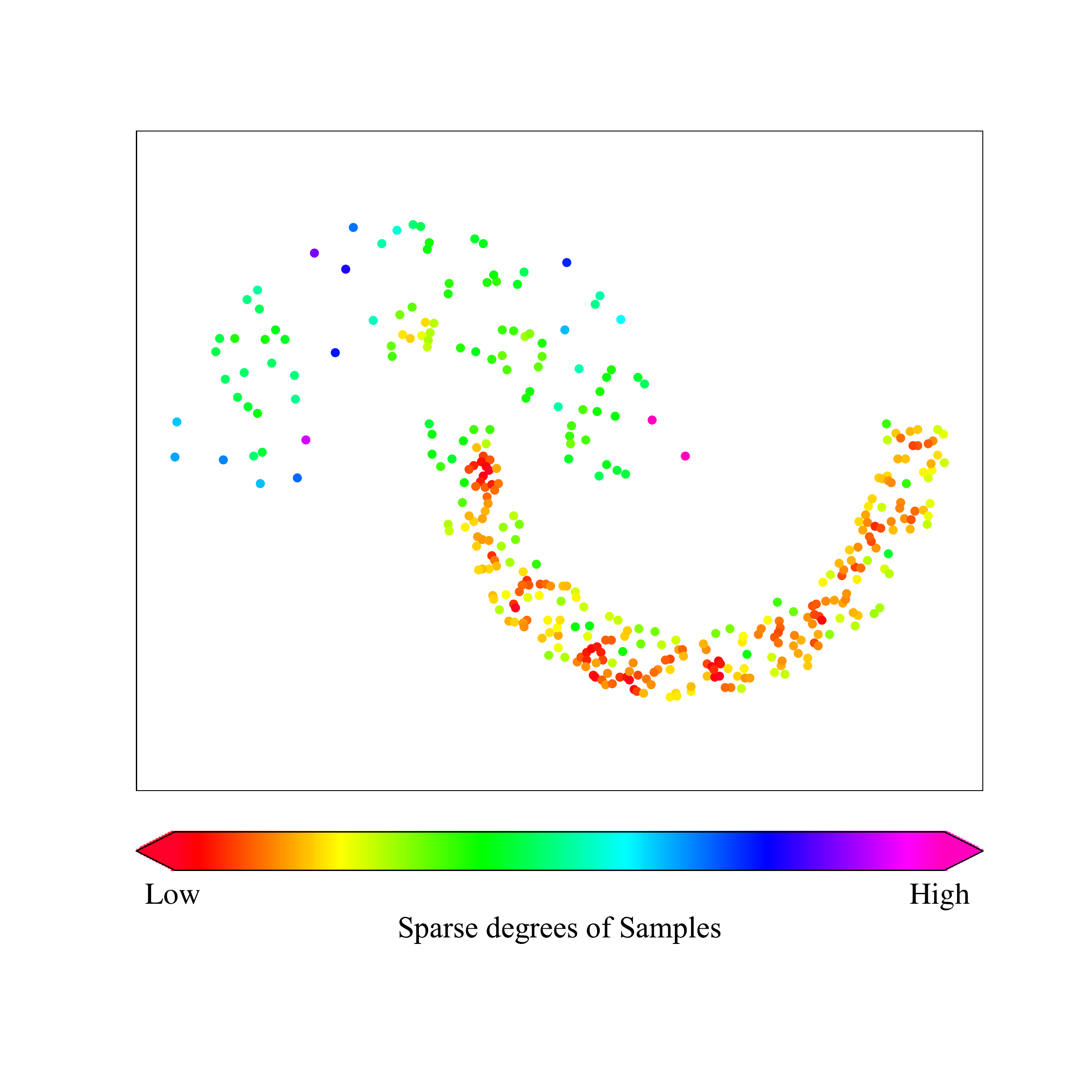}
		\label{fig: jain-1sparse_degree}
	}
	\hskip -30pt
	\subfigure[]{
		\includegraphics[width=8cm]{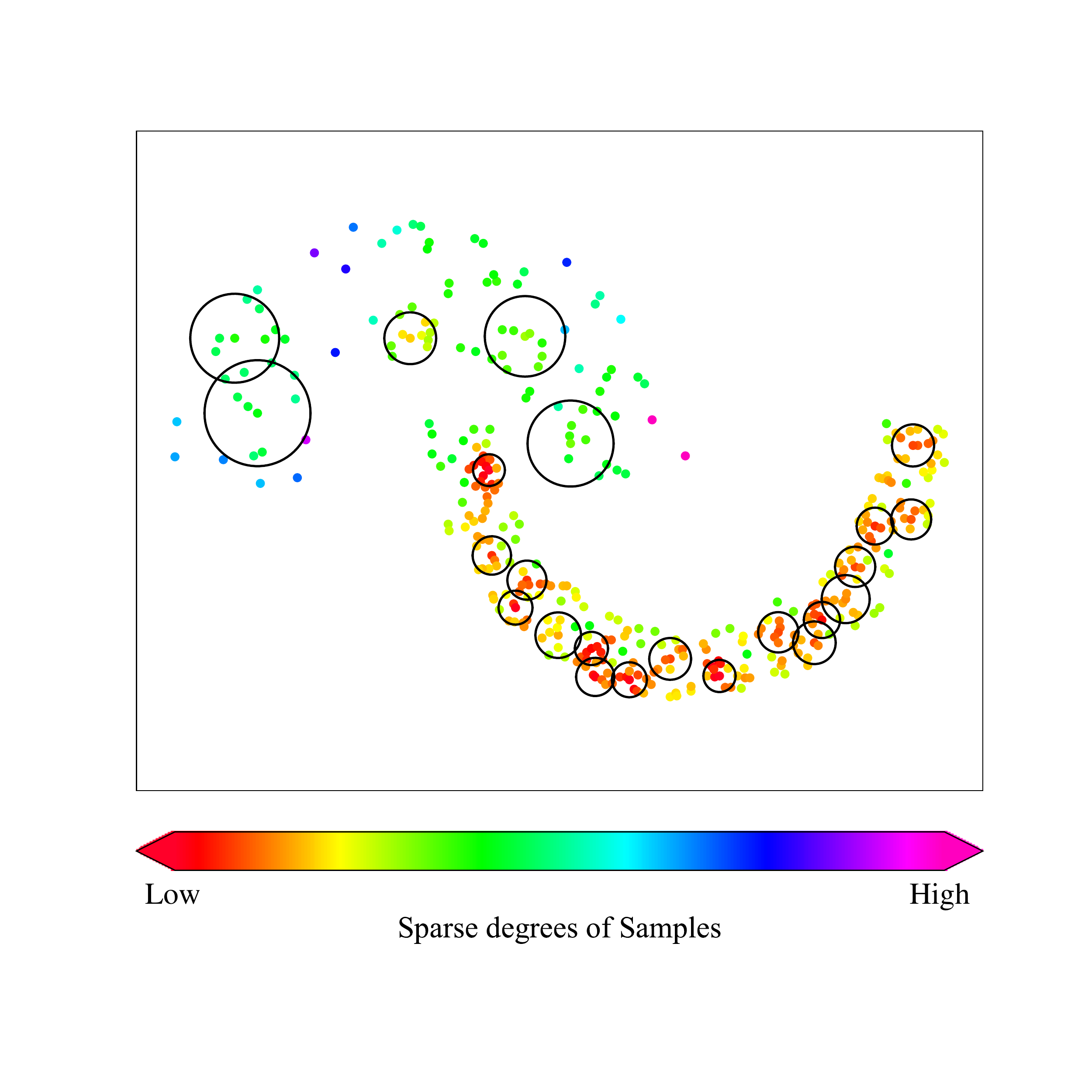}
		\label{fig: jain-2granulation}
	}
	\vskip -35pt  
	\subfigure[]{
		\includegraphics[width=8cm]{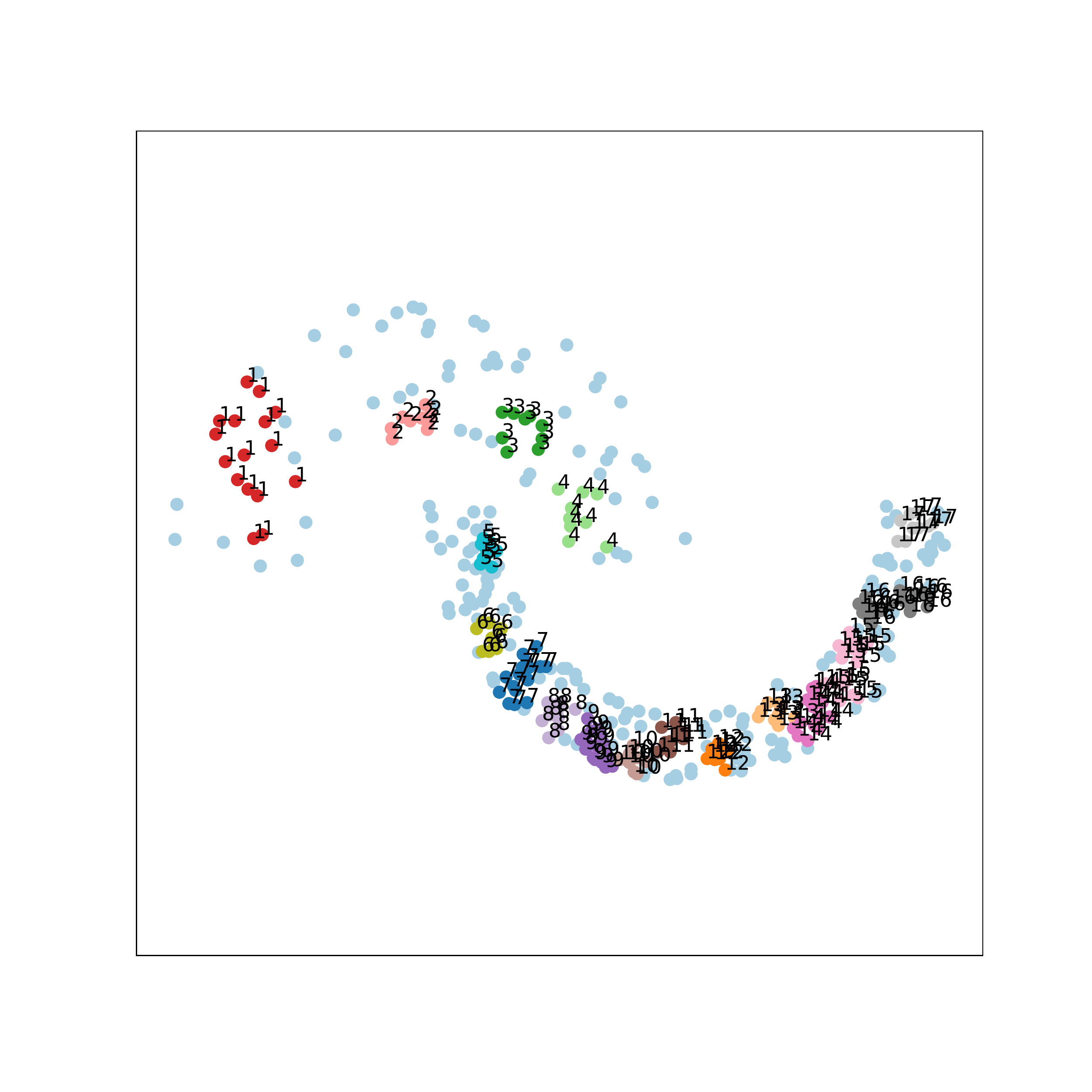}
		\label{fig: jain-3com1}
	}
	\hskip -30pt
	\subfigure[]{
		\includegraphics[width=8cm]{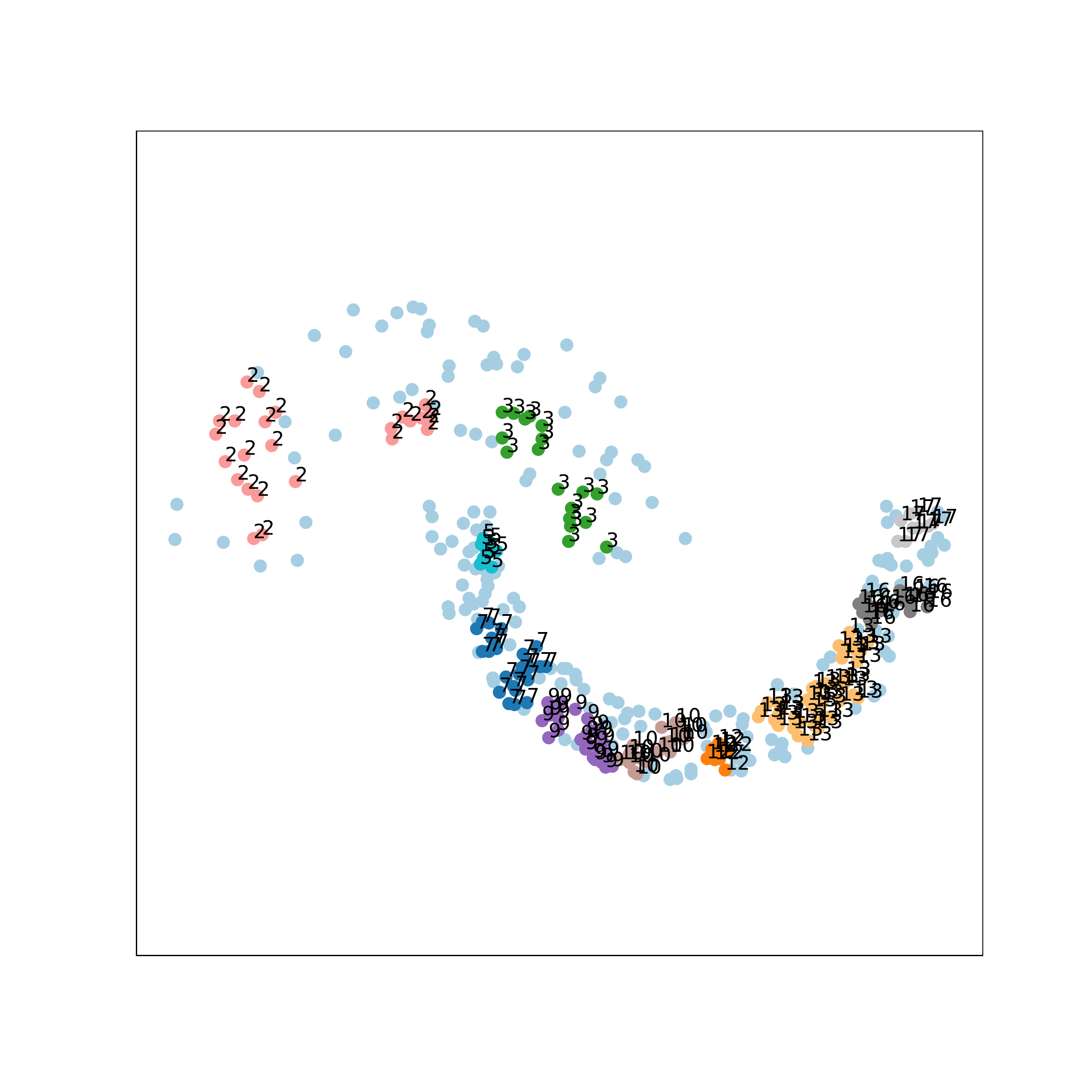}
		\label{fig: jain-4com2}
	}
	\vskip -35pt  
	\subfigure[]{
		\includegraphics[width=8cm]{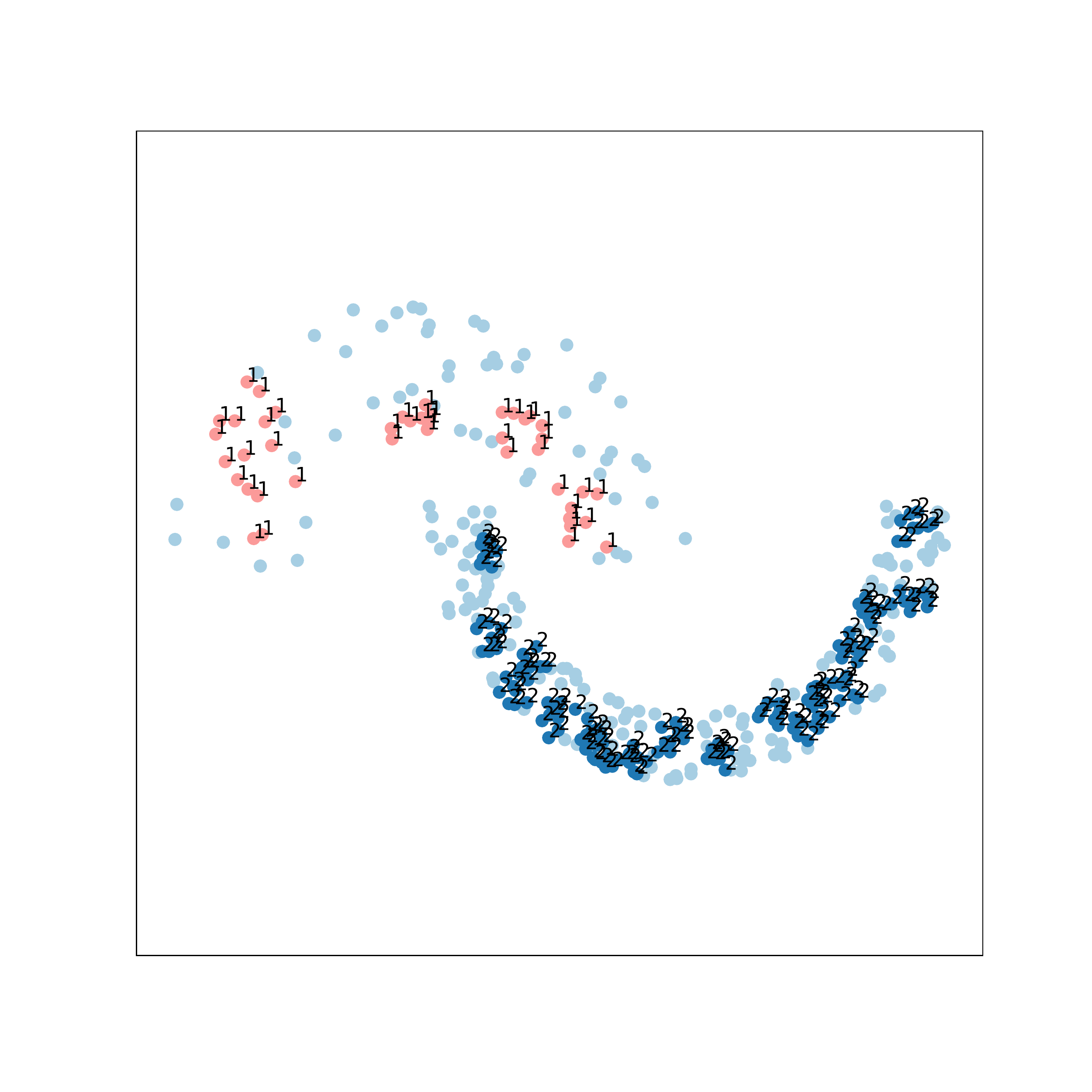}
		\label{fig: jain-5com3}
	}
	\hskip -30pt
	\subfigure[]{
		\includegraphics[width=8cm]{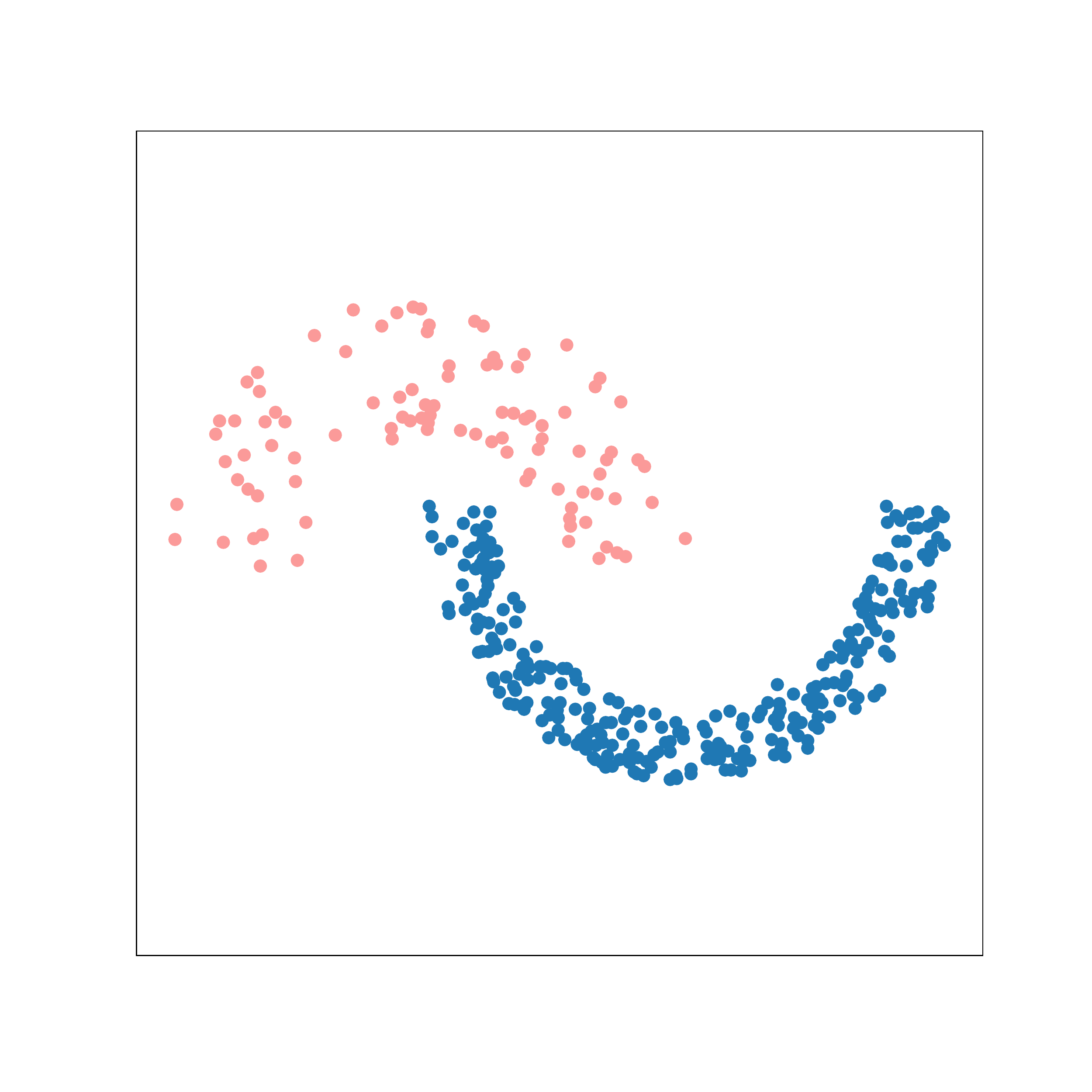}
		\label{fig: jain-6final}
	}
	\caption[process.]{The clustering process of the Jain dataset: (a) Sparse degrees of samples; (b) Granulation of samples; (c) Fusion based on intersection relationship; (d)Fusion based on density transmission; (e) Fusion based on distance; (f) Final clustering result.}
	\label{fig: jain-process}
\end{figure}

With the above granulation method, a group of granules $\{G_1,G_2,\dots,G_g\}$ with stable internal structure and clear density gradient can be obtained. For instance, the result of calculating sparse degrees of samples and granulation on the Jain dataset are shown in \autoref{fig: jain-1sparse_degree} and \autoref{fig: jain-2granulation}. The Jain dataset consists of two bowl-shaped clusters and there is a significant difference in the density of the two clusters. Observe \autoref{fig: jain-1sparse_degree} first, the color of a sample indicates its sparse degree. A sample with warm colors such as red, orange, yellow, etc., means that it is a high-density sample with a low sparse degree. And a sample with cold colors such as pink, purple, blue, etc., means that it is a low-density sample with a high sparse degree. After granulating the samples according to their sparse degrees, the black circles in \autoref{fig: jain-2granulation} represent granules. It can be seen from the figure that the granules are generated not only in the high-density regions but also in the low-density regions, that is, samples with clear density gradients in the low-density regions can form granules too. Consequently, it indicates that the granulation method coupled with the sparse degrees of samples can well deal with the kind of datasets with large density differences among clusters. Also, this method can avoid a situation where excessive attention is paid to high-density clusters at the expense of low-density clusters.

In addition, from the definition of a granule, it is known that each center of granules is a density peak sample within the granules, and the density peak sample is only for a small region, reflecting the density of a small region. In contrast, DPC determines the density peak samples for a whole dataset, so its density peak samples will affect the assignment of all other samples and thus influence the final result of clustering directly. As a result, the risk of error propagation and the cost of incorrectly selecting density peak samples are greatly increased. Unlike DPC, the density peak samples in our granulation method do not affect the structure of the final clusters directly and individually, which can weaken the force of the density peak samples and be more favorable to the generation of a better clustering result.

\subsection{Initial clusters obtained by granule fusion\label{subsection: Initial clusters obtained by combining}}
In the following, granules are combined according to the relationships among them, which leads to other forms of stable granule-related structures and initial clusters. Three fusion strategies for granules, based on intersection relationship, density transformation and distance respectively, are performed sequentially, which are beneficial to handling datasets with clusters of various shapes. These three fusion strategies are carried out sequentially, and possibly in a loop if special circumstances are encountered. The details of fusion strategies are described below.

\textbf{\uppercase\expandafter{\romannumeral1}. Fusion based on intersection relationship.}

It is taken for granted that the samples within two granules are very similar if the two granules intersect. So, it is natural that combine the granules based on their intersection relationship first.

Assume that any two granules are intersecting if they satisfy the condition: $|G_{i} \cap G_{j}| \geqslant 1$. On the basis of the intersection relationship of granules, all pairs of intersecting granules are combined and a set of \textbf{granule-clusters} (GCs) $\{GC_{1}, GC_{2}, \ldots, GC_{g^{*}}\}$ is generated, where $g^{*}$ indicates the number of granule-clusters.

\textbf{\uppercase\expandafter{\romannumeral2}. Fusion based on density transmission.}

In order to combine granule-clusters based on their characteristics of density, the idea of density transmission in the assignment strategy of DPC is improved and incorporated into the fusion strategy. This fusion strategy based on density transmission can make the shape of data distribution has little effect on the clustering results by transmitting low-density granule-clusters to high-density granule-clusters.

Since the idea of density transmission is derived, the definitions concerning distance and density are essential. Hence, the distance between granule-clusters is firstly defined as follows:
\begin{equation}
d^{*}(GC_{a},GC_{b})=\min\left\{d_{ij} \mid x_{i}\in GC_{a}, x_{j} \in GC_{b}\right\}, \label{distance between granule-clusters}
\end{equation}
where $a,b=1,\ldots,g^{*}$ and $a \neq b$. Namely, the distance between the nearest pair of samples located in the different granule-clusters is taken as the distance between the two granule-clusters.

In order to measure the density of granule-clusters, the sparse degrees of  samples are used again to define the sparse degrees of granule-clusters:
\begin{equation}
sd^{*}(GC_{a})=\frac{1}{\left|GC_{a}\right|} \sum_{x_{i} \in GC_{a}} sd(x_i), \label{sparse degree of granule-cluster}
\end{equation}
which indicates that the sparse degree of a granule-cluster is expressed by the average sparse degree of samples.

According to the improved idea of density transmission, given two granule clusters, if one granule cluster satisfies the requirements of being the nearest neighbor of the other and having a lower sparse degree than the other, then they will be combined. Exactly, granule-cluster $GC_a$ can be combined into granule-cluster $GC_b$, if they satisfy two requirements:\\
(1) $d^{*}(GC_{a},GC_{b})=\min \{d^{*}(GC_{a},GC_{y}) \mid GC_{y} \in \{ GC_{1}, GC_{2}, \ldots, GC_{g^{*}} \} \land GC_{y} \neq GC_{a}\}$,\\
(2) $sd^{*}(GC_{a}) \geq sd^{*}(GC_{b})$. 

In other words, the closest granule-cluster $GC_b$ to granule-cluster $GC_a$ is found at first, and then if the sparse degree of $GC_b$ is lower than or equal to that of $GC_a$, $GC_a$ combine into $GC_b$.

The above fusion strategy combines granule-clusters with a high sparse degree into granule-clusters with a low sparse degree, resulting in the transmission of low density to high density. Such transmission is favorable to address the problem of detecting clusters with arbitrary shapes. Based on density transmission, a set of granule-clusters (GCs) $\{GC_1,GC_2,\dots,GC_{g^{*}}\}$ are transformed into a set of \textbf{granule-flocks} (GFs) $\{GF_{1}, GF_{2}, \ldots, GF_{g^{\prime}}\}$, where $g^{\prime}$ indicates the number of granule-flocks. 

\textbf{\uppercase\expandafter{\romannumeral3}. Fusion based on distance.}

Like hierarchical agglomerative clustering, a set of granule-clusters or granule-flocks represents the initial clusters of a dataset at a certain granularity. For the purpose of being able to match practical requirements, given the number of clusters $c$ ($c<n$), a set of initial clusters at a specified granularity can be formed by utilizing the following fusion strategy.

Unlike the previous two fusion strategies, the fusion strategy base on distance is dynamic. Namely, during each iteration, the operation of combining a pair of granule-flocks will impact the operation in the next iteration. Details of the fusion strategy based on distance are as follows.

Similar to granule-clusters, the distance between two granule-flocks is defined as the distance between the pair of nearest samples located in the different granule-flocks:
\begin{equation}
d^{*}(GF_{a},GF_{b})=\min \left\{d_{ij} \mid x_{i} \in GF_{a}, x_{j} \in GF_{b}\right\}, \label{distance between granule-flocks}
\end{equation}
where $a,b=1,\ldots,g^{\prime}$ and $a \neq b$.

Given the number of clusters $c$, three relationships among $c$, the number of granule-clusters $g^{*}$ and the number of granule-flocks $g^{\prime}$ are considered: $c=g^{\prime}$ (or $c=g^{*}$), $c<g^{\prime}$ and $c>g^{\prime}$.
\begin{itemize}
	\setlength{\itemsep}{-1mm}
	\item For the case of $c=g^{\prime}$ (or $c=g^{*}$), the GFs (or the GCs) are taken directly as the initial clusters, and then the remaining samples are assigned using the assignment method proposed in \autoref{subsection: Unstable sample's evidential assignment} to obtain a final clustering result. 
	\item In the case of $c<g^{\prime}$, the fusion strategy based on distance is applied: a pair of GFs with the smallest distance is combined in each iteration, and the number of GFs and the distance between GFs are updated after each iteration until the current number of GFs to $c$. At last, the initial clusters $\{Cl_1,Cl_2,\ldots,Cl_c\}$ is produced. 
	\item To address the case of $c>g^{\prime}$, each GF is considered as a new small sub-dataset, and then repeat the processes including calculating the sparse degrees of samples, granulation, fusion based on intersection relationship and fusion based on density transmission on all sub-datasets respectively. In particular, following the fusion based on density transmission, every single sample without any new GF in a sub-dataset is considered as a new GF. After that, the new GFs generated from all sub-datasets are put together to form a new set of GFs. By judging the relationships between the number of the new GFs and $c$, the corresponding solution is selected according to the above discussion. What should be noted is that the iterations in this fusion strategy can definitely end up eventually, as described in the following remark.
\end{itemize}
	
\begin{remark}
	For the case of $c>g^{\prime}$, the iterations will be definitely stopped because sub-datasets become smaller during each round of iteration, i.e., the number of samples within each GF becomes smaller, and the limit case is that each GF contains only one sample. At this time, $g^{\prime}$ is equal to the number of samples within the granules and, in general, it will be greater than $c$. However, if a more extreme case occurs, where the dataset generates only one granule with $k$ samples, i.e., $g^{\prime}=k$, as well as $k<c$, the iterations will not be stopped. To cope with such a case, it is preferable to take $k$ as equal to $c$ in advance. Therefore, for a given $c$, the value of $k$ in the proposed algorithm is taken to be $\max\{c, \lceil log_{2}(n) \rceil\}$ to avoid getting into a dead loop. Finally, the iterations must be stopped and the initial clusters $\{Cl_1,Cl_2,\ldots,Cl_c\}$ can be achieved.
\end{remark}

Illustrative graphs of fusion based on intersection relationship, density transmission and distance on the Jain dataset are exemplified in \autoref{fig: jain-3com1} to \autoref{fig: jain-5com3}. Above all, following the fusion strategy based on intersection relationship, the intersecting granules are combined and 17 $GC$s are formed. As shown in \autoref{fig: jain-3com1}, different colors and serial numbers are used to distinguish the different GCs where samples are located. Secondly, observing \autoref{fig: jain-3com1} and \autoref{fig: jain-4com2}, 17 GCs are combined into 10 GFs by the fusion strategy based on density transmission. For a brief analysis, since the closest granule-cluster of $GC_1$ is $GC_2$ and $sd^{*}(GC_2)$ is lower than $sd^{*}(GC_1)$, the low-density $GC_1$ can be combined into the high-density $GC_2$; however, since the closest granule-clusters of $GC_{12}$ is $GC_{11}$ but $sd^{*}(GC_{11})$ is higher than $sd^{*}(GC_{12})$, the high-density $GC_{12}$ cannot be combined into low-density $GC_{11}$. In the end, according to the case where the number of GFs is greater than the true number of clusters, 10 GFs are combined into 2 initial clusters using the fusion strategy based on distance and the result is displayed in \autoref{fig: jain-5com3}. The samples marked with the number 1 indicate that they belong to the initial cluster $Cl_1$ and the samples marked with the number 2 indicate that they belong to the initial cluster $Cl_2$. The marked samples are stable samples, while the others are unstable samples. From the above, it can be seen that GFDC can generate reasonable initial clusters and detect clusters with arbitrary shapes.

\subsection{Evidential assignment of unstable samples \label{subsection: Unstable sample's evidential assignment}}

After the initial clusters obtained, samples within the initial clusters are defined as stable samples which are with definite label information, and samples outside the initial clusters are defined as unstable samples (assume there are $t$ unstable samples) which need to be further assigned. In order to assign these unstable samples as well as identify outliers, the EK-NN rule \cite{EK-NN_rule}  based on the Dempster-Shafer theory is used to design a method for evidential assignment of unstable samples and create a credal partition of $U$.

Firstly, assume that the frame of discernment is $\Omega=\{Cl_{1}, Cl_{2}, \ldots, Cl_{c}\}$ and for all stable samples in the initial clusters, i.e., $x_{i} \in Cl_u$, $u=1,2,\ldots,c$, the mass function $m_{i}^{\Omega}$ is defined as:

\begin{subequations}
\begin{flalign}
	&\label{initialize credal partition1}
	m_{i}^{\Omega}(Cl_{u}) =
	\begin{cases}
		1, & \text { if } x_{i} \in Cl_{u}, \\
		0, & \text { if } x_{i} \notin Cl_{u}, 
	\end{cases}\\
	&\label{initialize credal partition2}
	m_{i}^{\Omega}(A) = 0,  \\ 
	&\label{initialize credal partition3}
	m_{i}^{\Omega}(\Omega) = 1-\sum_{u}m_{i}^{\Omega}(Cl_{u})-\sum_{A}m_{i}^{\Omega}(A) = 0, 
\end{flalign} \notag
\end{subequations}
where $A \in 2^{\Omega} \backslash \{Cl_{1}, Cl_{2}, \ldots, Cl_{c}, \Omega\}$. The value of $m_{i}^{\Omega}(\cdot)$ represents the degree of belief that stable sample $x_{i}$ belong to a single-cluster $Cl_u$, a meta-cluster A or $\Omega$. It is worth noting that, according to the EK-NN rule and the purpose of this paper, the degree of belief that a sample belongs to a meta-cluster is not considered. Therefore, there is $m^{\Omega}(A)=0$ for all samples.

Secondly, a stable set $S$ with all stable samples within the initial clusters is initialized. Next, sample $x_i$ with the lowest sparse degree from the unstable samples outside $S$ is considered. Then, pieces of evidence are calculated, which regard the label of $x_i$ provided by $k$ nearest neighbors of $x_i$ within $S$. So, for an unstable sample $x_i$ with the lowest sparse degree outside $S$, the corresponding mass function is defined as follows:
\begin{equation}
\begin{cases}
m_{ij}^{\Omega}(Cl_{u})=\exp (-(d_{ij}+sd(x_j))) \cdot m_{j}^{\Omega}(Cl_{u}), \\
m_{ij}^{\Omega}(\Omega)=1-\sum_{u} m_{ij}^{\Omega}(Cl_{u}),
\end{cases}  \label{calculate masses}
\end{equation}
for all $x_{j} \in N_{k}^{S}(x_{i})$, where $ N_{k}^{S}(x_{i})$ denotes the set comprising $k$ nearest neighbors of sample $x_i$ in $S$. Here it is easy to know that $k$ is less than or equal to $|S|$ because there are at least $k$ samples in a set of granules according to the definition of the granule. The $m_{ij}^{\Omega}(Cl_{u})$ denotes a piece of evidence that $x_i$ belongs to $Cl_u$ provided by $x_j$, and $m_{ij}^{\Omega}(\Omega)$ denotes a piece of evidence that $x_i$ is an outliers provided by $x_j$. Unlike the EK-NN rule and its improved versions, when designing the function for unstable samples, not only the distance factor is taken into account but also the density factor by adding the sparse degrees of the nearest neighbors to the mass function.

From \autoref{calculate masses}, it can be seen that if $x_j$ is far from $x_i$ and $x_j$ is a sparse sample, i.e., the distance between $x_i$ and $x_j$ is long and the sparse degree of $x_j$ is high, the label of $x_j$ will provide very little information regarding the label of $x_i$. In contrast, if $x_j$ is close to $x_i$ and $x_j$ is a dense sample, i.e., the distance between $x_i$ and neighbor $x_j$ is short and the sparse degree of $x_j$ is low, there will be a large degree of belief that $x_i$ and $x_j$ belong to the same cluster.

Thirdly, with the Dempster's rule of combination (\autoref{Dempster's rule of combination}), these $k$ pieces of evidence can be combined depending on
\begin{equation}
m_{i}^{\Omega}(\cdot)=\underset{x_{j} \in N_{k}^{S}\left(x_{i}\right)}{\bigoplus} m_{ij}^{\Omega}(\cdot). \label{combine masses}
\end{equation}
After all basic belief masses of sample $x_i$ are calculated, add the processed sample $x_i$ to $S$ and consider the next unstable sample with the lowest sparse degree outside $S$ to repeat the above process until $S=U$.

Finally, after obtaining the credal partition of $U$, labels can be assigned to samples based on 
\begin{equation}
x_{i} \in \underset{Cl_{u}}{\operatorname{arg~max}}~m_{i}^{\Omega}(Cl_{u}). \label{assign}
\end{equation}
This means that a sample is assigned to the cluster which makes the value of mass largest. Particularly, when it is desired to identify outliers, a threshold $\tau$ can be set such that if $m_i^{\Omega}(\Omega)>\tau$, sample $x_i$ is considered as an outlier. Generally, $\tau$ can take the value from 0.99 to 1.

The result of the evidential assignment of the unstable samples for the Jain dataset is shown in \autoref{fig: jain-6final}, which is without a threshold for outliers. An example of evidential assignment of unstable samples with a threshold for outliers is illustrated in \autoref{subsection: Illustrative examples}.

\subsection{Analysis of algorithm\label{subsection: Analysis of algorithm}}

According to the exposition in \autoref{subsection: Sample's sparse degree} to \autoref{subsection: Unstable sample's evidential assignment}, GFDC is summarized as \autoref{algorithm: Combine granules into initial clusters} and \autoref{algorithm: main algorithm}. The details of the proposed algorithm for combining granules into initial clusters are shown in the former, where the application of three fusion strategies for generating stable granule-related structures and initial clusters is carefully elaborated. The latter describes the complete algorithm.

In the following, the time complexity of GFDC is analyzed in terms of \autoref{algorithm: main algorithm} as follows:
\begin{itemize}
	\setlength{\itemsep}{-1mm}  
	\setlength{\topsep}{-1mm}  
	\item Computing the sparse degree in step 1: $O(wn(n-1))+O(wn^2)+O(n(n-1))+O(n^2)$, where $w$ denotes the number of attributes, and $n$ denotes the number of samples in a dataset. It includes calculating the matrix of distance, calculating the relative density, finding the maximum and sorting.
	\item The process of fusion in step 3: $O(g^*(g^*-1)/2 \cdot (n^2/(g^*)^2-1))+O(g^*)+O(g^*-1)+O((g^{\prime}-c) \cdot g^{\prime}(g^{\prime}-1)/2 \cdot (n^2/(g^{\prime})^2-1))$, where $g^{*}$ denotes the number of granule-clusters, $g^{\prime}$ denotes the number of granule-flocks and $c$ is the number of clusters. It includes calculating the distance between each pair of granule-clusters, calculating the sparse degree of each granule-cluster, finding the minimum, and calculating the distance between each pair of granule-flocks in iteration. Moreover, it should be noted that it is impossible to predict the time complexity for the case of $g^{\prime} < c$.
	\item Creating the credal partition of the dataset and determining the labels in step 4 to step 11: $O(tn^2/2)+O(t \cdot (log_2(n))^2 \cdot (c+1))+O(tc)$, where $t$ represents the number of samples outside initial clusters. It includes calculating the masses and combining them, as well as finding the maximum.
\end{itemize}

In summary, the total time complexity of GFDC is approximately $O((w+g^{\prime}-c+t)n^2)$.

\vspace{0.5cm}
\begin{spacing}{1.2}
\begin{algorithm}[H]\footnotesize
	\caption{Initial clusters obtained by granule fusion.}
	\label{algorithm: Combine granules into initial clusters}
	\LinesNumbered
	\KwIn{The granules $\{G_1,G_2,\ldots,G_g\}$ and the number of clusters $c$.}
	\KwOut{The initial clusters $\{Cl_1,Cl_2,\ldots,Cl_c\}$}
	
	\For{each granule $G_i$}{
		\For{each granule $G_j$ and $G_i \ne G_j$}{
			\If{$|G_{i} \cap G_{j}| \geqslant 1$}{
				Combine $G_i$ and $G_j$;}}}
	
	Renumber and get a set of granule-clusters $\{GC_{1}, GC_{2}, \ldots, GC_{g^{*}}\}$;
	
	\uIf{$g^{*}=c$}{
		\Return{The initial clusters $\{Cl_1,Cl_2,\ldots,Cl_c\} \gets \{GC_{1}, GC_{2}, \ldots, GC_{g^{*}}\}$};}
	\Else{Compute distance $d^*(GC_a,GC_b)$ between each pair of GCs according to \autoref{distance between granule-clusters};\\
		Compute sparse degree $sd^*(GC_a)$ of each GC according to \autoref{sparse degree of granule-cluster};\\
		\For{each granule-cluster $GC_a$}{
			Find the nearest granule-cluster $GC_b$ of $GC_a$;\\
			\If{$sd^*(GC_a) \geq sd^*(GC_b)$}{
				Combine $GC_a$ into $GC_b$;}}
		
		Renumber and get a set of granule-flocks $\{GF_{1}, GF_{2}, \ldots, GF_{g^{\prime}}\}$;
		
		\uIf{$g^{\prime} = c$}{
			\Return{The initial clusters $\{Cl_1,Cl_2,\ldots,Cl_c\} \gets \{GF_{1}, GF_{2}, \ldots, GF_{g^{\prime}}\}$};}
		\uElseIf{$g^{\prime} > c$}{
			\While{$g^{\prime} \neq c$}{
				Compute distance $d^*(GF_a,GF_b)$ between each pair of GFs according to \autoref{distance between granule-flocks};\\
				Find the pair of GFs which has minimal distance and combine them;\\
				Update the GFs and $g^{\prime} \gets g^{\prime}-1$;}
			\Return{The initial clusters $\{Cl_1,Cl_2,\ldots,Cl_c\} \gets \{GF_{1}, GF_{2}, \ldots, GF_{g^{\prime}}\}$};}
		\Else{
			Consider each GF as a new small sub-dataset;\\
			\For{each sub-dataset}{
				Compute sparse degrees of samples and generate granules via steps 1 and 2 from \autoref{algorithm: main algorithm};\\
				Generate GCs and GFs via step 1 to step 20;}
			Aggregate all GFs generated from all sub-datasets into a new set of granule-flocks; \\
			Execute step 21 to step 38 based on the new set of granule-flocks.}
	}

\end{algorithm}
\end{spacing}

\begin{spacing}{1.2}
	\begin{algorithm}[H]\footnotesize
		\caption{GFDC: A granule fusion density-based clustering with evidential reasoning.}
		\label{algorithm: main algorithm}
		\LinesNumbered
		\KwIn{Dataset $U$, the number of cluster $c$ and the threshold $\tau$ (if necessary).}
		\KwOut{A clustering result and outliers (if threshold $\tau$ is given).}
		Compute sparse degree $sd(x_i)$ of each sample according to \autoref{sparse degree};
		
		Generate the granules $\{G_{1}, G_{2}, \ldots, G_{g}\}$ based on sparse degrees of samples;
		
		Combine granules into initial clusters $\{Cl_{1}, Cl_{2}, \ldots, Cl_{c}\}$ via \autoref{algorithm: Combine granules into initial clusters};
		
		Constitute a set $S$ with samples in the initial clusters and compute the masses $m_{\cdot}^{\Omega}(Cl_u)$ of stable samples according to \autoref{initialize credal partition1} to \autoref{initialize credal partition3};
		
		\While{$|S| \ne |U|$}{
			Consider the unassigned sample $x_{i}$ with lowest sparse degree in $U \backslash S$, find its $k$ nearest neighbours in $S$ and compute the masses $m_{ij}^{\Omega}(Cl_u)$ and $m_{ij}^{\Omega}(\Omega)$ according to \autoref{calculate masses};\\
			Combine the masses $m_{i}^{\Omega}(\cdot)$ according to \autoref{combine masses};\\
			$S \gets S \cup x_{i}$;}
		Assign each sample to the corresponding cluster according to \autoref{assign} or treat it as an outlier.
		
		\Return{A clustering result and outliers (if threshold $\tau$ is given).}
	\end{algorithm}
\end{spacing}

\section{Experimental results\label{section: Experimental results}}
In this section, to verify the validity of the proposed algorithm, some experiments are conducted on datasets, including 20 synthetic datasets\footnote{Synthetic datasets derived from the website: http://cs.joensuu.fi/sipu/datasets/.} (as listed in \autoref{table: Synthetic datasets}) and 9 real-world datasets\footnote{Real-world datasets derived from the website: http://archive.ics.uci.edu/ml/datasets.php.} (as listed in \autoref{table: Real-world datasets}), which are commonly used to test the performance of a clustering algorithm. To be more convincing, the performance of the proposed algorithm is compared with several classical clustering algorithms, including $k$-means++ \cite{k-means++}, SC \cite{SC2007}, DBSCAN \cite{DBSCAN}, DPC \cite{DPC}, DPC-KNN \cite{DPC-KNN} and ECM \cite{ECM}. Experiments are performed on a computer with Intel(R) Xeon(R) E5-2650 v4 @ 2.20GHz CPU with 64 GB RAM.
\vspace{0.5cm}

In the experiments, four effective evaluation measures are made use of to quantify the comparative results:
\begin{itemize}
	\setlength{\itemsep}{-1mm}
	\item \emph{Purity} \cite{purity}: Purity is equivalent to the accuracy of the clustering result. It is equal to the number of correctly assigned samples divided by the total number of samples. Since the true class corresponding to each cluster of the clustering result is unknown, the maximum value in each case is taken. It takes a value between 0 and 1, and the larger the purity, the better the clustering results.
	\item \emph{Adjusted Rank Index} (ARI) \cite{ARI}: ARI is proposed to solve the problem that \emph{Rank Index} (RI) cannot guarantee that its values are close to zero for random label assignments. Compared with RI, ARI has higher discrimination. ARI takes values in the range of $[-1,1]$ and larger values mean that clustering results match the real situations.
	\item \emph{Adjusted Mutual Information} (AMI) \cite{AMI}: \emph{Mutual Information} (MI) is used to evaluate the similarity between the real labels and the clustering labels, whose common form is \emph{Normalized Mutual Information} (NMI) and AMI. In particular, like ARI, AMI can identify the case of random labels. The value of AMI ranges from $[-1,1]$, and the larger the value, the better the clustering result matches the real situation.
	\item \emph{Fowlkes-Mallows scores} (FMI) \cite{FMI}: FMI is represented by the number of \emph{True Positive} (TP), the number of \emph{False Positive} (FP) and the number of \emph{False Negative} (FN), which means the geometric mean of pairwise precision and recall. FMI scores range from $[0,1]$ and a higher score indicates that the clustering result is more similar to the ground truth.
\end{itemize}

Furthermore, the parameters of compared algorithms in the experiments are set as follows.
\begin{itemize}
	\setlength{\itemsep}{-1mm}
	\item For $k$-means++, the number of clusters in the algorithm is set to the true number of clusters and the initial centers are given randomly. 
	\item In SC, the dimension of the projection subspace is set to the true number of clusters, $k$-means is used as the strategy for assigning labels in the embedding space, and parameter $gamma$ is set to the value that makes the clustering result optimal for each dataset. 
	\item For two parameters $\varepsilon$ and $minpts$ in DBSCAN, the grid search method is used to find the optimal parameters for each dataset, where $\varepsilon$ is searched in the range of 0.01 to the average distance of any pair of samples in the dataset, and $minpts$ is searched from $5$ to $10$, with a step size of $1$. 
	\item In DPC, there is a parameter cutoff distance $d_c$, and it can be defined as $d_c=D_{\lceil |D| \times p \rceil}$, where $D=\{D_1,D_2,\dots\}$ is a set of the distances between any pair of samples in the dataset and the distances are in ascending order. And $p$ is a user-defined parameter, which is found the best value from 0.01 to 0.90, with a step size of 0.01, for each dataset in our experiments. In addition, the first $C$ samples with the largest $\gamma$ are used as cluster centers automatically, where $\gamma=\delta \times \rho$, $\delta$ and $\rho$ indicate the distance and density of samples respectively, and $C$ is the true number of clusters. 
	\item In DPC-KNN, the number of nearest neighbors is computed as $p$ percent of the number of samples, and the parameter $p$ is selected form $[0.1\%,0.2\%,0.5\%,1\%,2\%,6\%]$. Also, the method of selecting cluster centers in DPC-KNN is the same as in DPC. 
	\item For ECM, the number of clusters is set to the true number of clusters and the initial centers are given randomly. Only the parameter $\alpha$ in ECM is searched from 1 to 3 with a step size of 0.5, parameters $\beta$ and $\delta$ take the default values of 2 and 10.
\end{itemize}

When adjusting the parameters of SC and DPC, a set of parameters that makes the value of ARI of a clustering result highest is considered as the optimal parameters and the corresponding clustering result is the optimal result of the algorithm. For DBSCAN, it should be noted that in the following experiments, those results, where the number of detected clusters is the same as or closest to the true number of clusters and where there is as little noise as possible, are chosen as the optimal results.

\subsection{Illustrative example\label{subsection: Illustrative examples}}
\begin{figure}[htbp]\footnotesize
    \centering
    \vspace{-25pt}  
    \setlength{\abovecaptionskip}{-10pt}  
    \subfigcapskip=-25pt  
    \subfigure[]{
    \includegraphics[width=8.7cm]{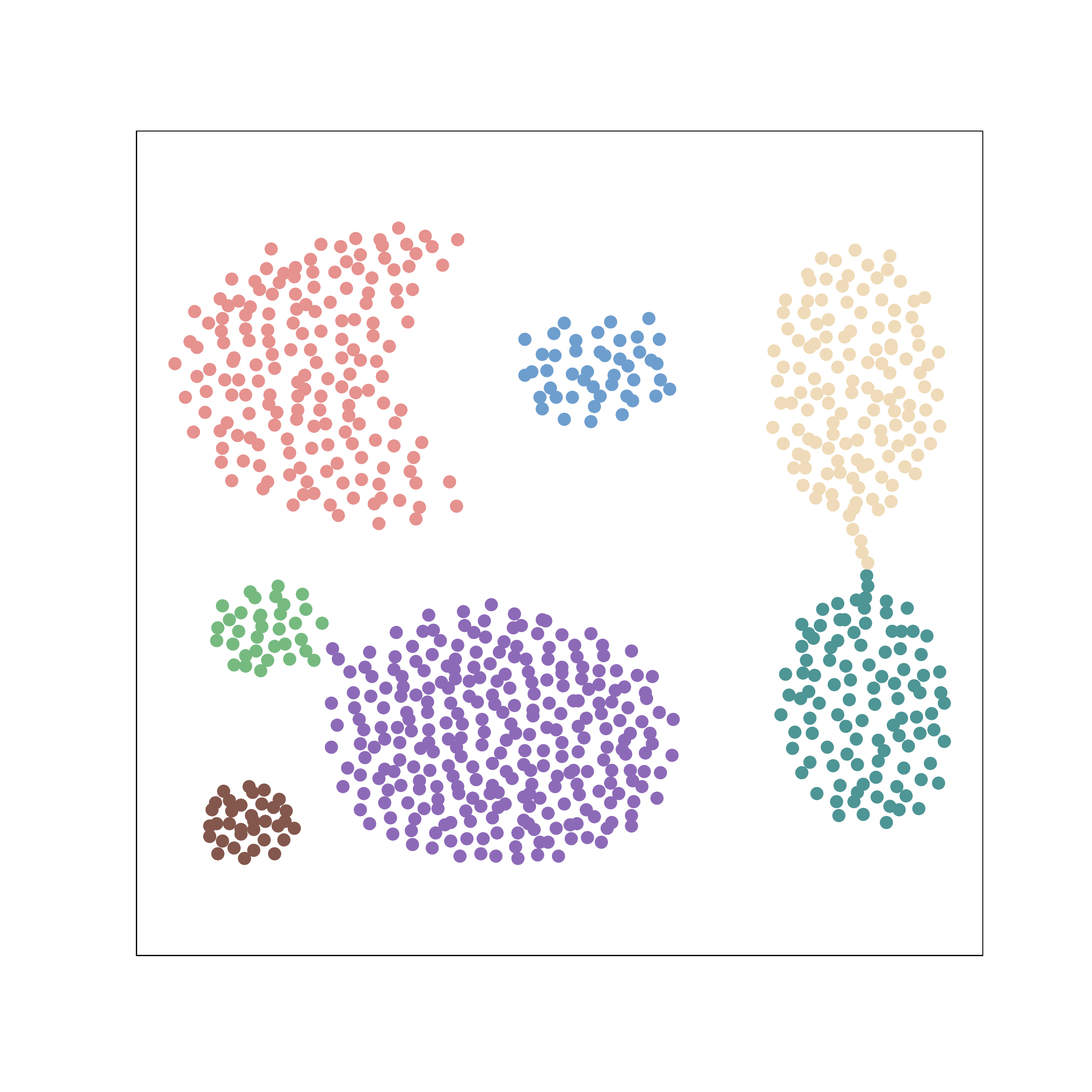}
    \label{fig: aggregation-groundtruth-outlier}
    }
    \hskip -30pt
    \subfigure[]{
    \includegraphics[width=8.7cm]{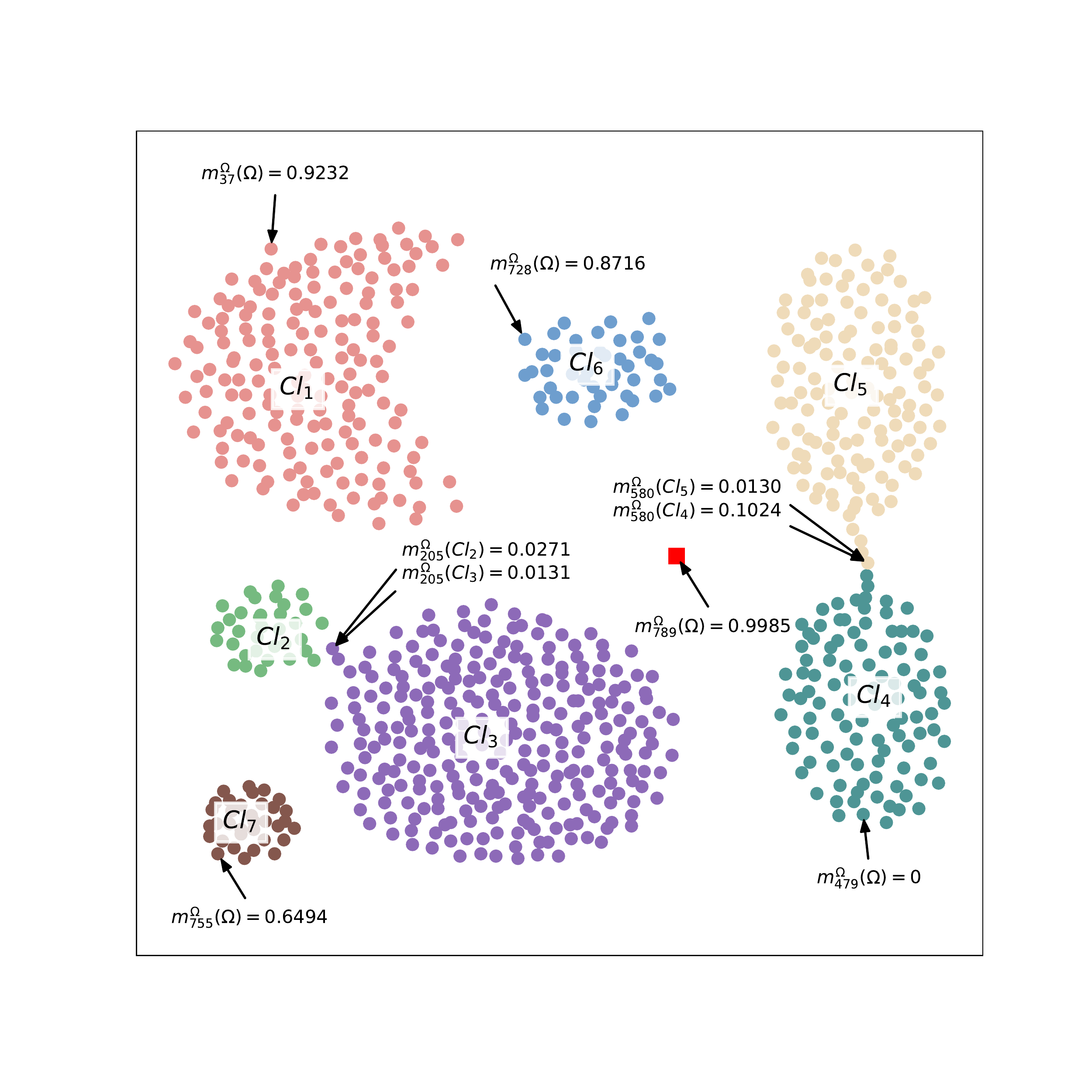}
    \label{fig: outlier}
    }
    \caption{Illustration of identifying outliers: (a) the ground truth of the Aggregation dataset; (b) the clustering result obtained by GFDC.}
    \label{fig: Illustration of identifying outliers}
\end{figure}

In this subsection, an illustrative example is used to demonstrate the ability of GFDC for clustering and identifying outliers. The Aggregation dataset is considered in this example, consisting of 788 samples and 7 clusters, and the ground truth is shown in \autoref{fig: aggregation-groundtruth-outlier}. Furthermore, an outlier is manually added to the dataset to test whether GFDC can identify it accurately.

Analyzing the clustering result without considering outliers first, in \autoref{table: ARI comparison on synthetic datasets}, the comparison results of  $ARI_{k-means++}=0.7579$, $ARI_{SC}=0.9949$, $ARI_{DBSCAN}=0.9065$, $ARI_{DPC}=0.9942$, $ARI_{DPC-KNN}=0.9978$, $ARI_{ECM}=0.6392$ as well as $ARI_{GFDC}=0.9949$ indicate that GFDC is good in this case. Here, in the result of the proposed method, samples 205 and 580 are misclassified samples. Sample 205 originally belongs to $Cl_3$, but due to $m_{205}^{\Omega}(Cl_2)=0.0271$ and $m_{205}^{\Omega}(Cl_3)=0.0131$, the sample is misclassified into $Cl_2$, and sample 580 originally belongs to $Cl_5$, but due to $m_{580}^{\Omega}(Cl_4)=0.1024$ and $m_{580}^{\Omega}(Cl_5)=0.0130$, the sample is misclassified into $Cl_4$. From \autoref{fig: Illustration of identifying outliers}, it can be seen that, for most clustering algorithms, samples like samples 205 and 508 which are at the junction of two clusters are very difficult to be classified correctly. In addition, the comparison of clustering results of other algorithms on the Aggregation dataset is shown in \autoref{fig: aggregation}.

Analyzing the ability of GFDC to identify outliers next, in the Aggregation dataset which includes an outlier and the outlier is marked with a red square in \autoref{fig: outlier}, it can be seen that, as $m_{789}^{\Omega}(\Omega)=0.9985$ is close to 1, sample 789 can be identified as an outlier. While considering the samples located at the boundaries of other clusters, there are $m_{37}^{\Omega}(\Omega)=0.9232$, $m_{479}^{\Omega}(\Omega)=0$, $m_{728}^{\Omega}(\Omega)=0.8716$ and $m_{755}^{\Omega}(\Omega)=0.6494$. Although some of the boundary samples have large values of $m^{\Omega}(\Omega)$, they are not very close to 1, so they are not identified as outliers. In addition, since sample 479 is the sample in the initial clusters, its value of $m^{\Omega}(\Omega)$ is 0. In sum, it is thus clear that GFDC can effectively identify outliers, and the threshold $\tau$ is generally taken in the range of $[0.99,1]$, which can be determined by the user according to the practice.

\subsection{Experiments on synthetic datasets and results analysis\label{subsection: Experiments on synthetic datasets}}

\begin{figure}[H]\footnotesize
	\centering
	\vspace{-25pt}  
	\setlength{\abovecaptionskip}{0pt}  
	\subfigcapskip=-15pt  
	\subfigure[Ground truth]{
		\includegraphics[width=4.5cm]{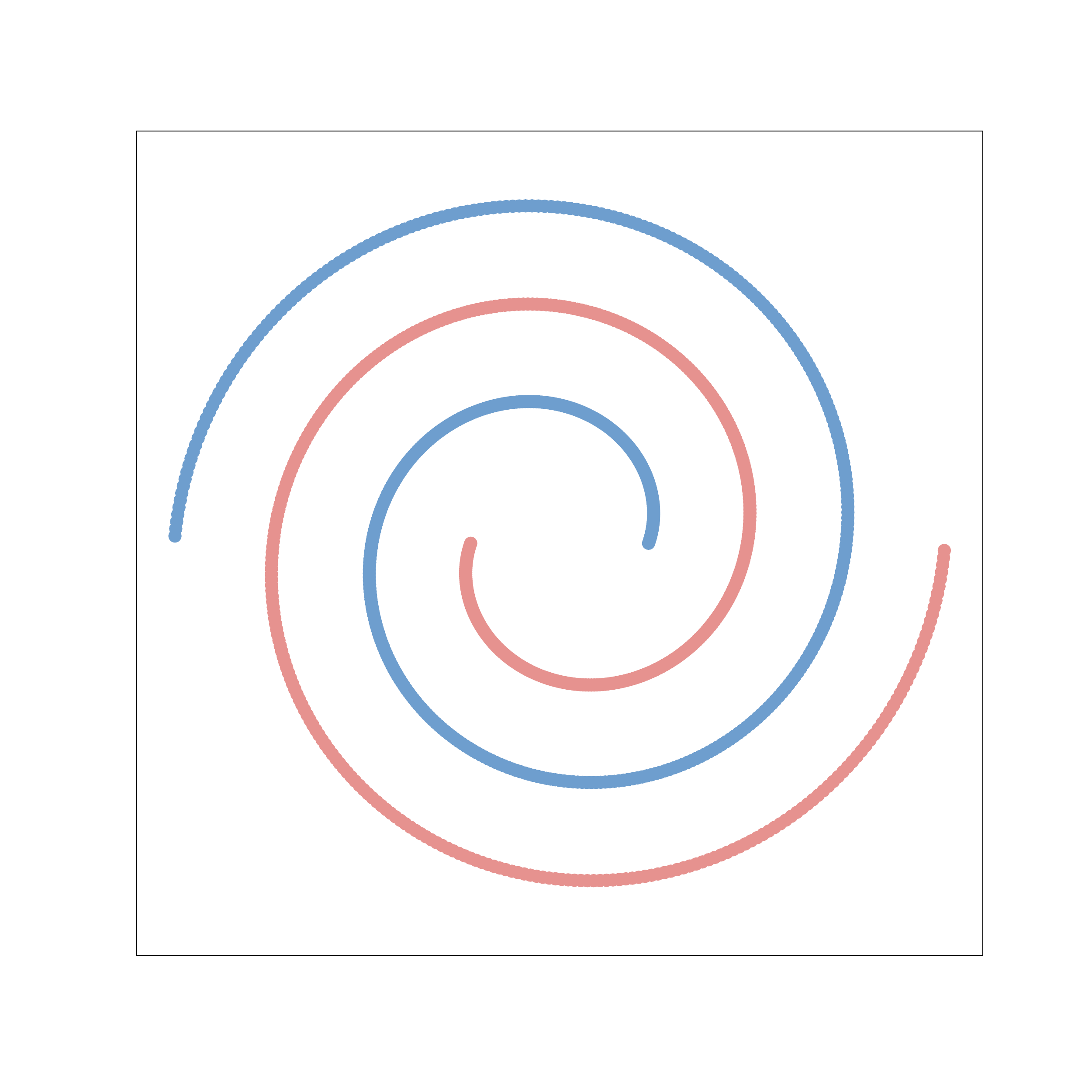}
		\label{fig: 2spiral ground truth}
	}
	\hskip -30pt  
	\subfigure[$k$-means++]{
		\includegraphics[width=4.5cm]{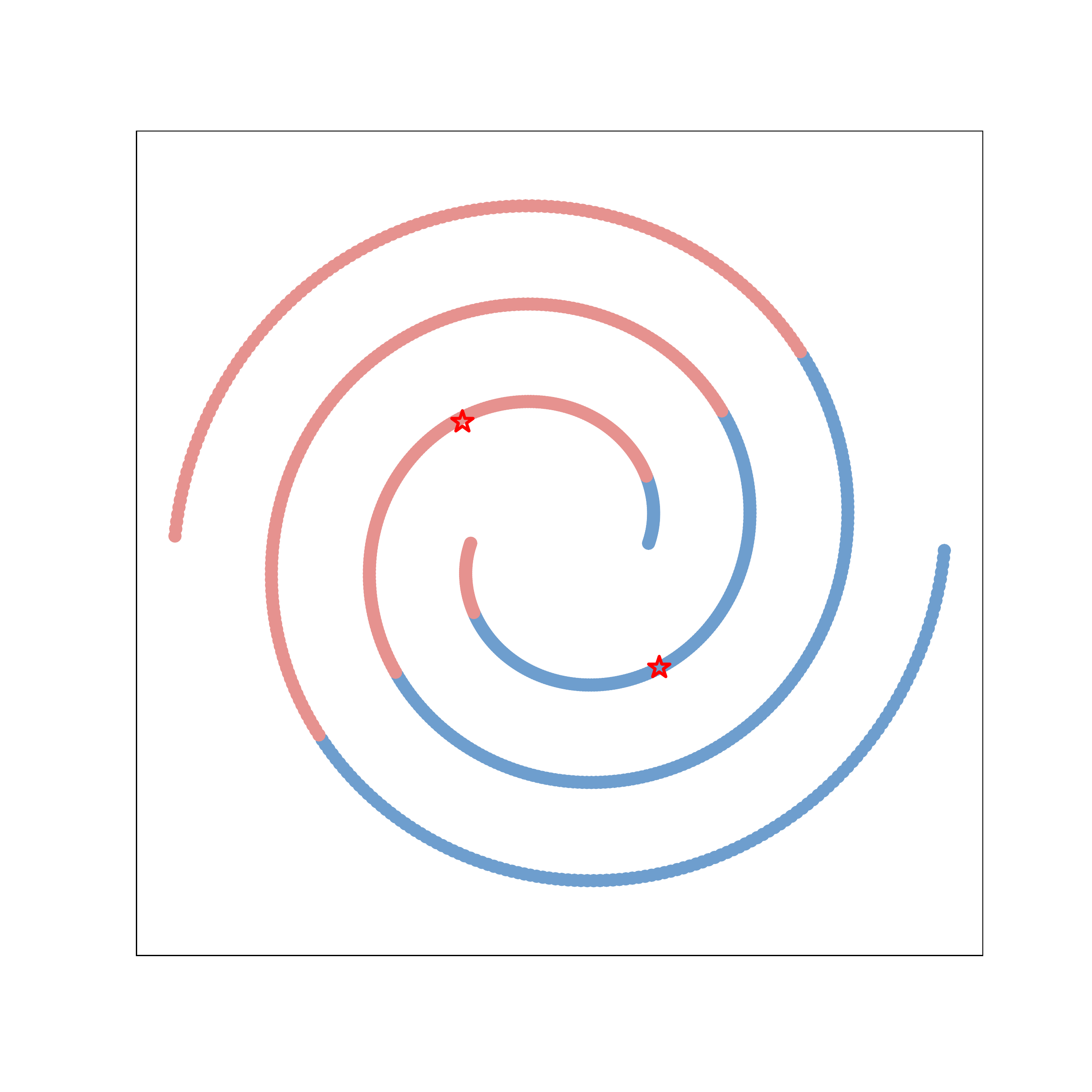}
		\label{fig: 2spiral k_means}
	}
	\hskip -30pt  
	\subfigure[SC]{
		\includegraphics[width=4.5cm]{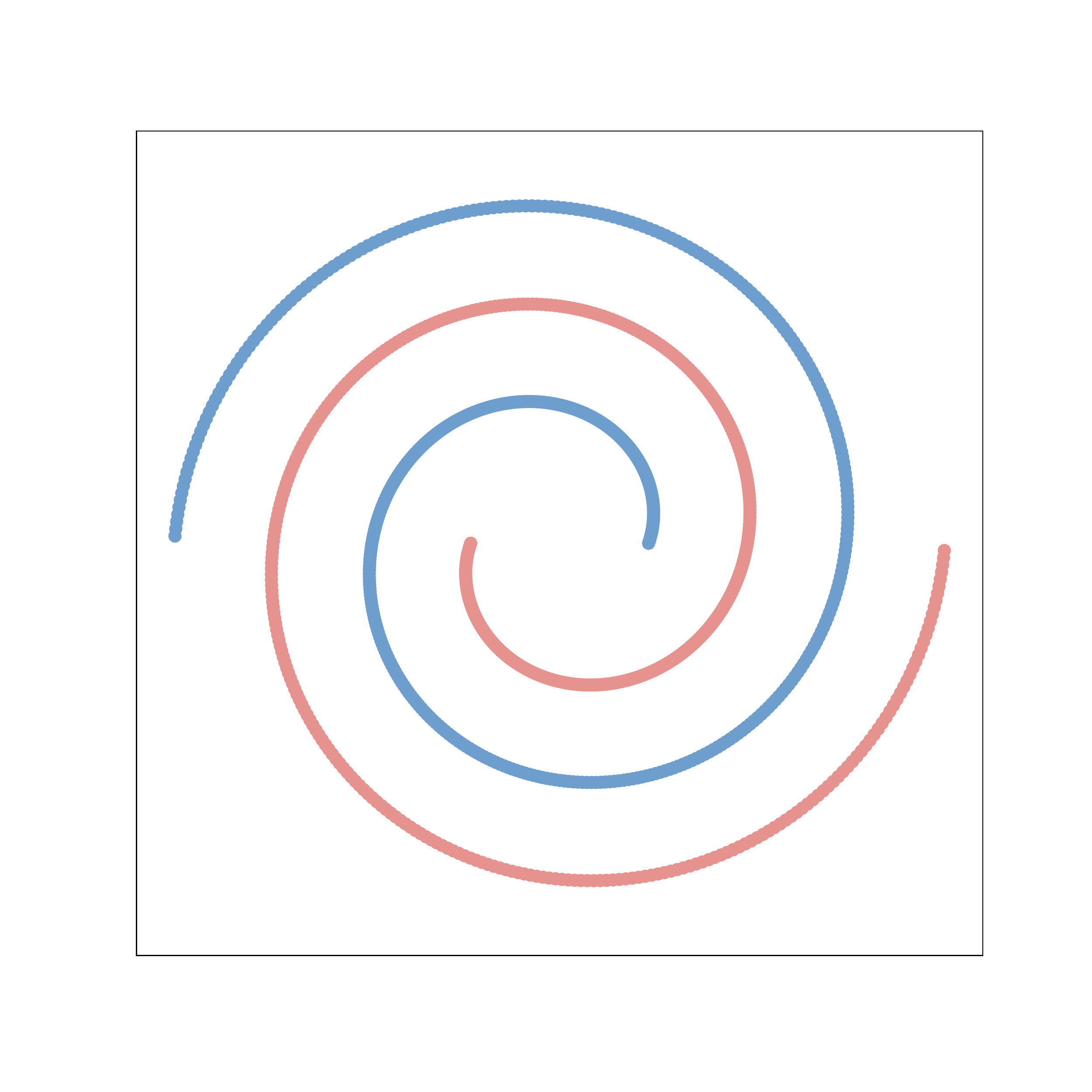}
		\label{fig: 2spiral SC}
	}
	\hskip -30pt  
	\subfigure[DBSCAN]{
		\includegraphics[width=4.5cm]{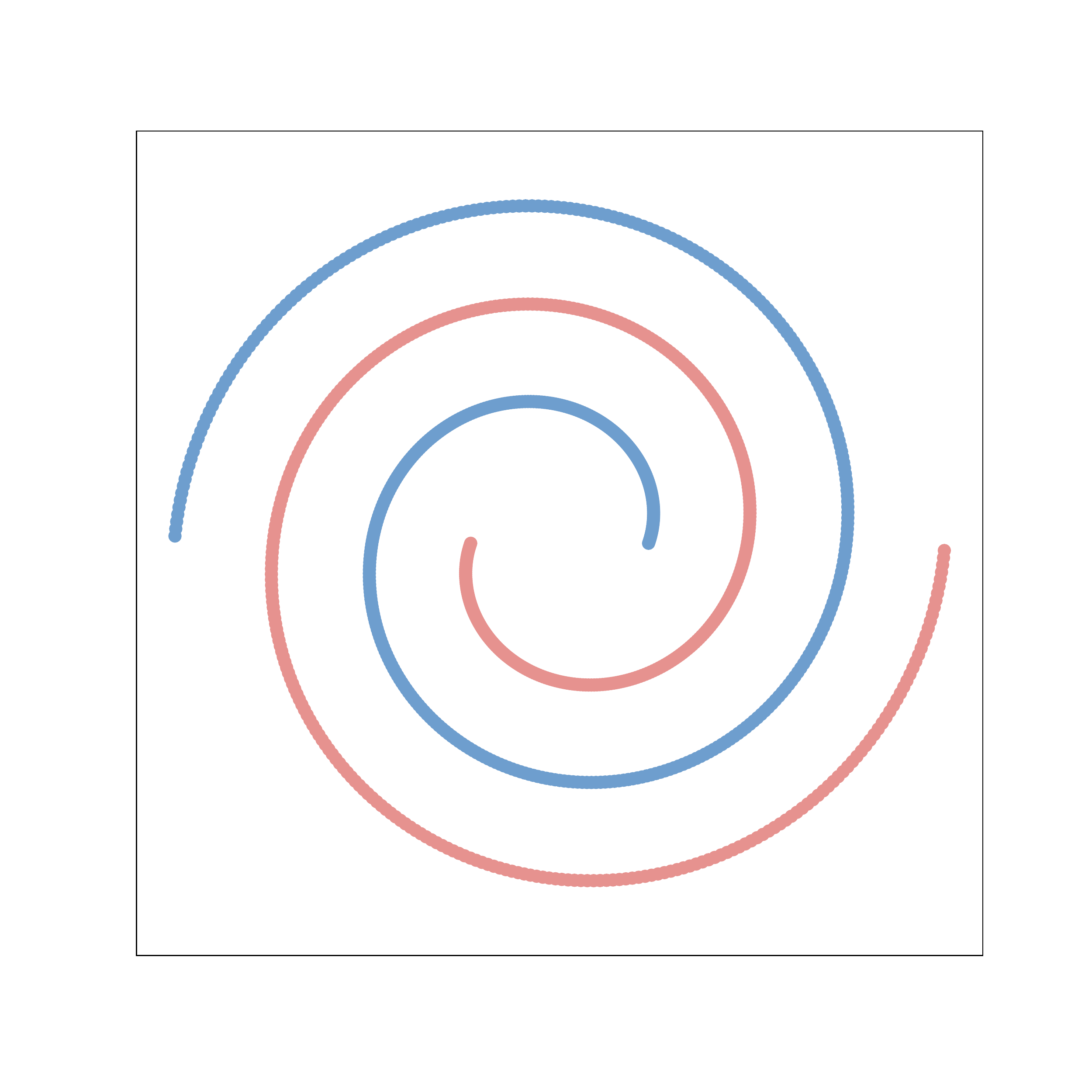}
		\label{fig: 2spiral DBSCAN}
	}
	\vskip -20pt  
	\subfigure[DPC]{
		\includegraphics[width=4.5cm]{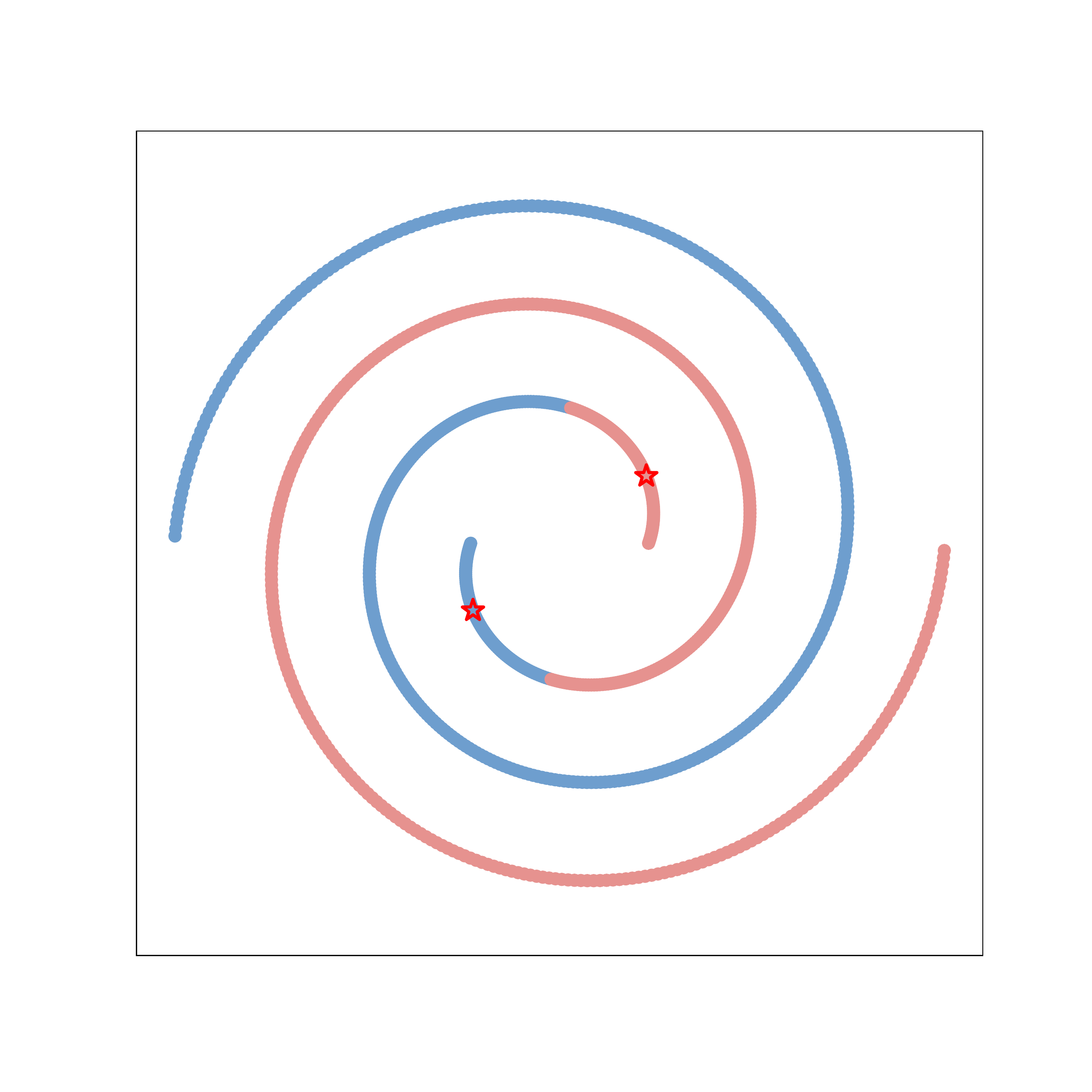}
		\label{fig: 2spiral DPC}
	}
	\hskip -30pt  
	\subfigure[DPC-KNN]{
		\includegraphics[width=4.5cm]{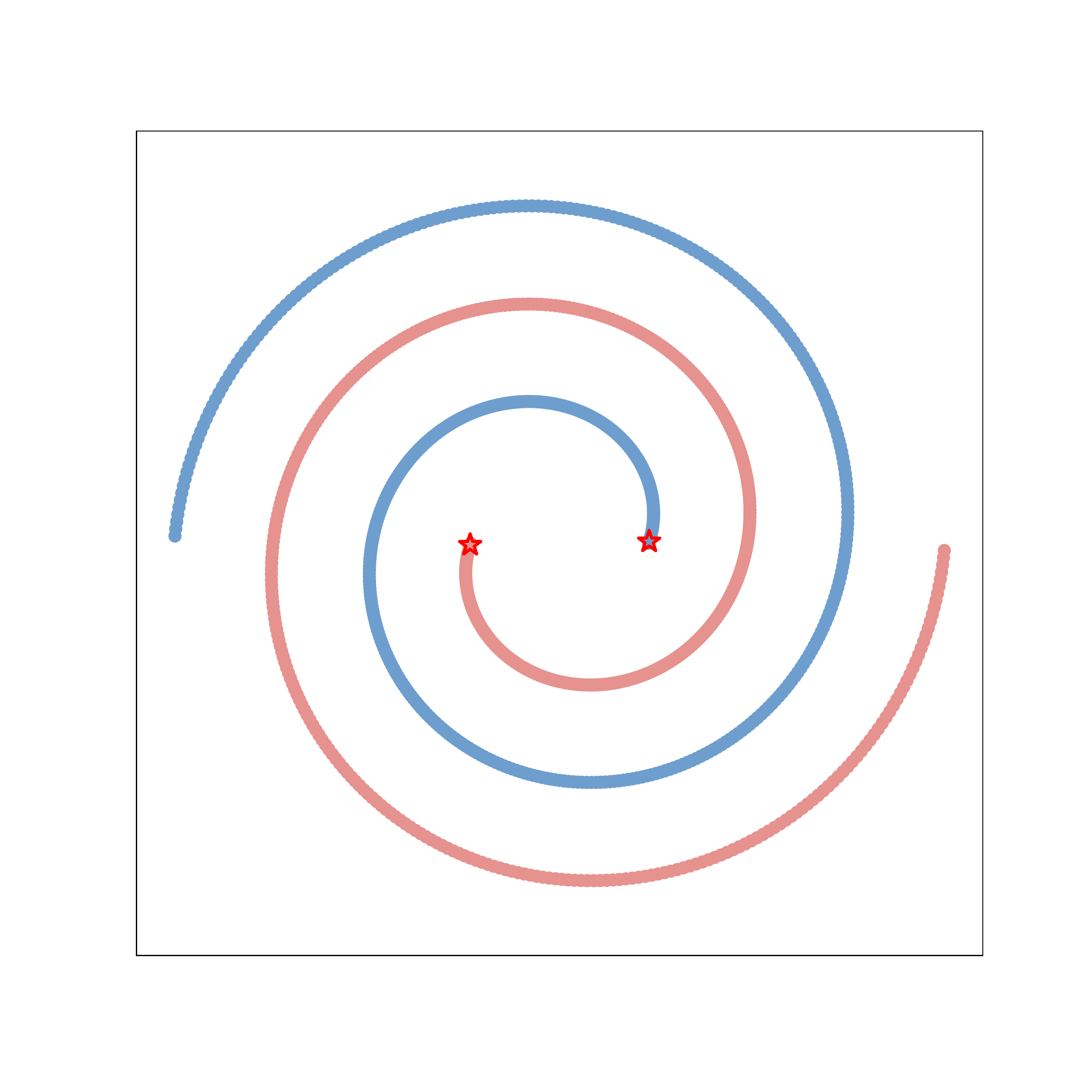}
		\label{fig: 2spiral DPCKNN}
	}
	\hskip -30pt  
	\subfigure[ECM]{
		\includegraphics[width=4.5cm]{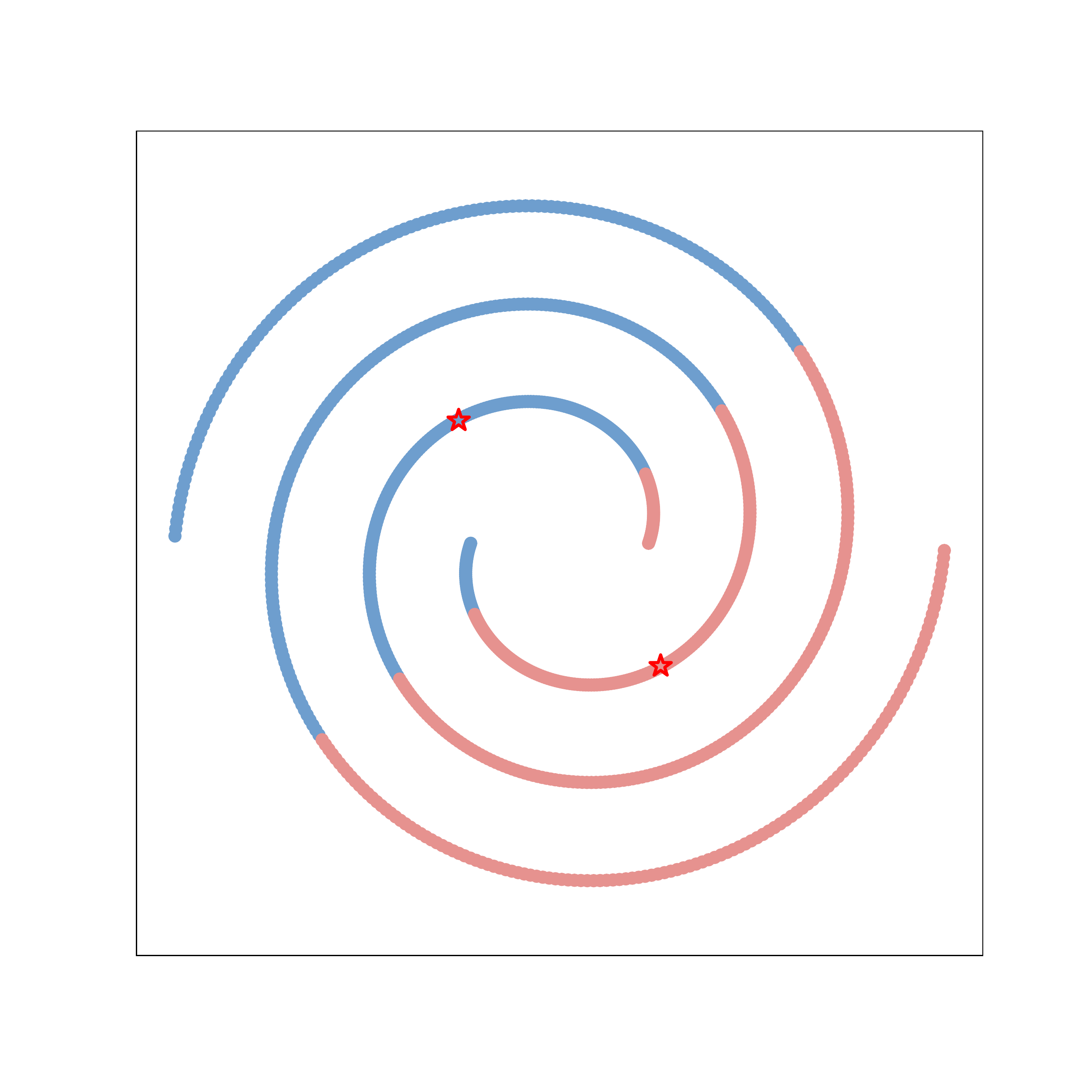}
		\label{fig: 2spiral ECM}
	}
	\hskip -30pt  
	\subfigure[GFDC]{
		\includegraphics[width=4.5cm]{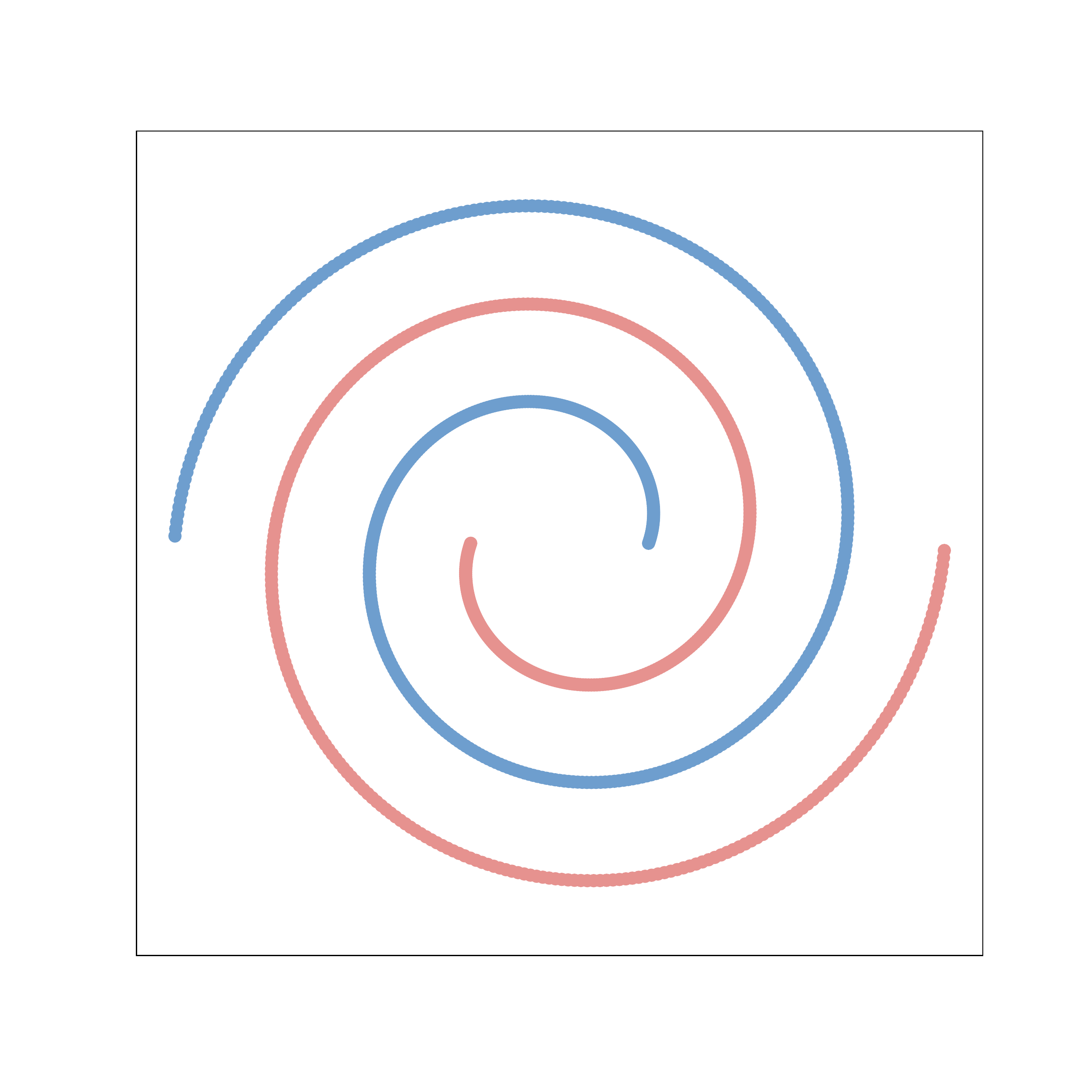}
		\label{fig: 2spiral proposed}
	}
	\caption{2spiral.}
	\label{fig: 2spiral}
\end{figure}

\begin{figure}[H]\footnotesize
	\centering
	\vspace{-25pt}  
	\setlength{\abovecaptionskip}{0pt}  
	\subfigcapskip=-15pt  
	\subfigure[Ground truth]{
		\includegraphics[width=4.5cm]{aggregation-groundtruth.pdf}
		\label{fig: aggregation ground truth}
	}
	\hskip -30pt
	\subfigure[$k$-means++]{
		\includegraphics[width=4.5cm]{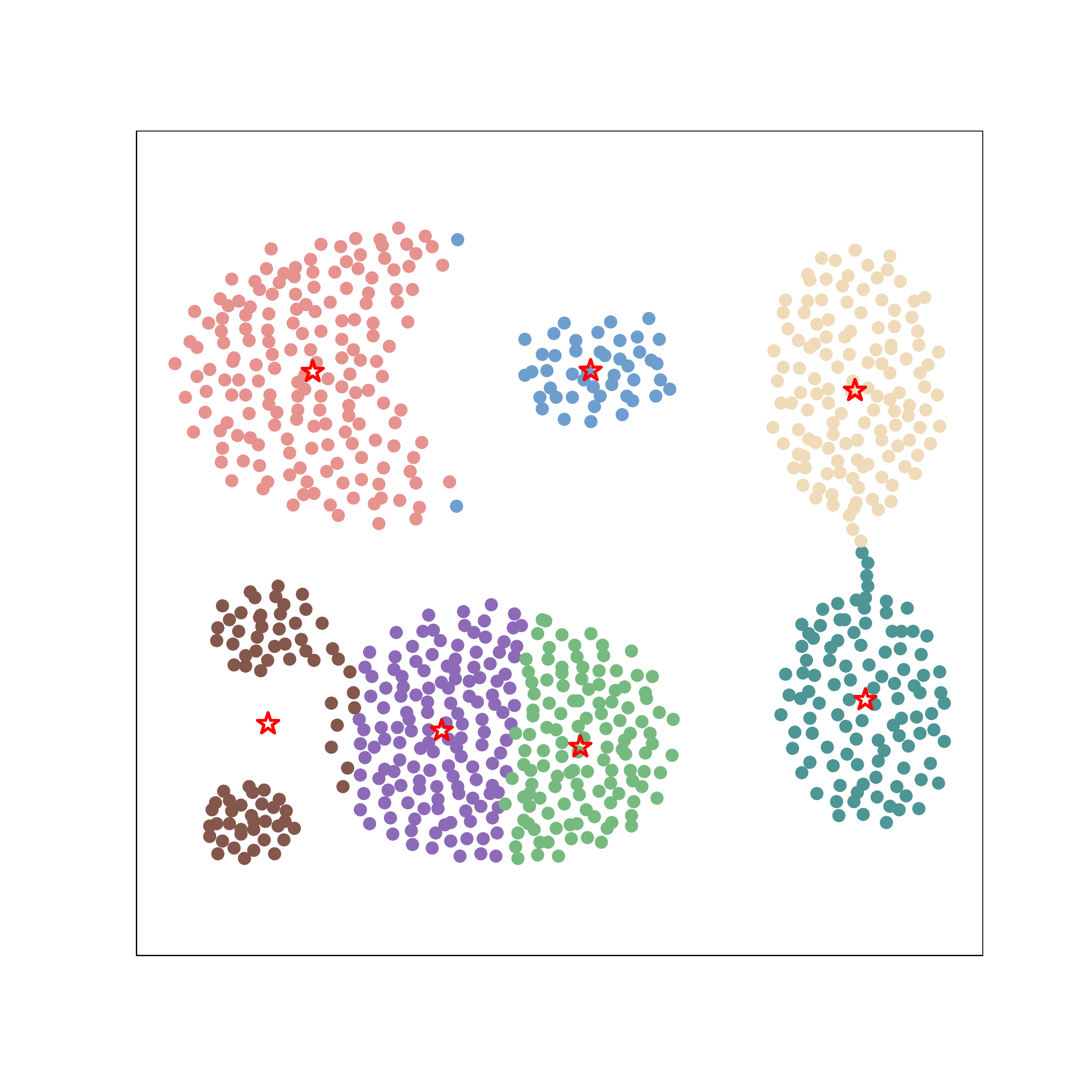}
		\label{fig: aggregation k_means}
	}
	\hskip -30pt
	\subfigure[SC]{
		\includegraphics[width=4.5cm]{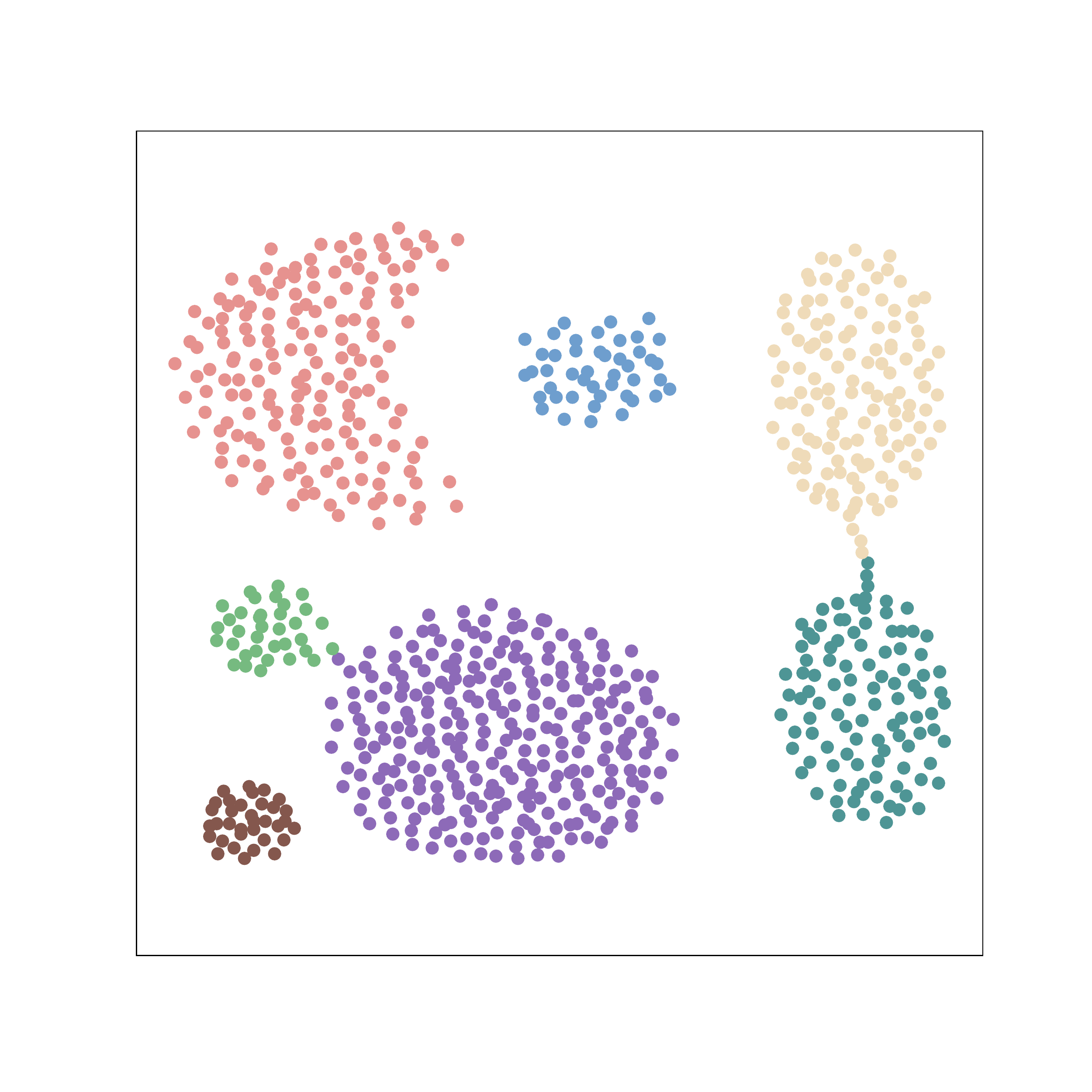}
		\label{fig: aggregation SC}
	}
	\hskip -30pt  
	\subfigure[DBSCAN]{
		\includegraphics[width=4.5cm]{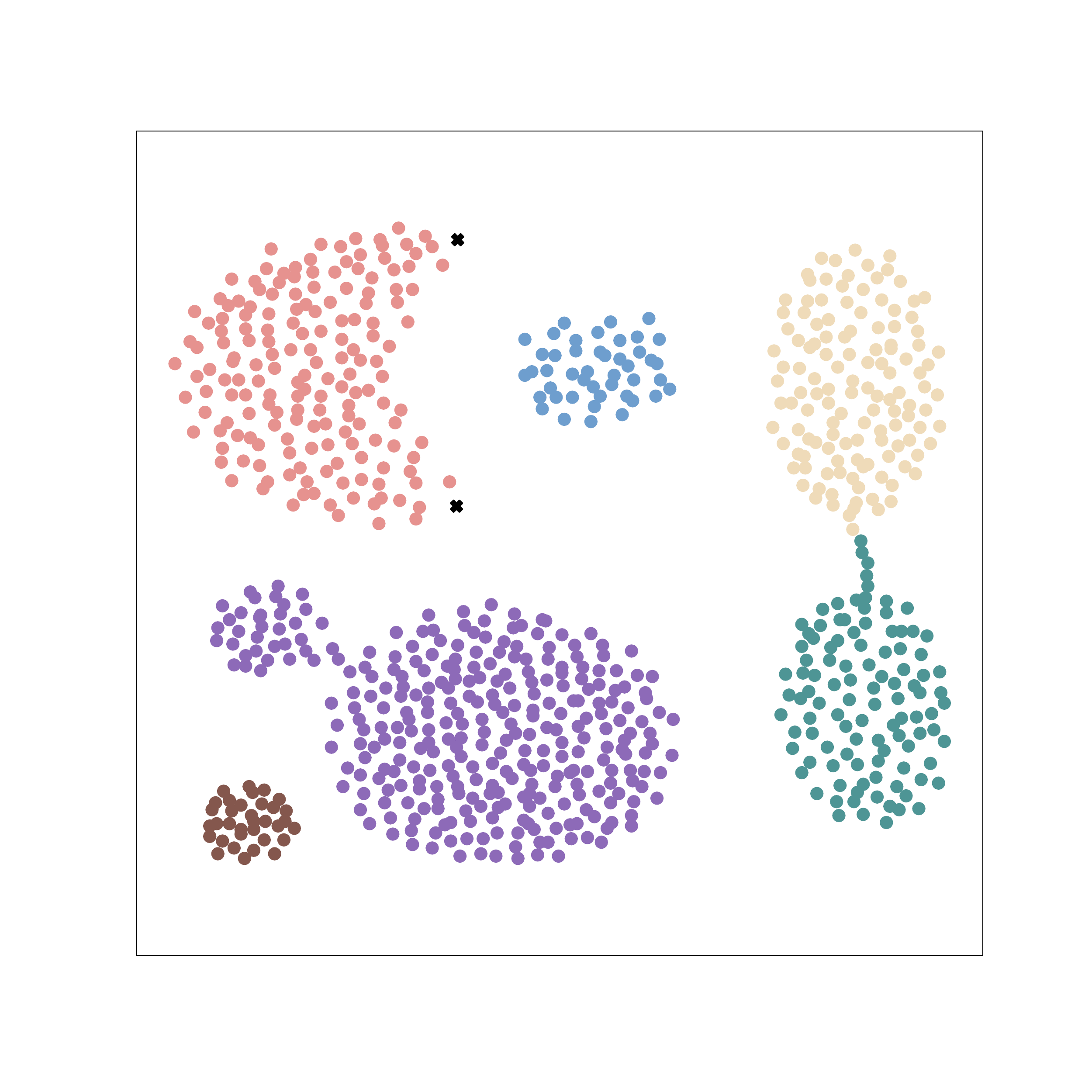}
		\label{fig: aggregation DBSCAN}
	}
	\vskip -20pt
	\subfigure[DPC]{
		\includegraphics[width=4.5cm]{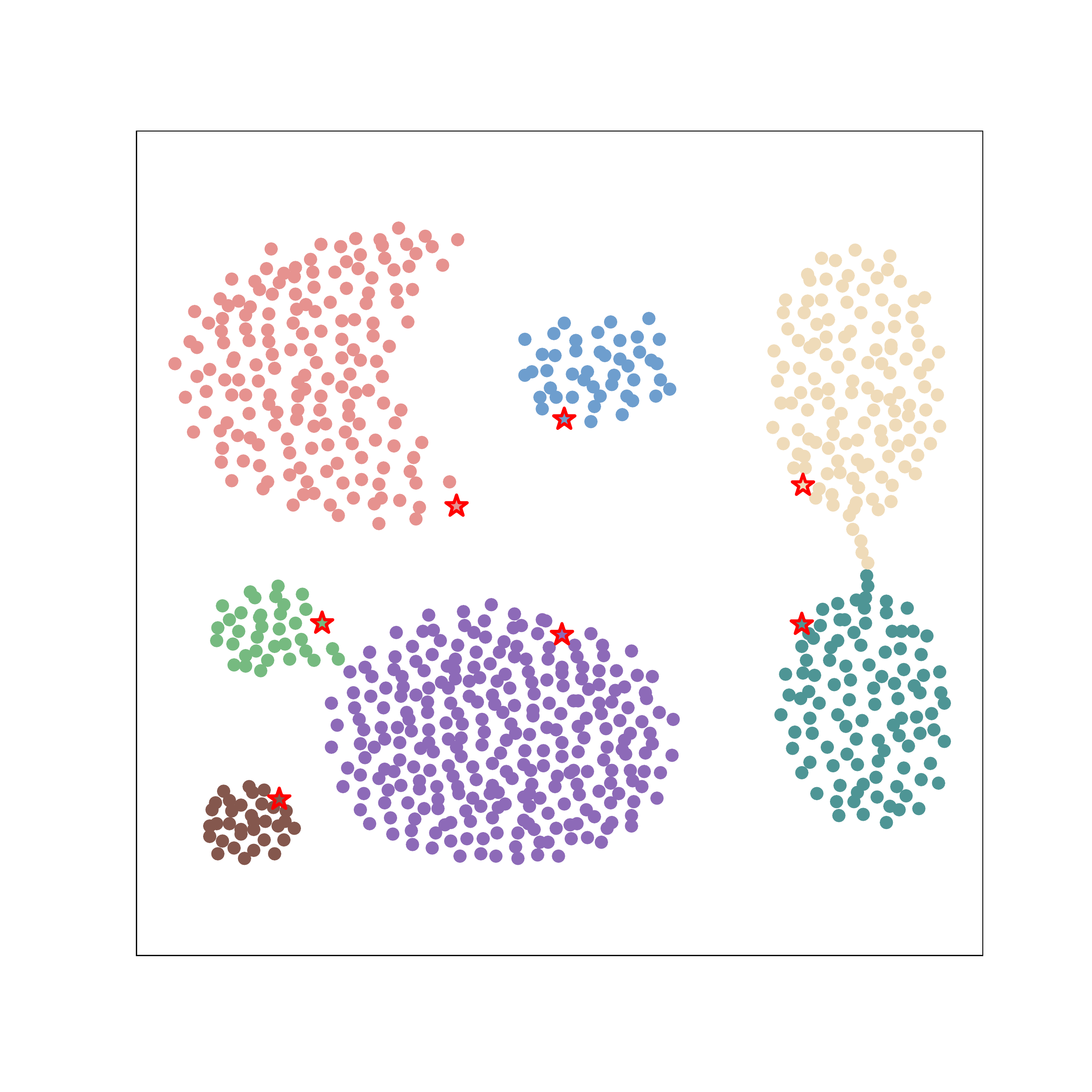}
		\label{fig: aggregation DPC}
	}
	\hskip -30pt
	\subfigure[DPC-KNN]{
		\includegraphics[width=4.5cm]{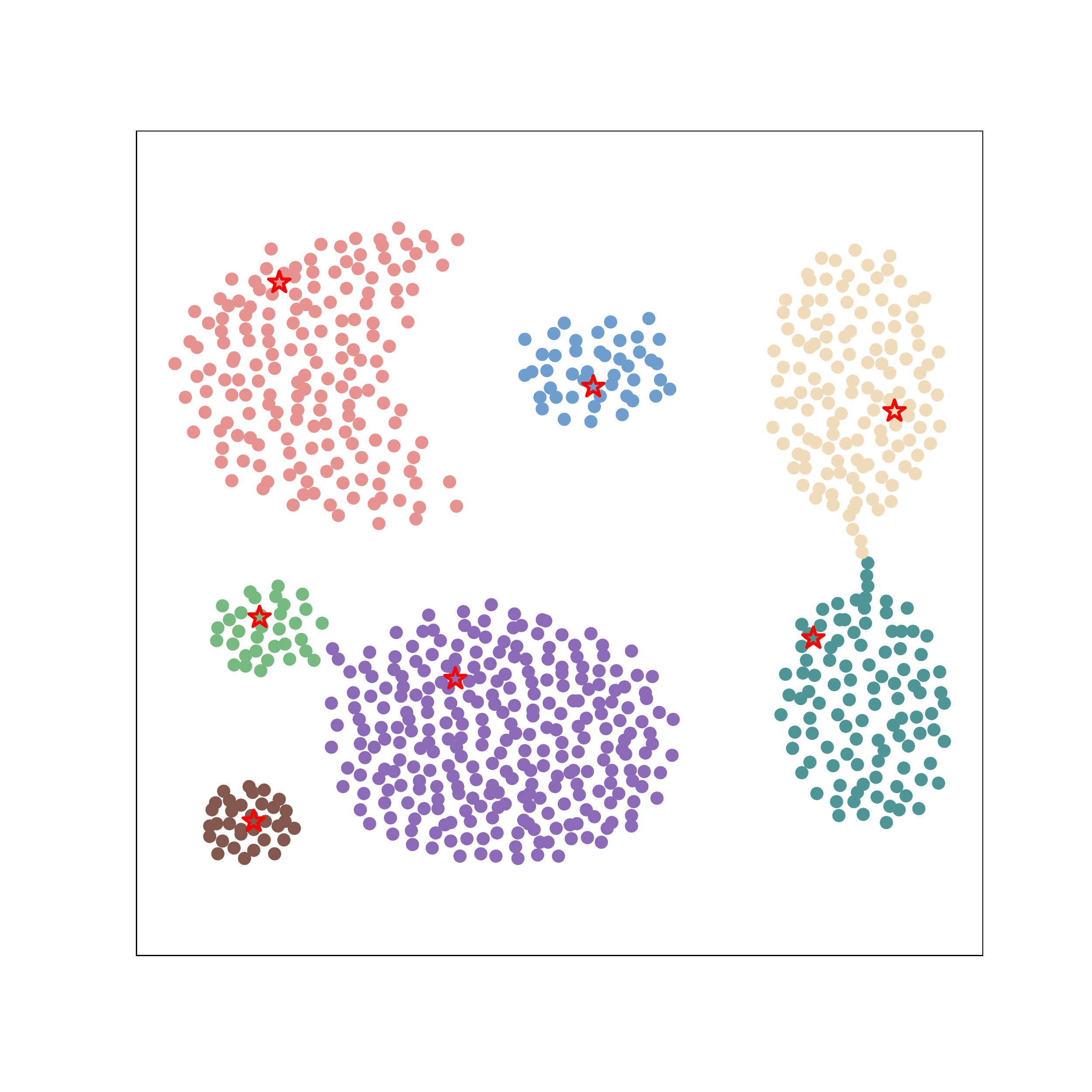}
		\label{fig: aggregation DPCKNN}
	}
	\hskip -30pt
	\subfigure[ECM]{
		\includegraphics[width=4.5cm]{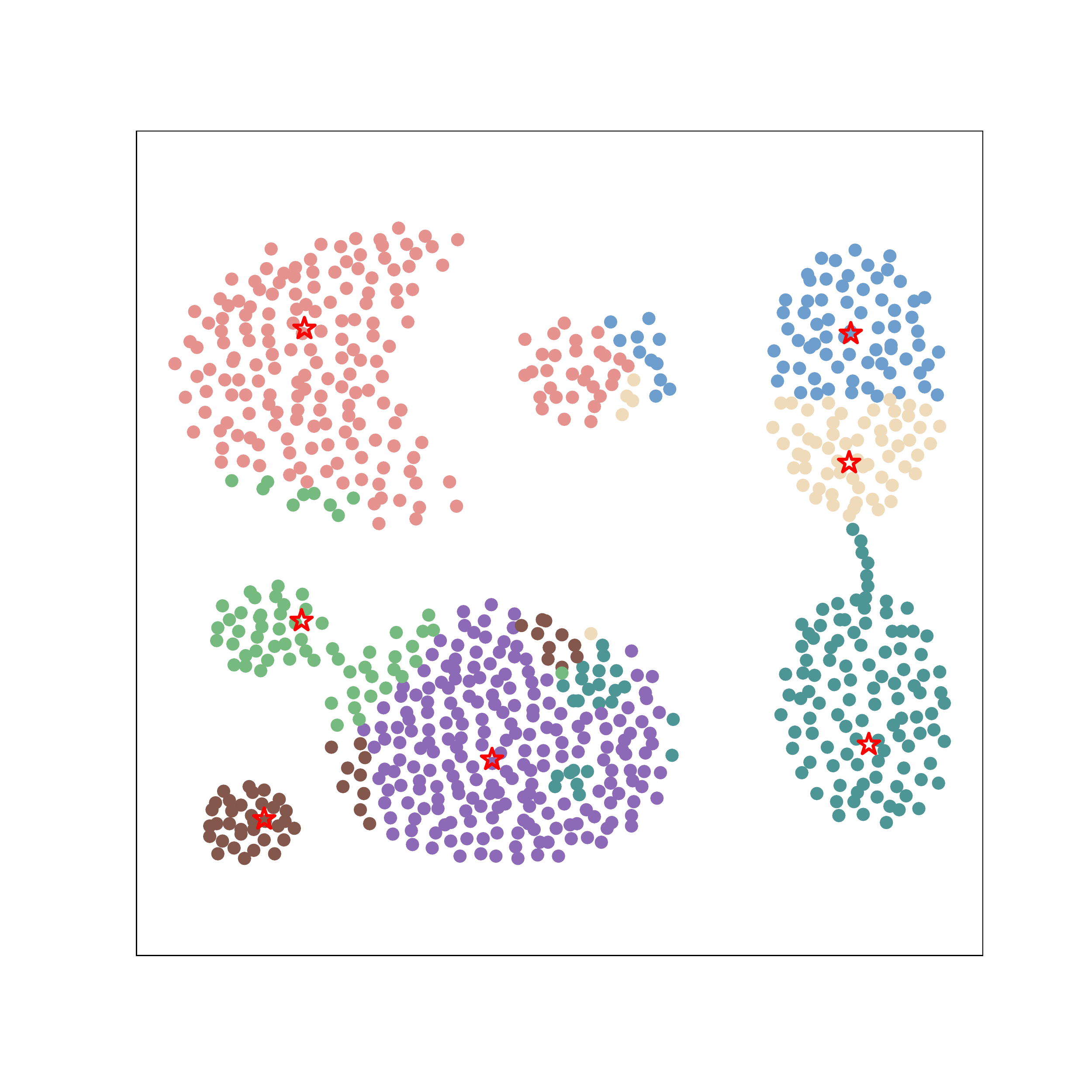}
		\label{fig: aggregation ECM}
	}
	\hskip -30pt
	\subfigure[GFDC]{
		\includegraphics[width=4.5cm]{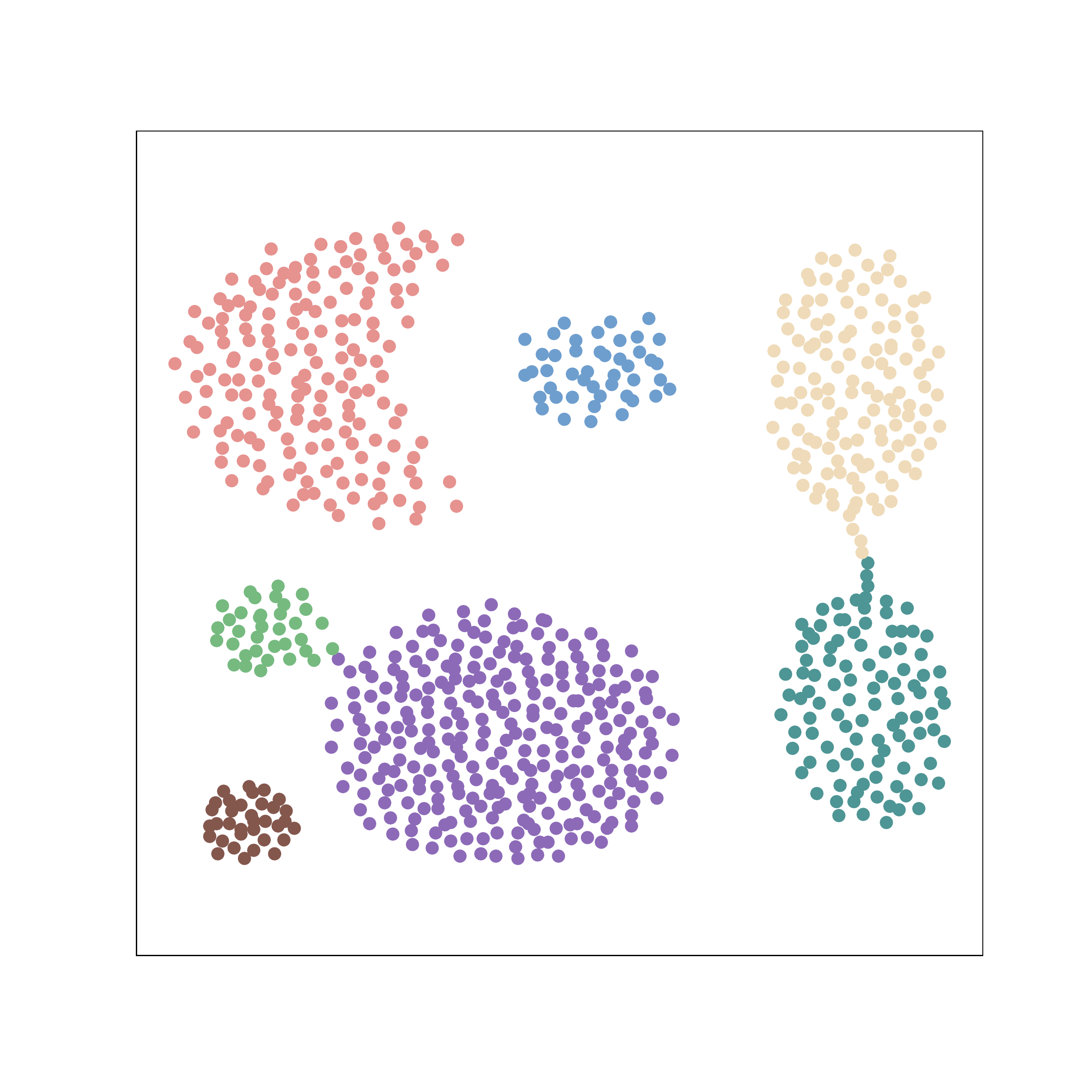}
		\label{fig: aggregation proposed}
	}
	\caption{Aggregation.}
	\label{fig: aggregation}
\end{figure}

\begin{figure}[htbp]\footnotesize
	\centering
	\vspace{-25pt}  
	\setlength{\abovecaptionskip}{0pt}  
	\subfigcapskip=-15pt  
	\subfigure[Ground truth]{
		\includegraphics[width=4.5cm]{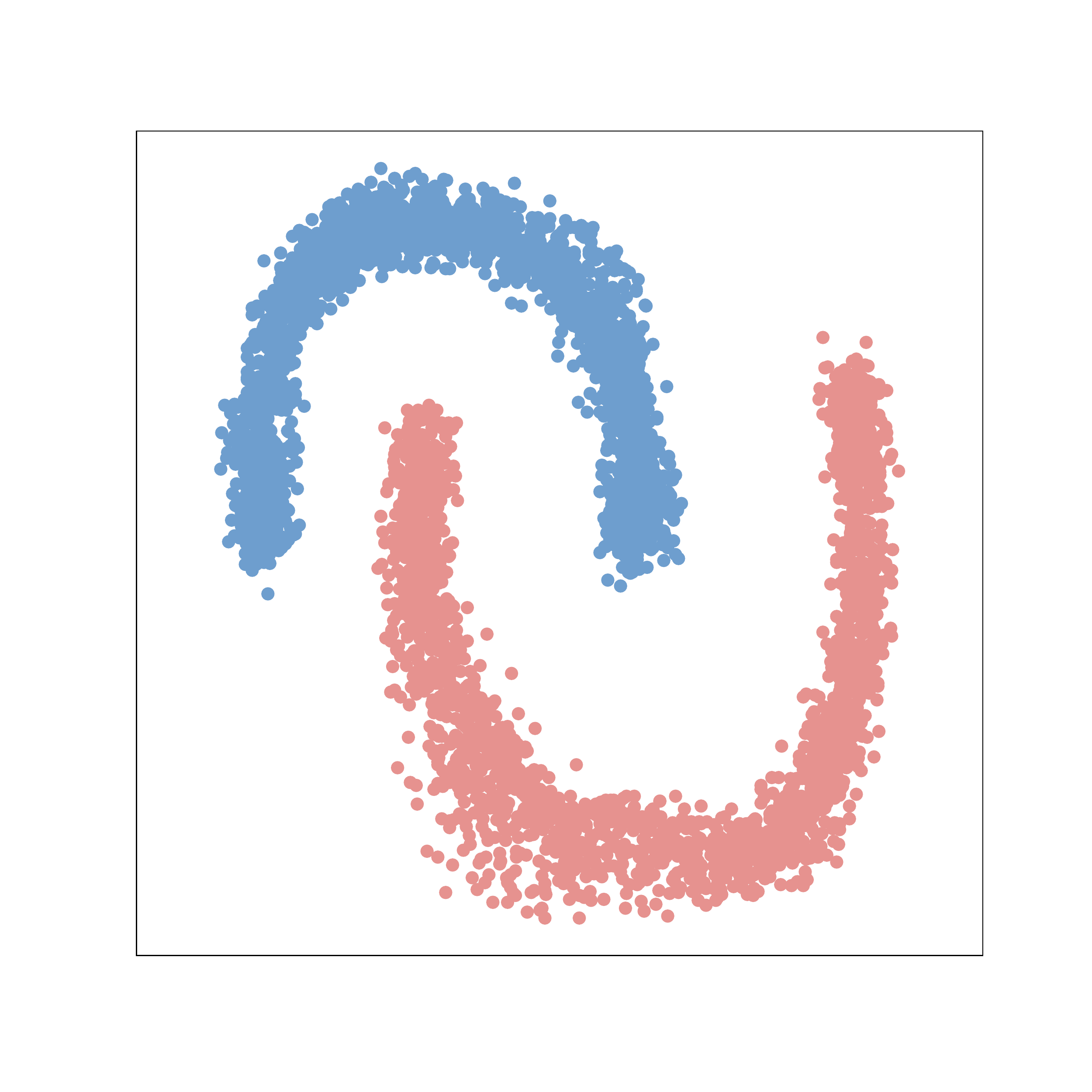}
		\label{fig: banana Ori ground truth}
	}
	\hskip -30pt
	\subfigure[$k$-means++]{
		\includegraphics[width=4.5cm]{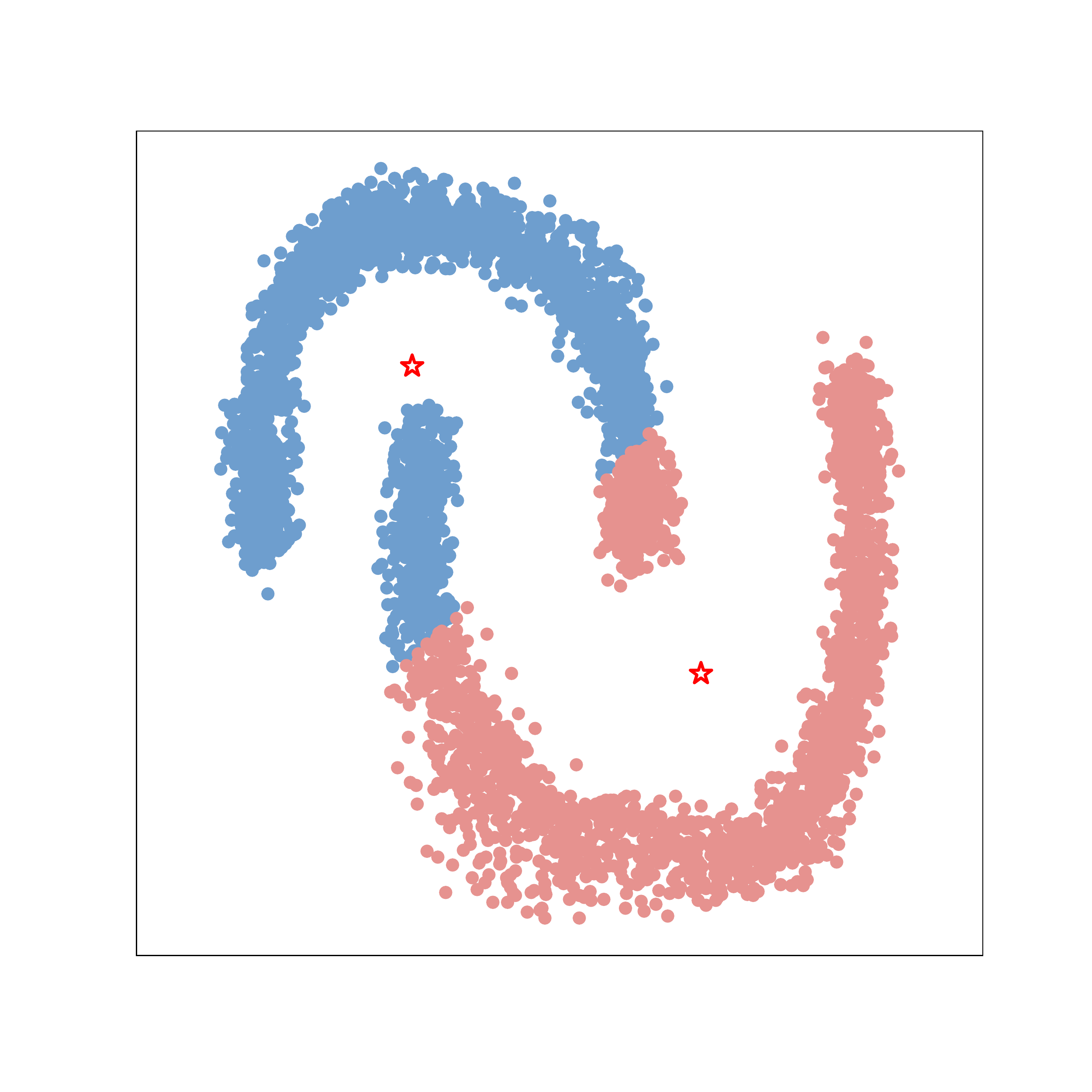}
		\label{fig: banana Ori k_means}
	}
	\hskip -30pt
	\subfigure[SC]{
		\includegraphics[width=4.5cm]{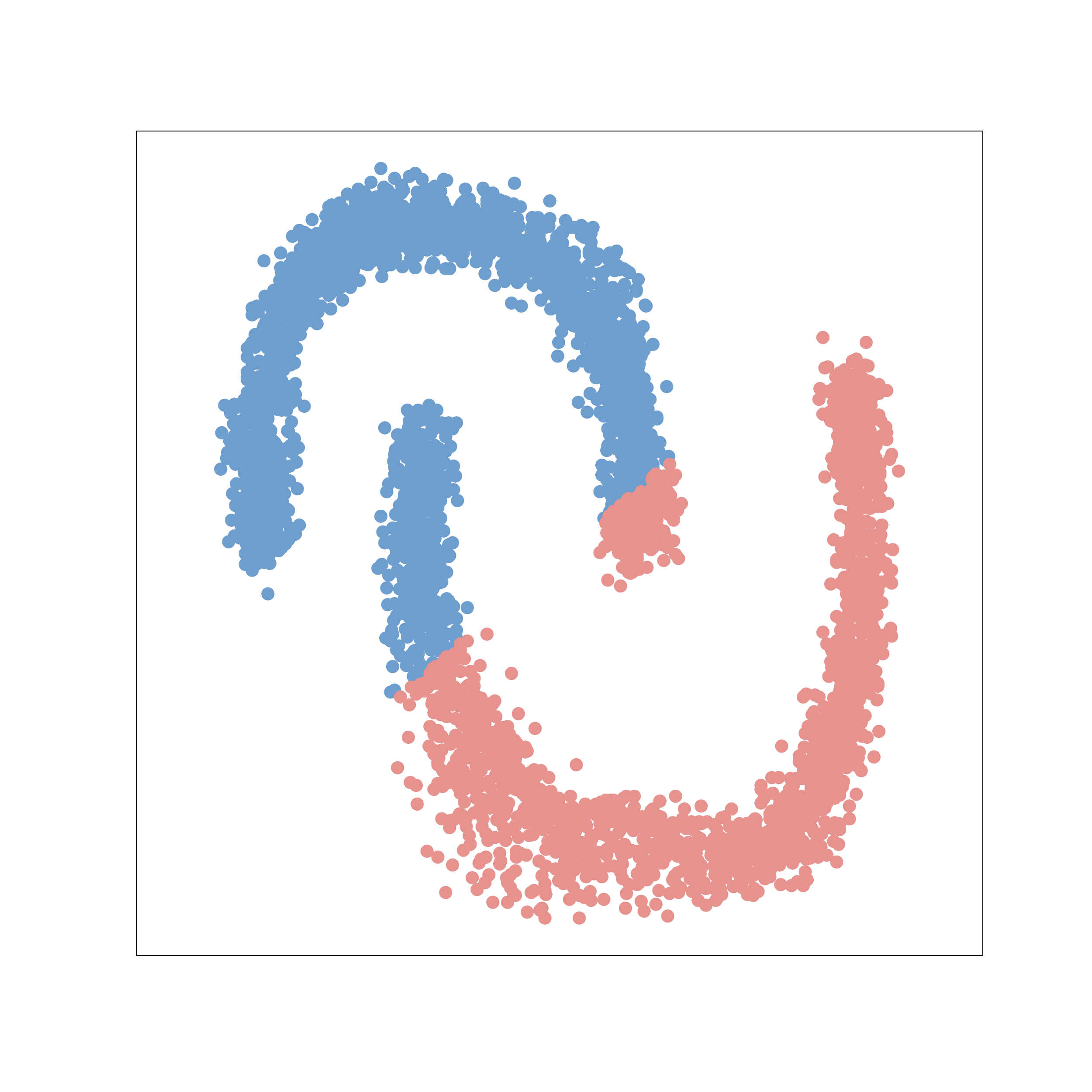}
		\label{fig: banana Ori SC}
	}
	\hskip -30pt  
	\subfigure[DBSCAN]{
		\includegraphics[width=4.5cm]{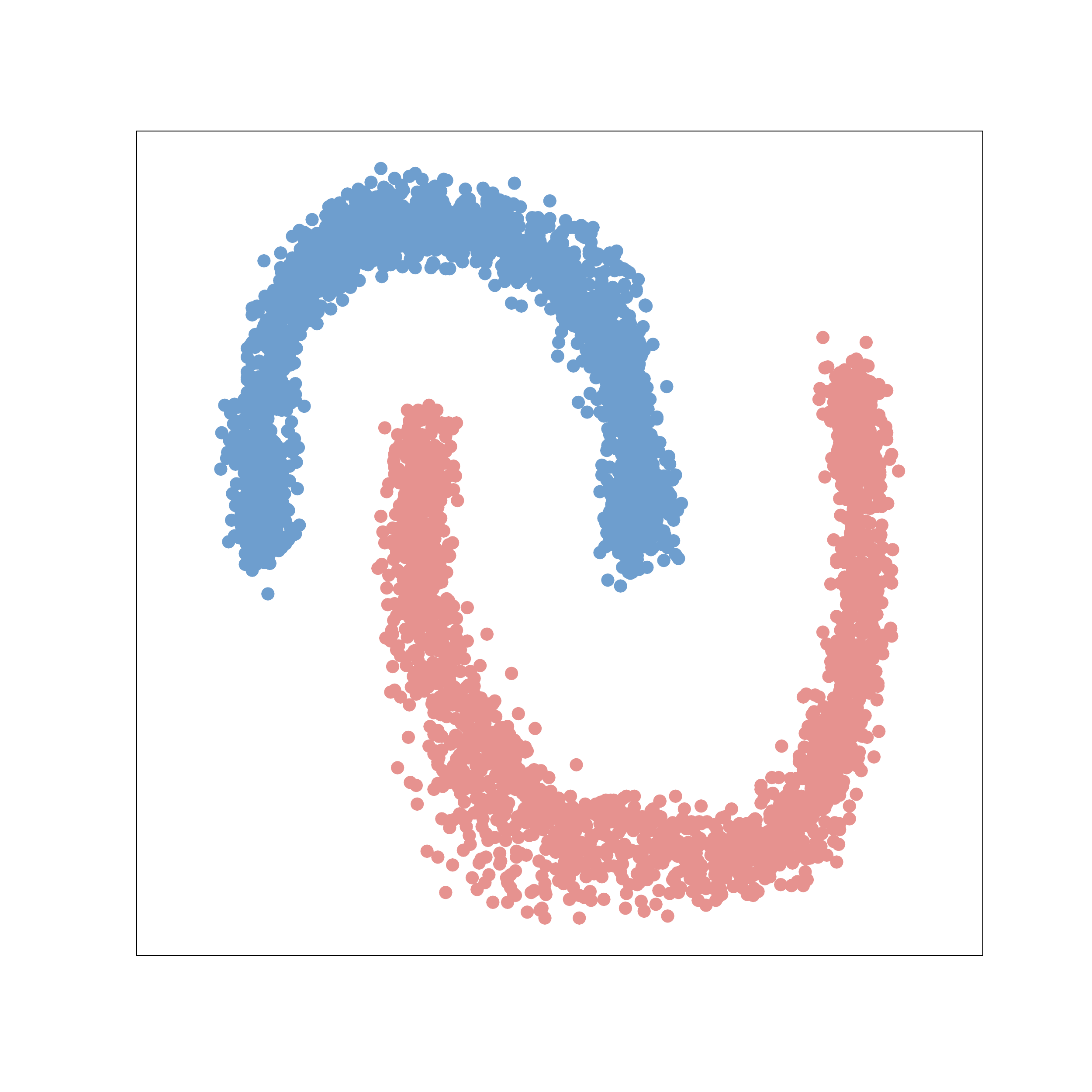}
		\label{fig: banana Ori DBSCAN}
	}
	\vskip -20pt
	\subfigure[DPC]{
		\includegraphics[width=4.5cm]{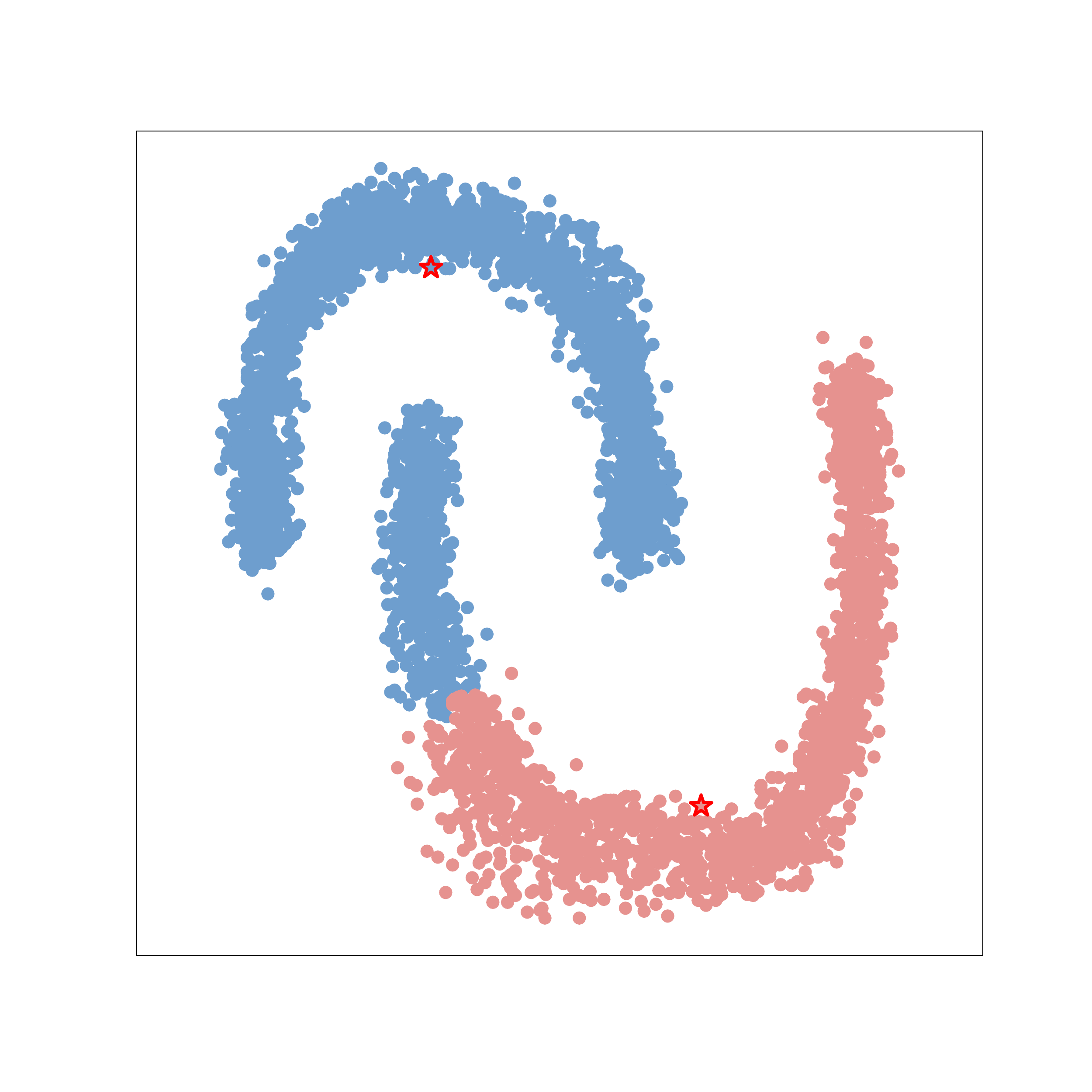}
		\label{fig: banana Ori DPC}
	}
	\hskip -30pt
	\subfigure[DPC-KNN]{
		\includegraphics[width=4.5cm]{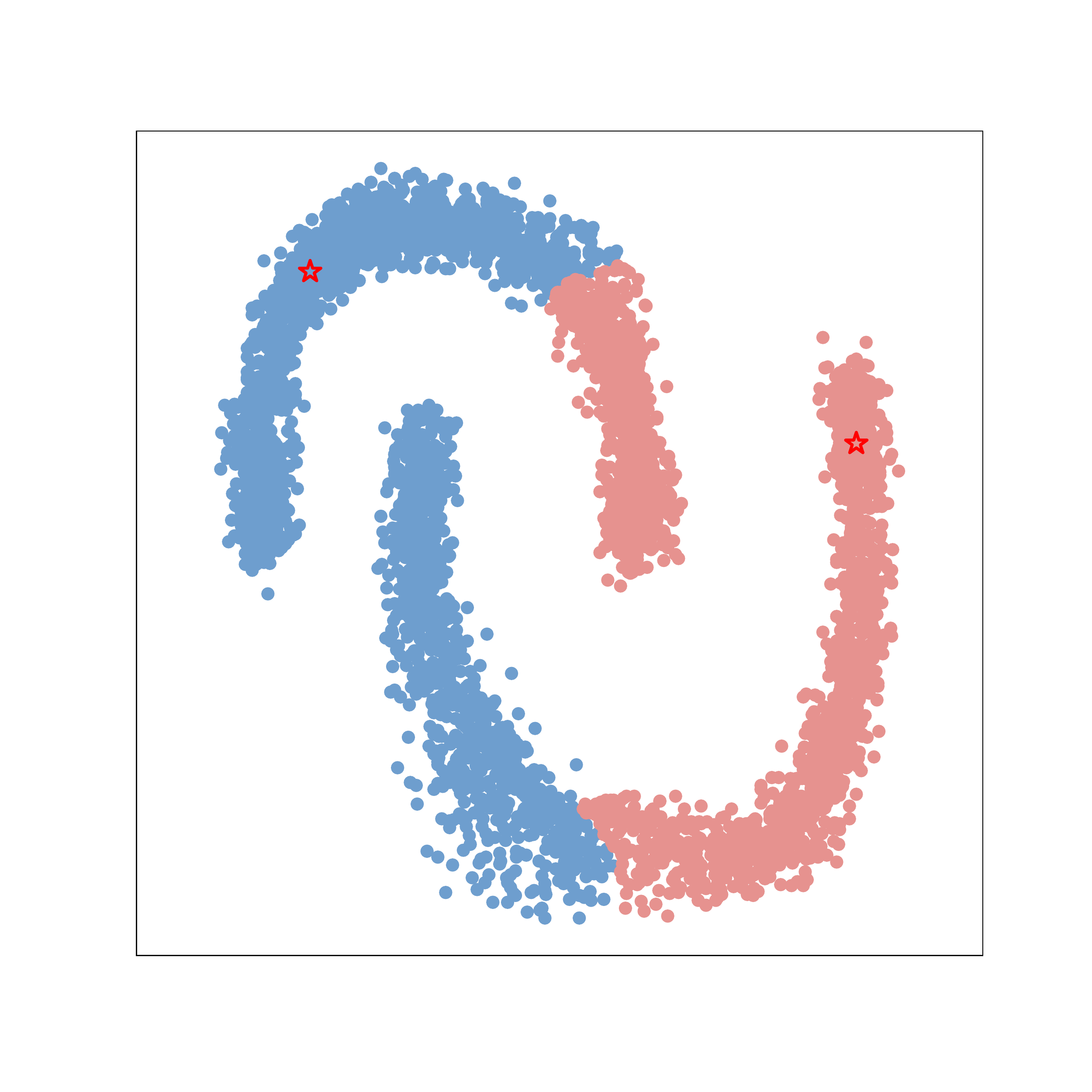}
		\label{fig: banana Ori DPCKNN}
	}
	\hskip -30pt
	\subfigure[ECM]{
		\includegraphics[width=4.5cm]{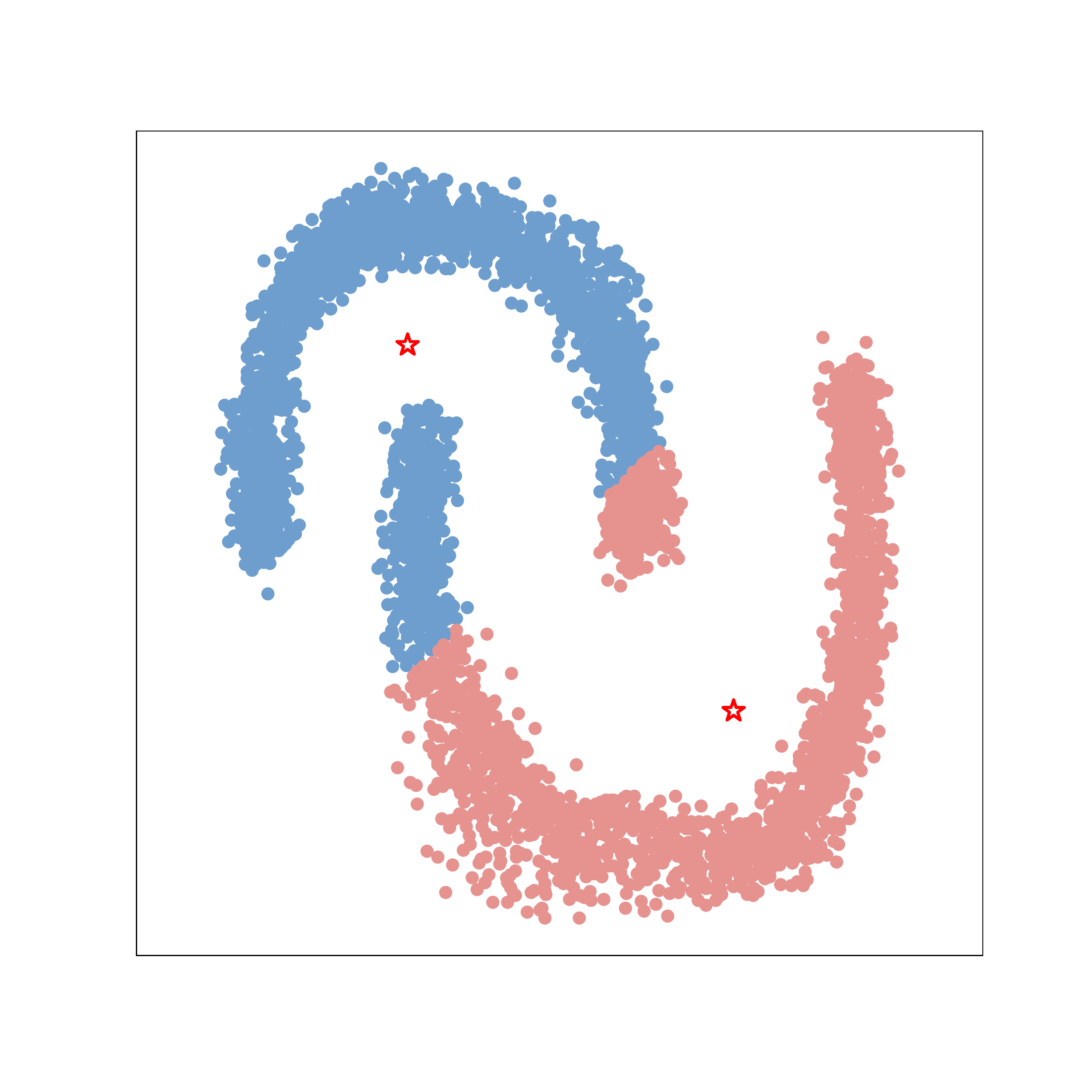}
		\label{fig: banana Ori ECM}
	}
	\hskip -30pt
	\subfigure[GFDC]{
		\includegraphics[width=4.5cm]{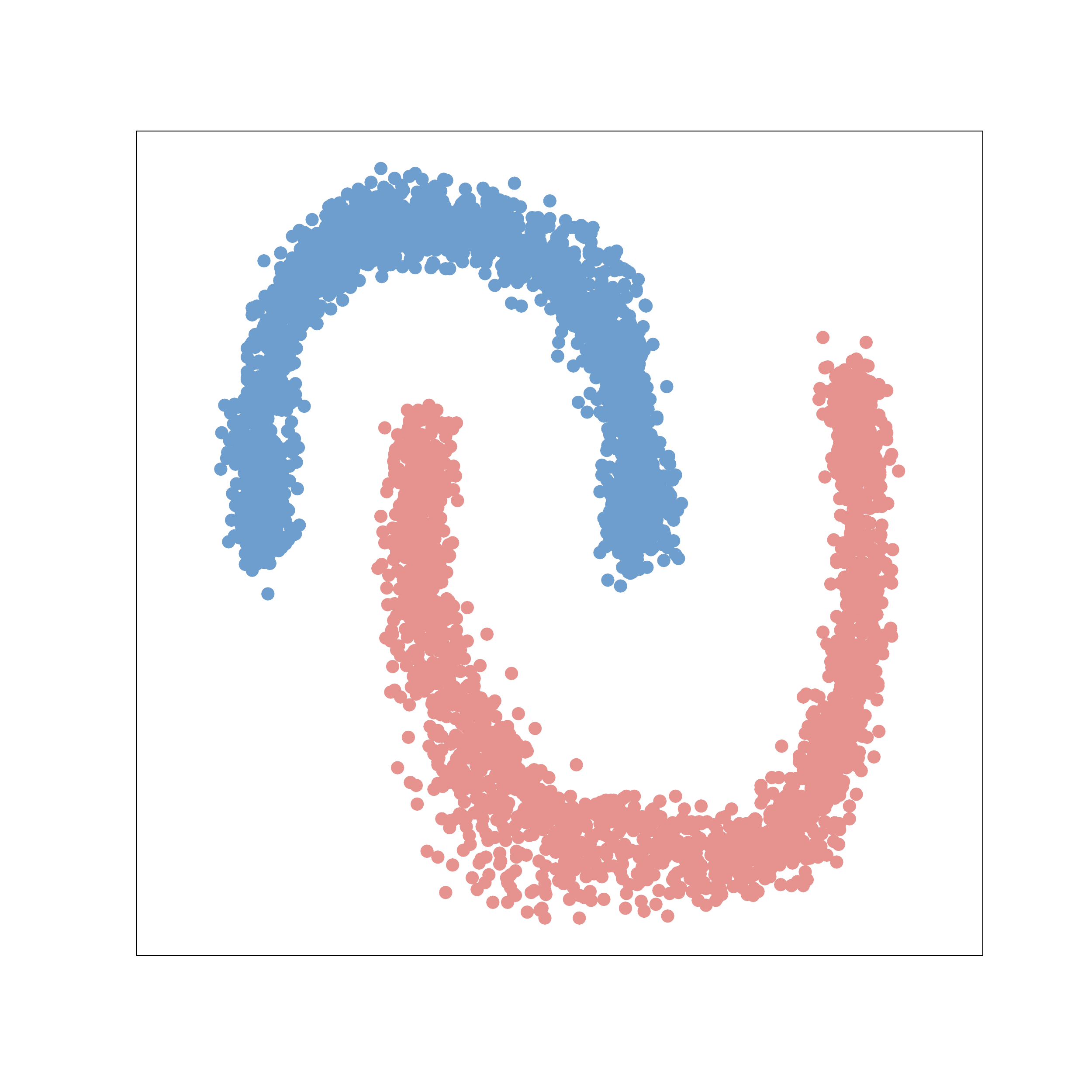}
		\label{fig: banana Ori proposed}
	}
	\caption{Banana-Ori.}
	\label{fig: banana Ori}
\end{figure}

\begin{figure}[H]\footnotesize
	\centering
	\vspace{-25pt}  
	\setlength{\abovecaptionskip}{0pt}  
	\subfigcapskip=-15pt  
	\subfigure[Ground truth]{
		\includegraphics[width=4.5cm]{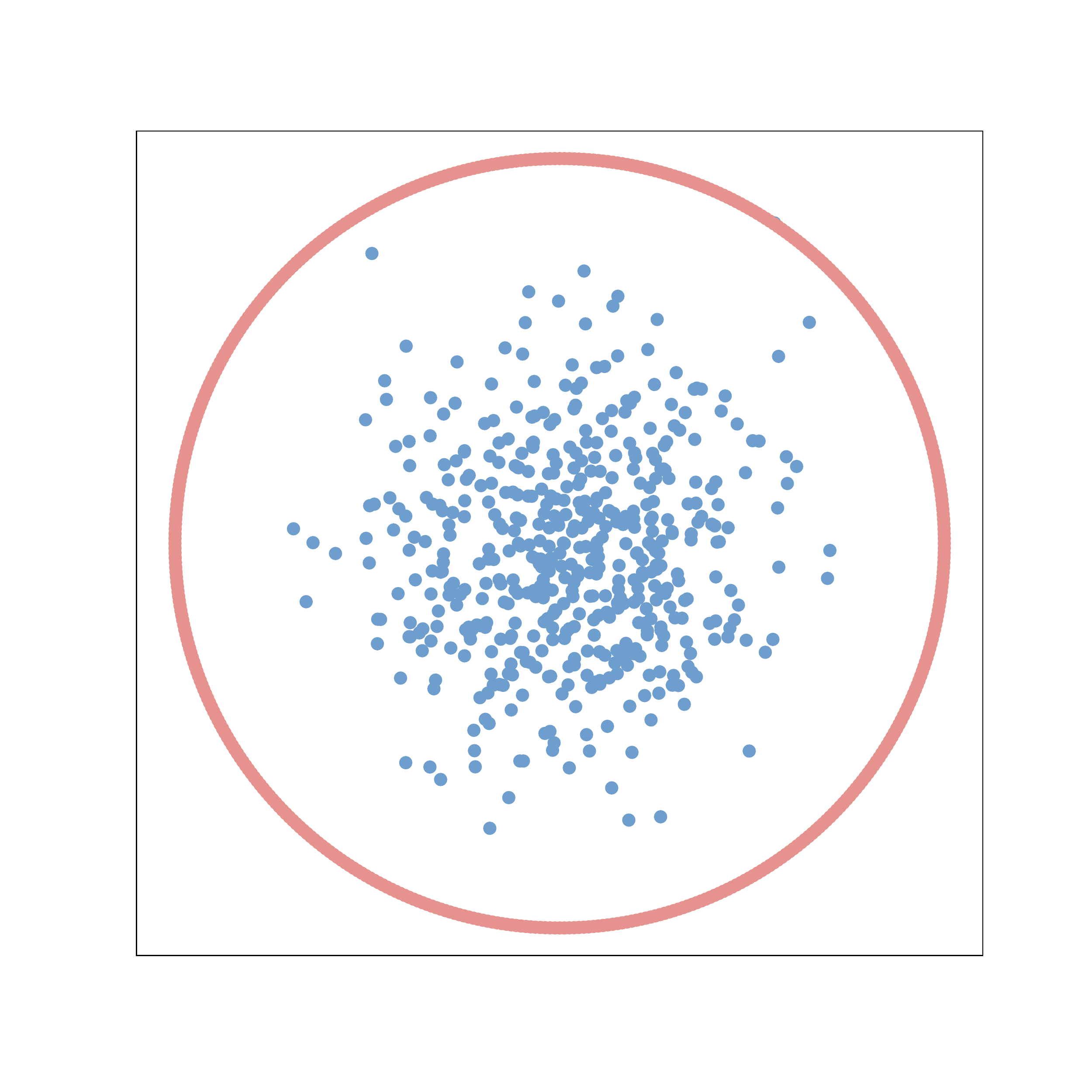}
		\label{fig: donut2 ground truth}
	}
	\hskip -30pt
	\subfigure[$k$-means++]{
		\includegraphics[width=4.5cm]{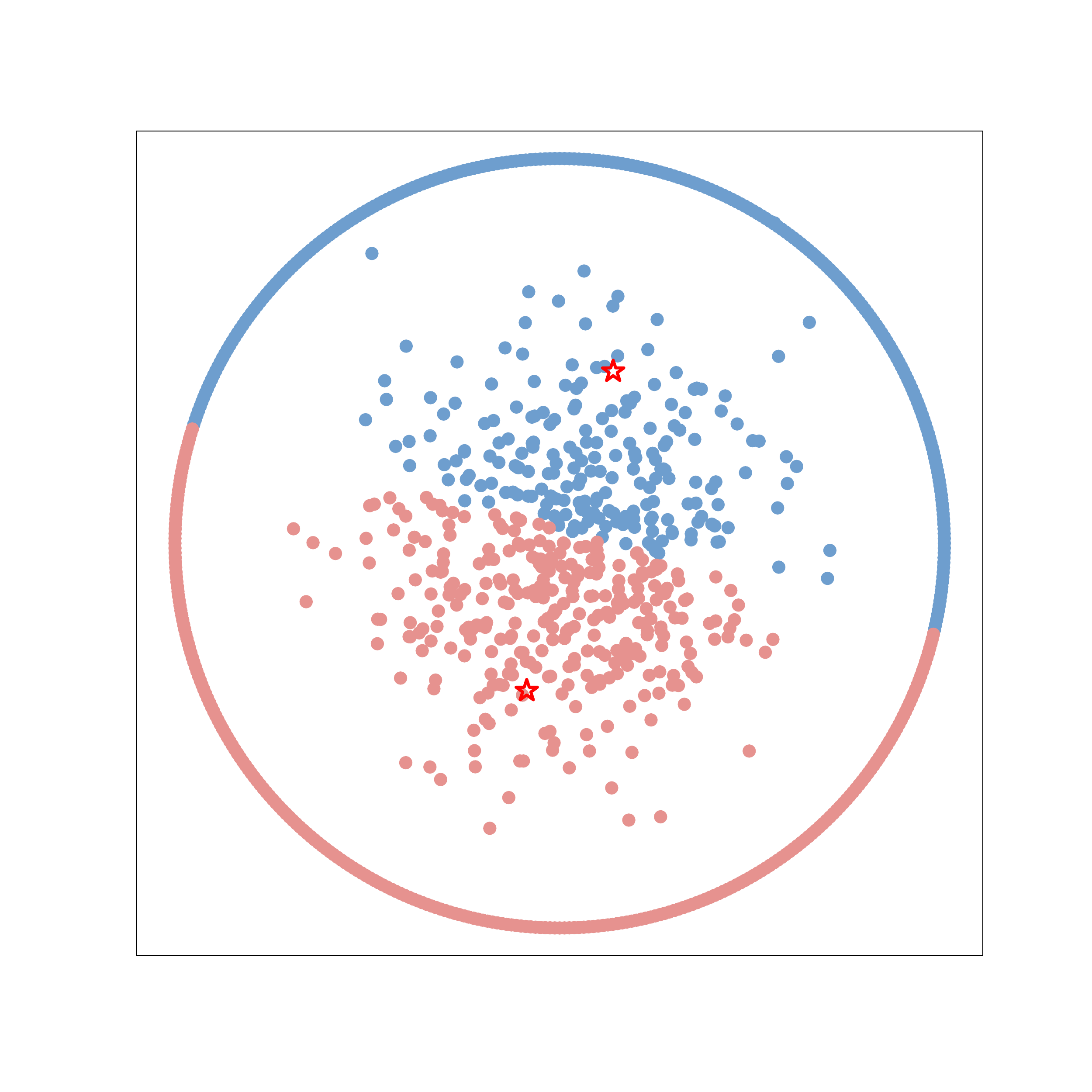}
		\label{fig: donut2 k_means}
	}
	\hskip -30pt
	\subfigure[SC]{
		\includegraphics[width=4.5cm]{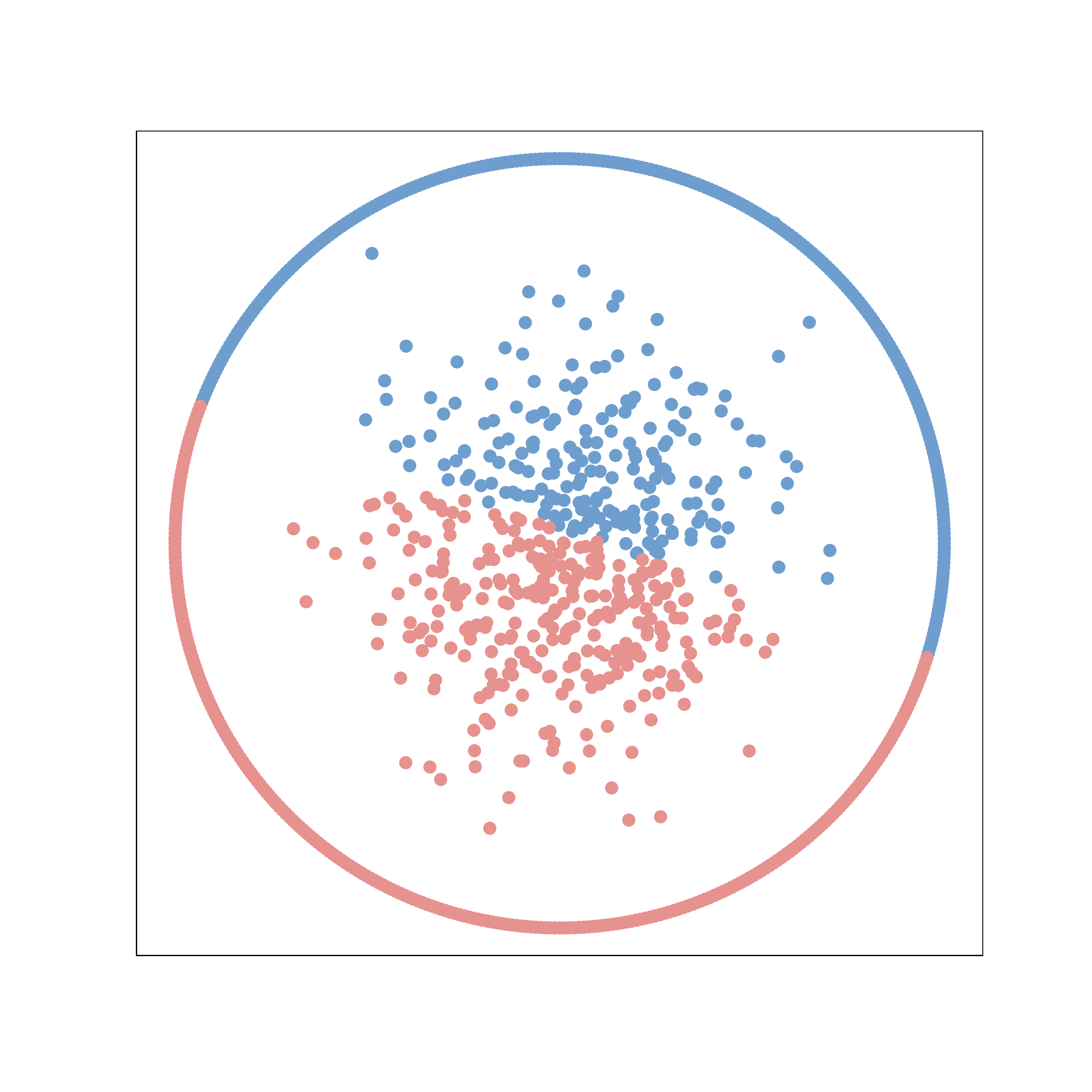}
		\label{fig: donut2 SC}
	}
	\hskip -30pt  
	\subfigure[DBSCAN]{
		\includegraphics[width=4.5cm]{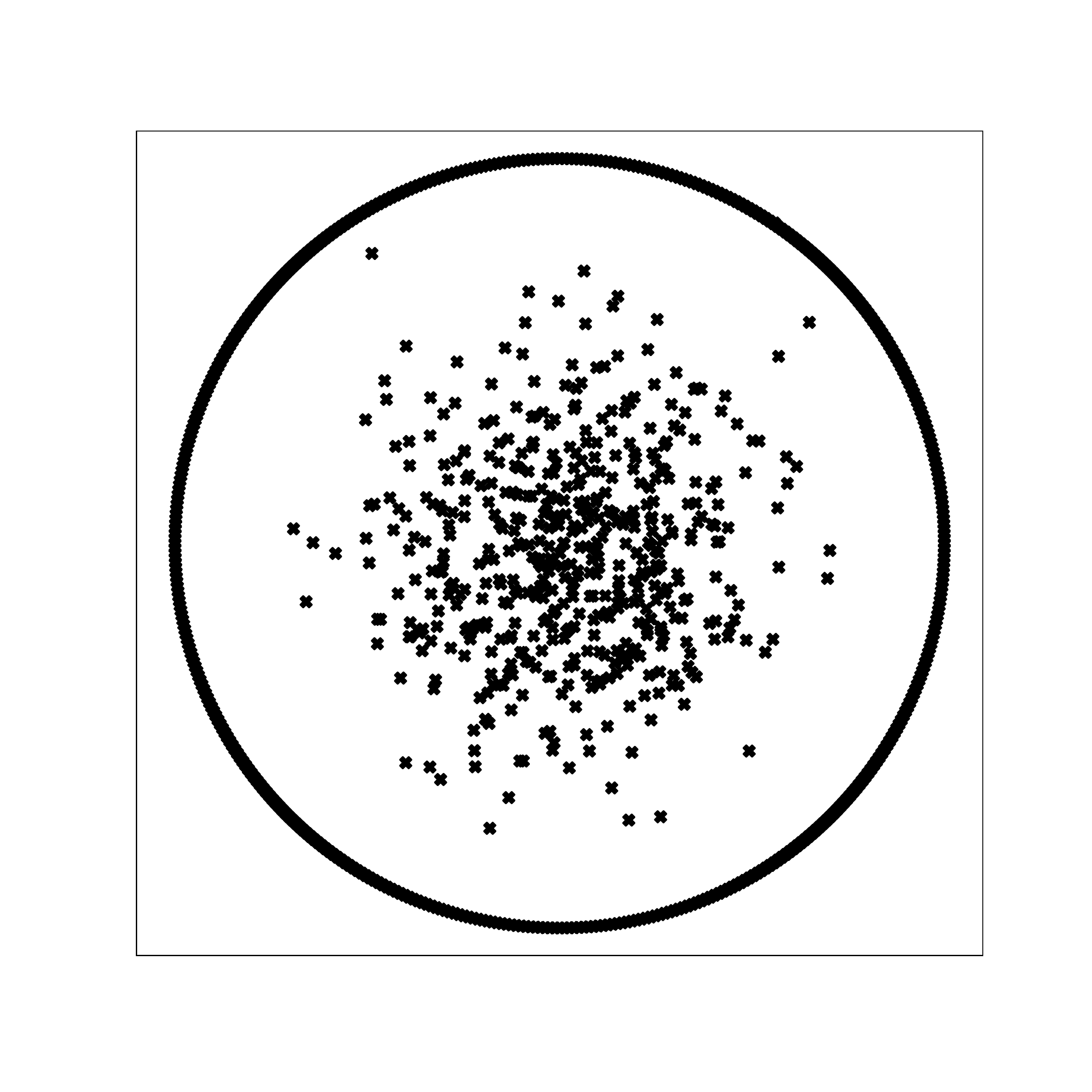}
		\label{fig: donut2 DBSCAN}
	}
	\vskip -20pt
	\subfigure[DPC]{
		\includegraphics[width=4.5cm]{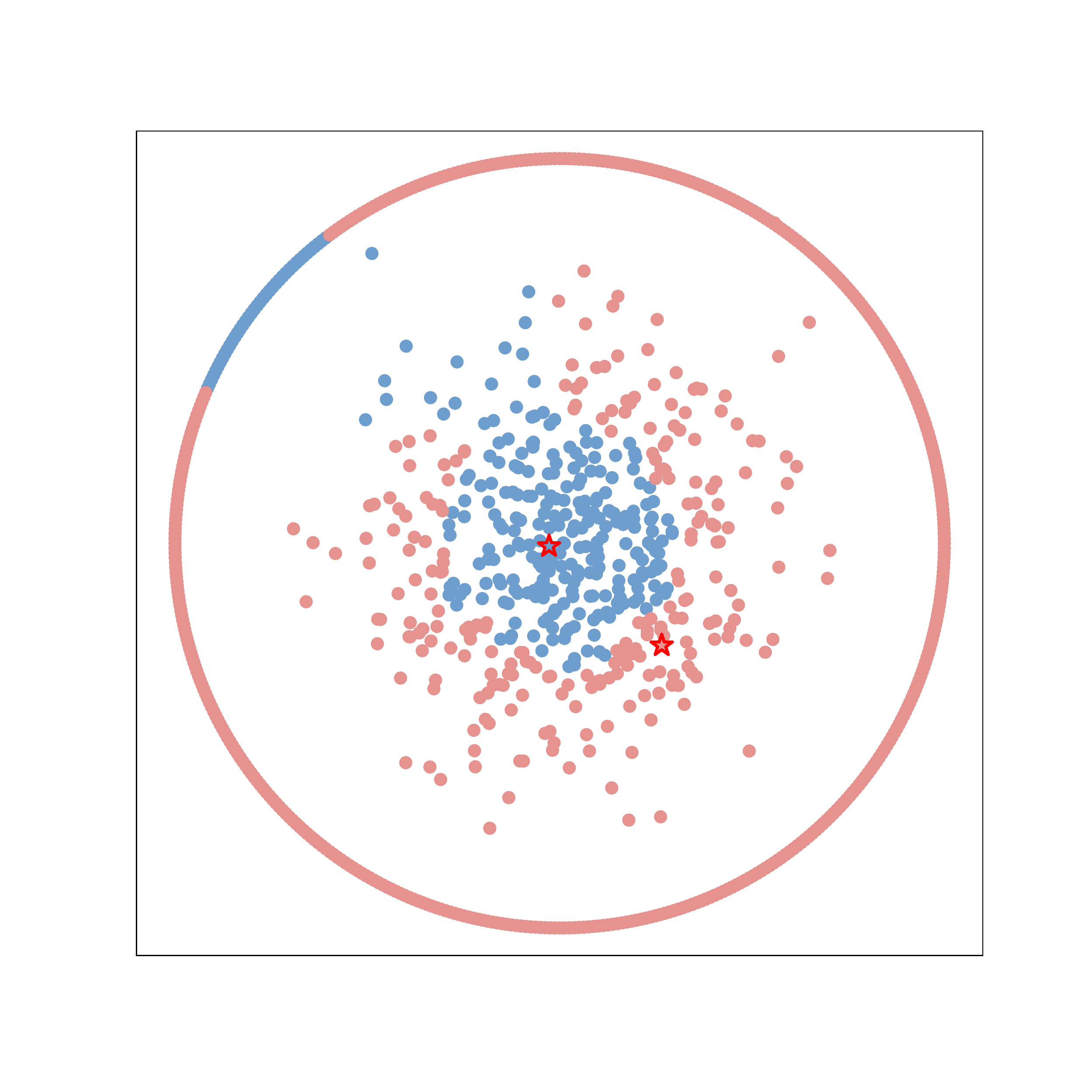}
		\label{fig: donut2 DPC}
	}
	\hskip -30pt
	\subfigure[DPC-KNN]{
		\includegraphics[width=4.5cm]{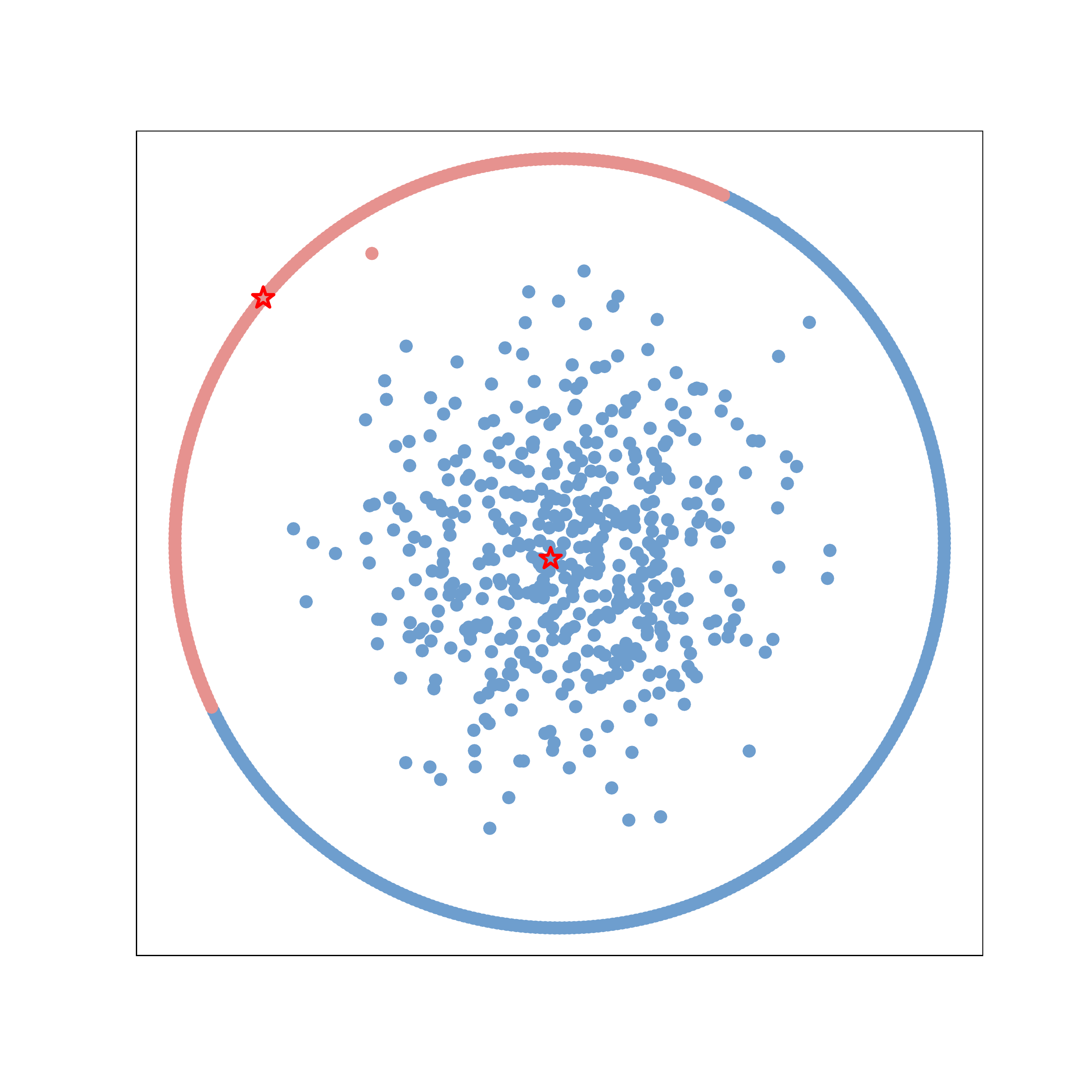}
		\label{fig: donut2 DPCKNN}
	}
	\hskip -30pt
	\subfigure[ECM]{
		\includegraphics[width=4.5cm]{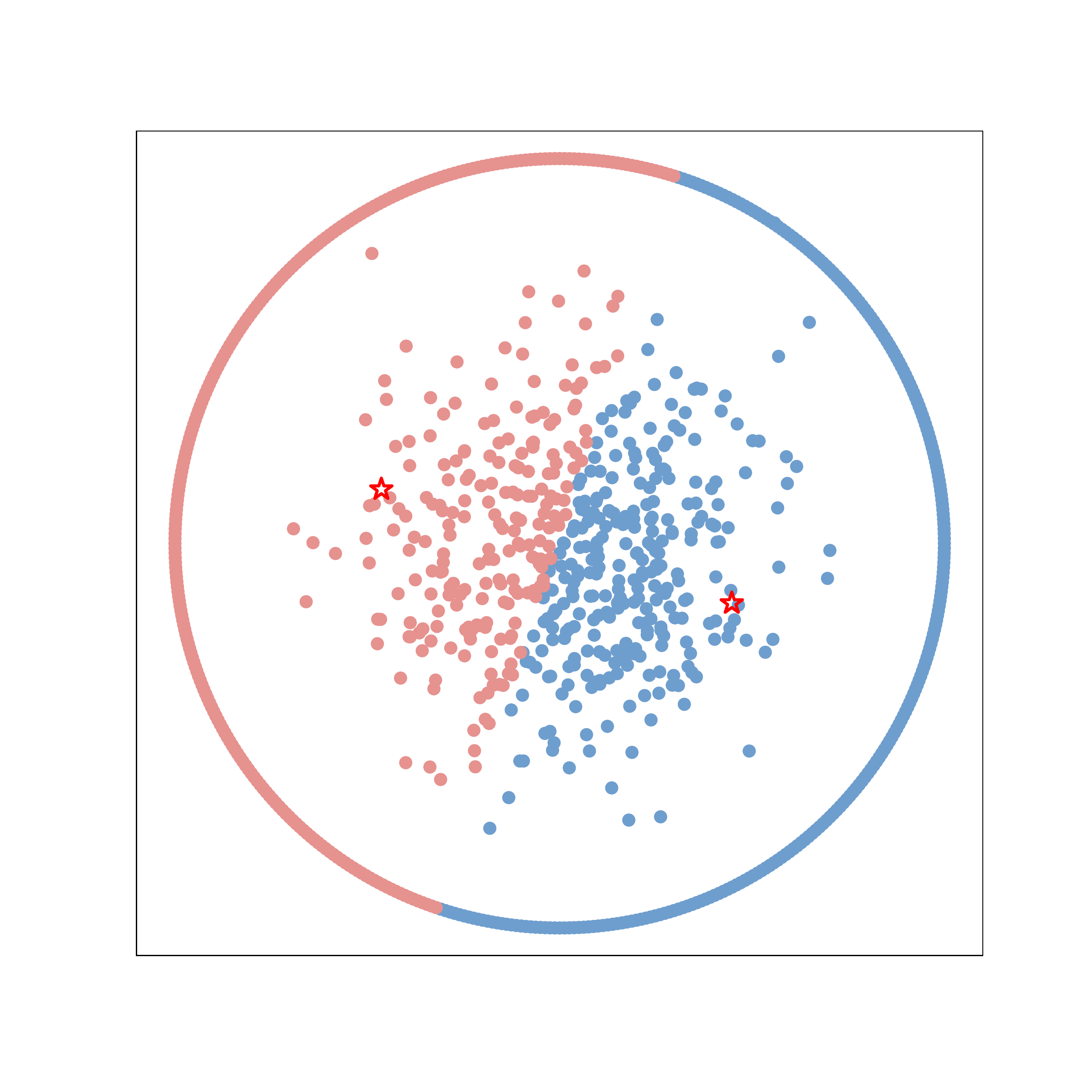}
		\label{fig: donut2 ECM}
	}
	\hskip -30pt
	\subfigure[GFDC]{
		\includegraphics[width=4.5cm]{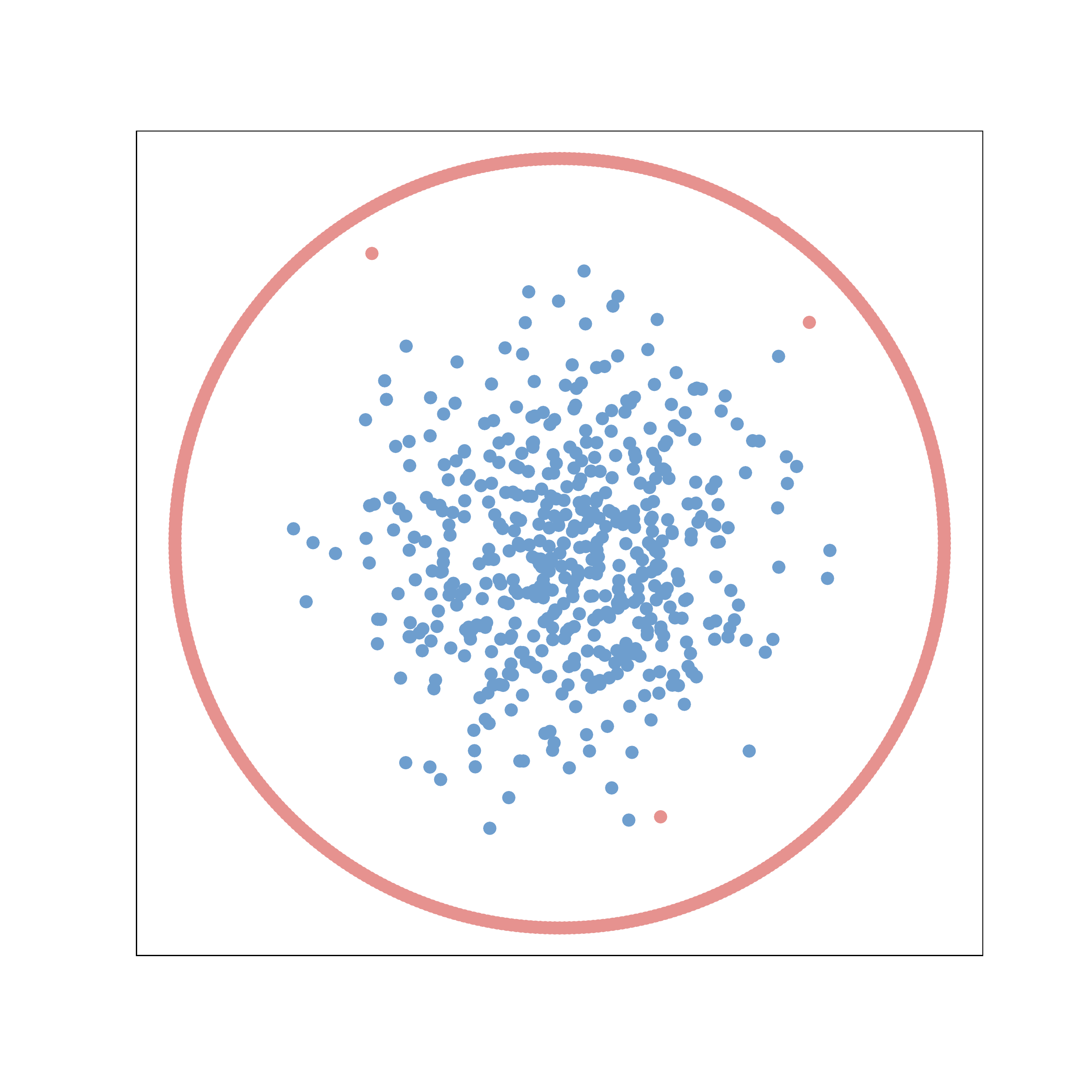}
		\label{fig: donut2 proposed}
	}
	\caption{Donut2.}
	\label{fig: donut2}
\end{figure}

\begin{figure}[H]\footnotesize
	\centering
	\vspace{-25pt}  
	\setlength{\abovecaptionskip}{0pt}  
	\subfigcapskip=-15pt  
	\subfigure[Ground truth]{
		\includegraphics[width=4.5cm]{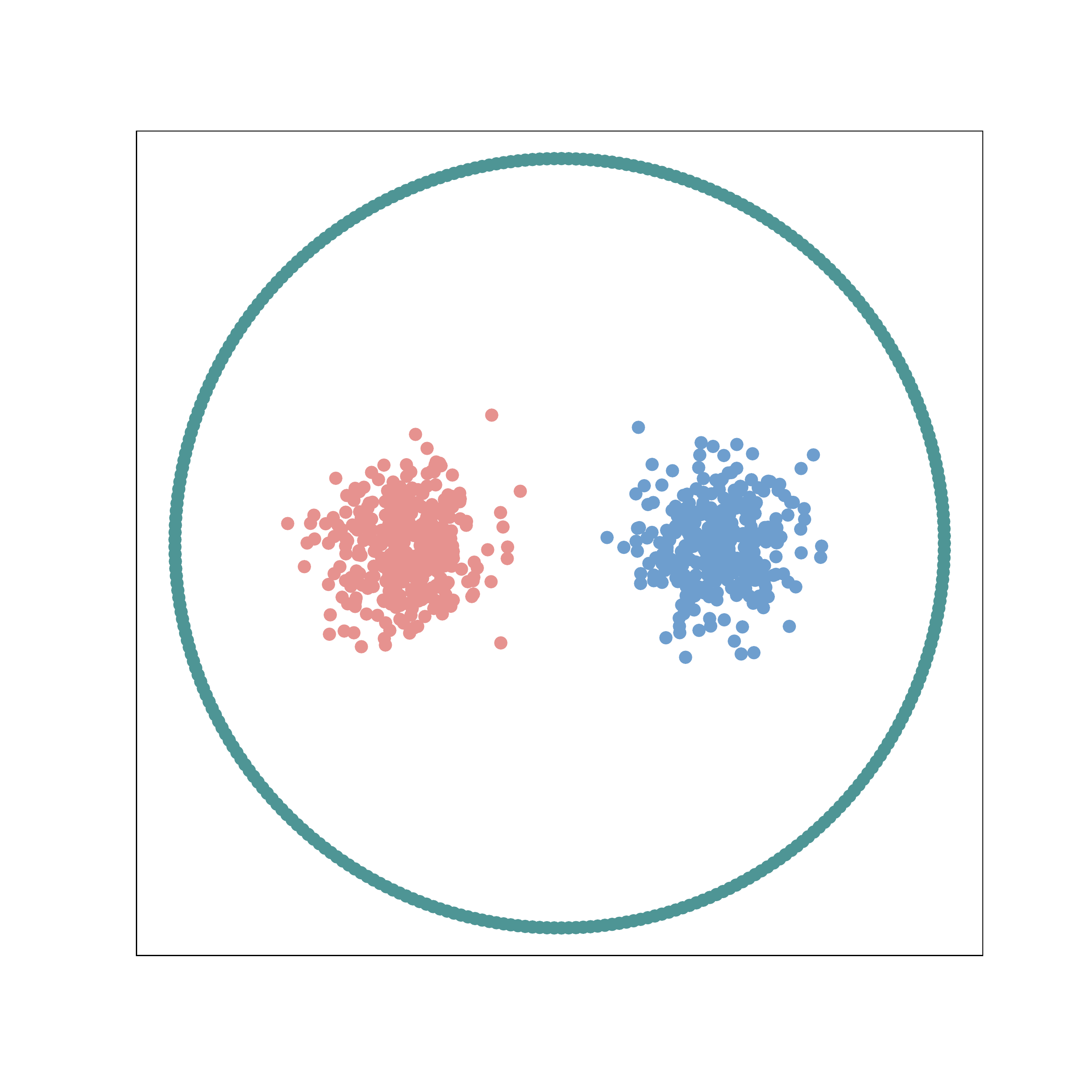}
		\label{fig: donut3 ground truth}
	}
	\hskip -30pt  
	\subfigure[$k$-means++]{
		\includegraphics[width=4.5cm]{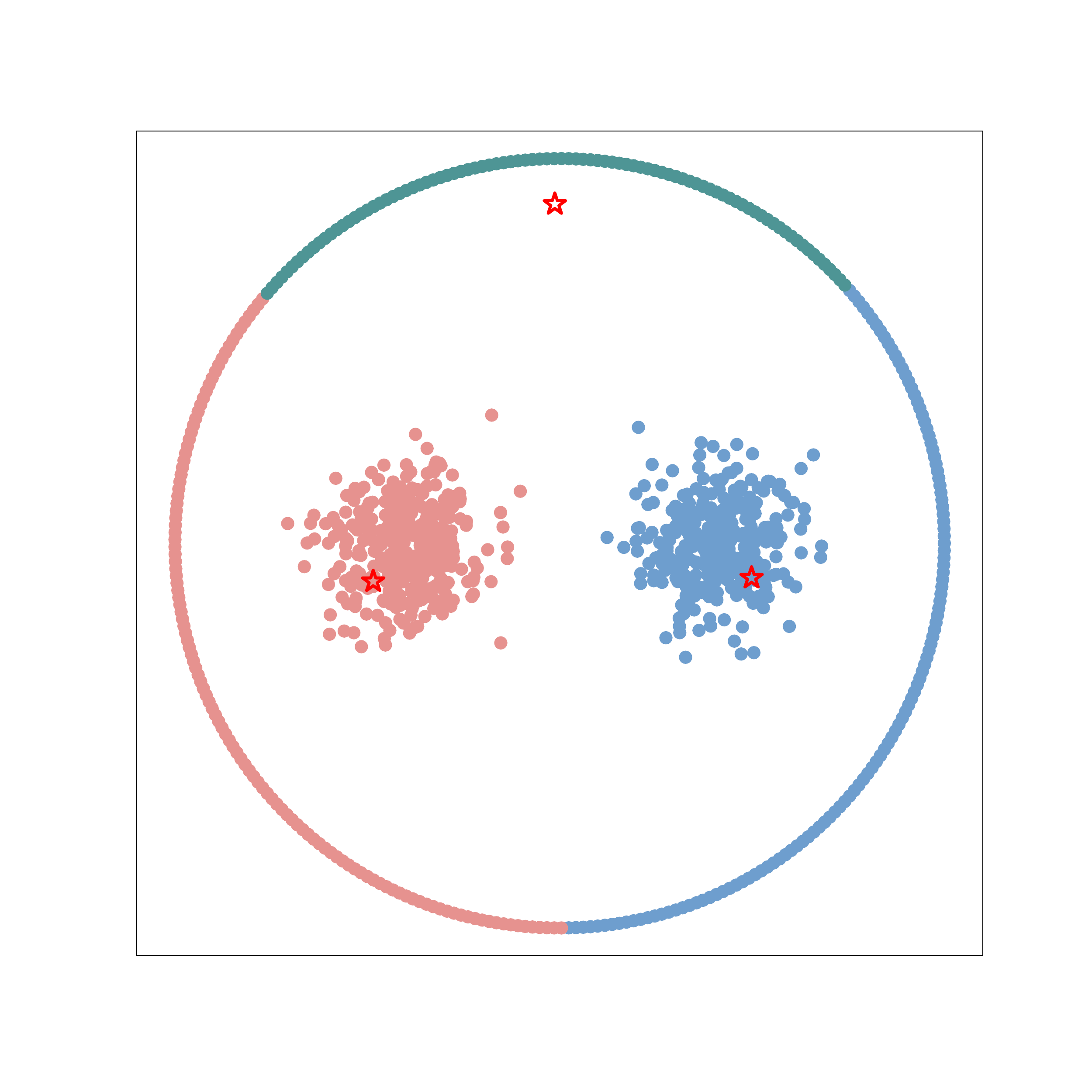}
		\label{fig: donut3 k_means}
	}
	\hskip -30pt  
	\subfigure[SC]{
		\includegraphics[width=4.5cm]{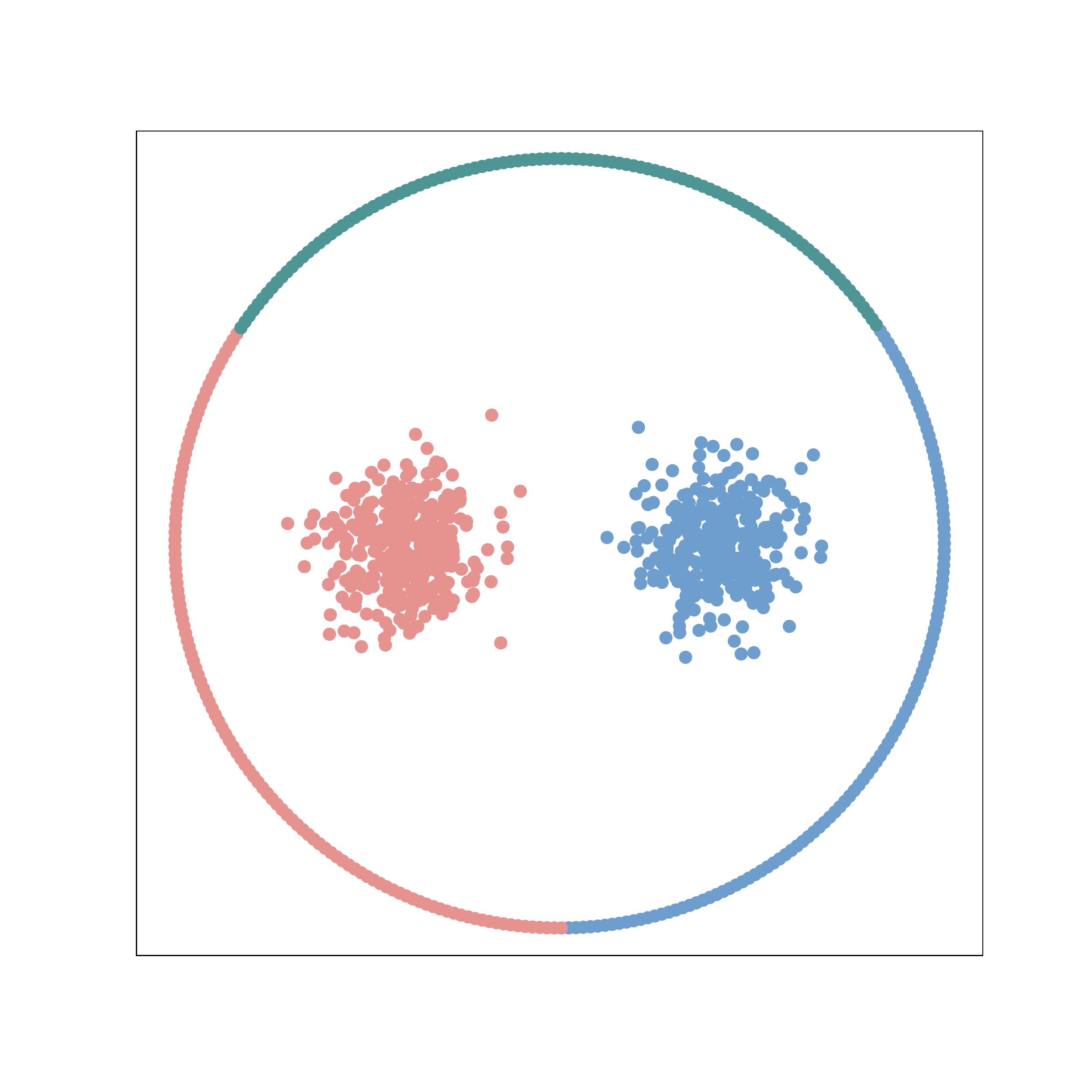}
		\label{fig: donut3 SC}
	}
	\hskip -30pt  
	\subfigure[DBSCAN]{
		\includegraphics[width=4.5cm]{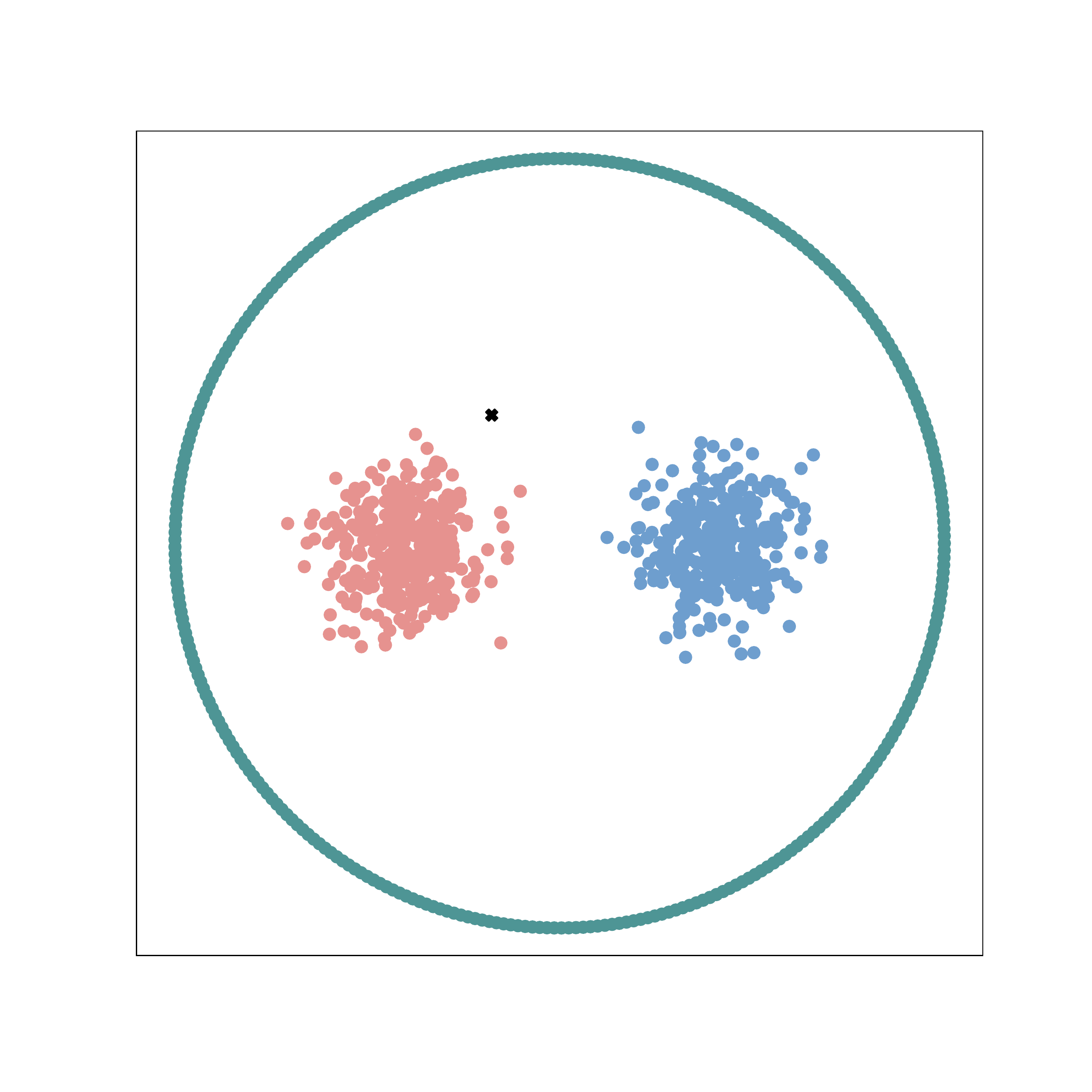}
		\label{fig: donut3 DBSCAN}
	}
	\vskip -20pt  
	\subfigure[DPC]{
		\includegraphics[width=4.5cm]{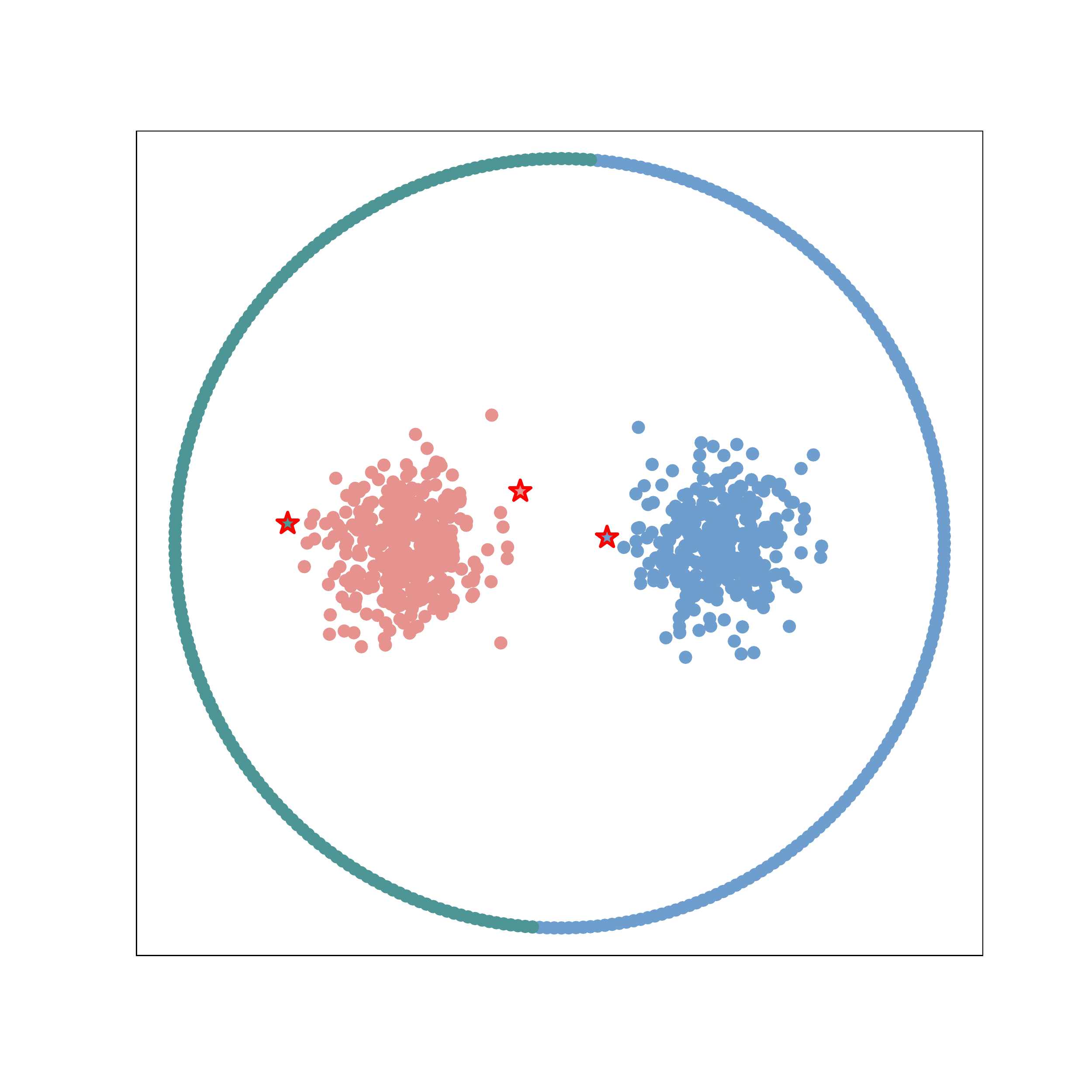}
		\label{fig: donut3 DPC}
	}
	\hskip -30pt  
	\subfigure[DPC-KNN]{
		\includegraphics[width=4.5cm]{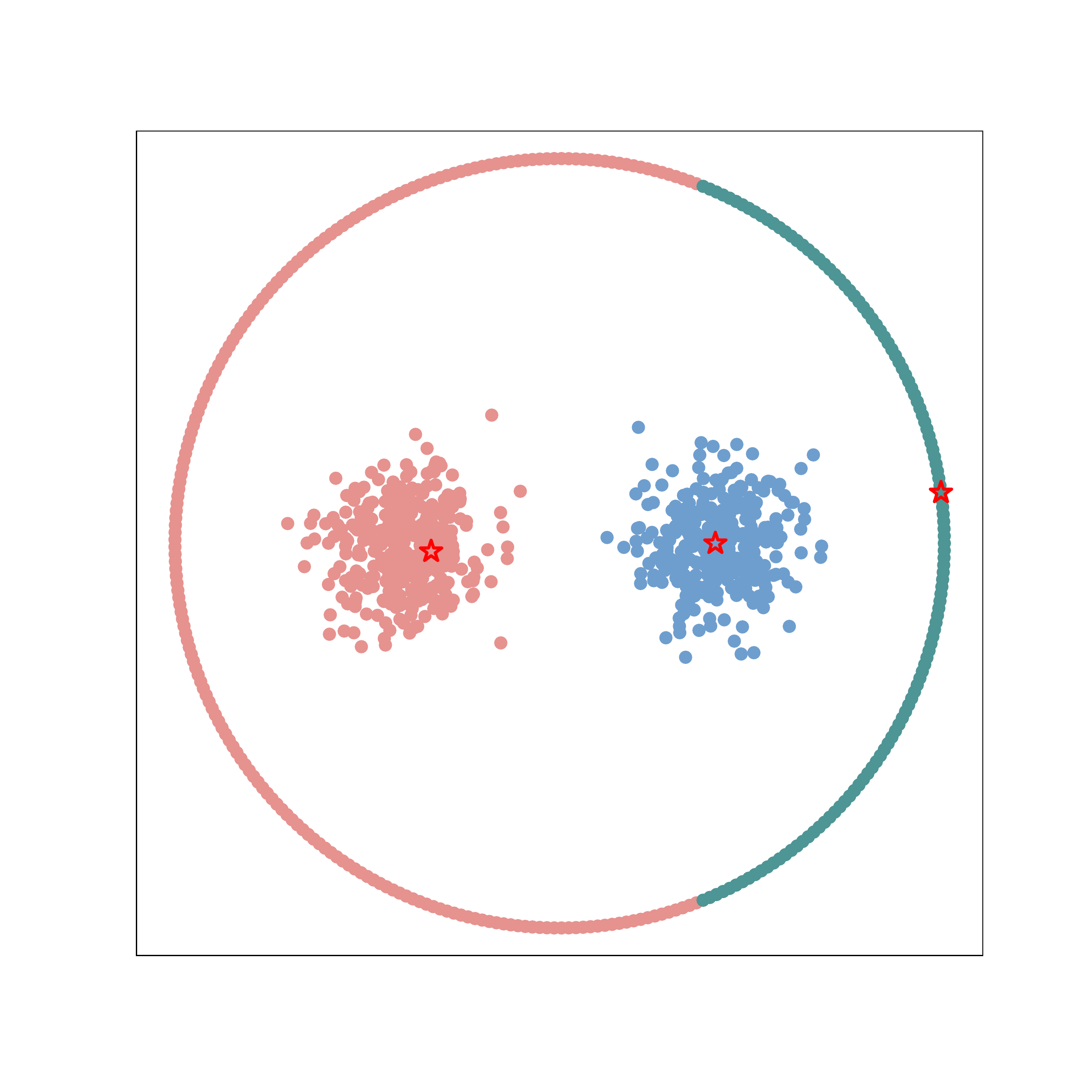}
		\label{fig: donut3 DPCKNN}
	}
	\hskip -30pt  
	\subfigure[ECM]{
		\includegraphics[width=4.5cm]{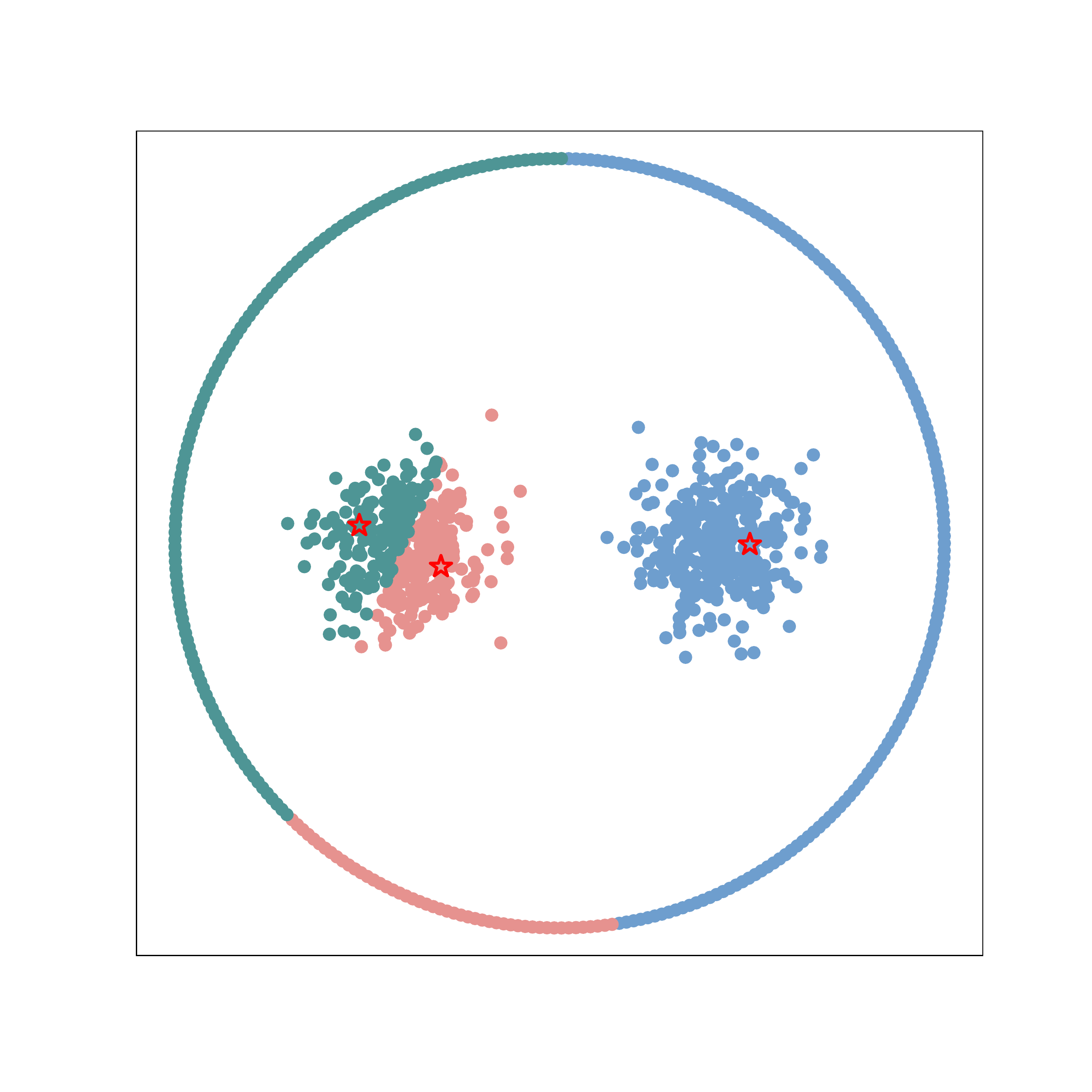}
		\label{fig: donut3 ECM}
	}
	\hskip -30pt  
	\subfigure[GFDC]{
		\includegraphics[width=4.5cm]{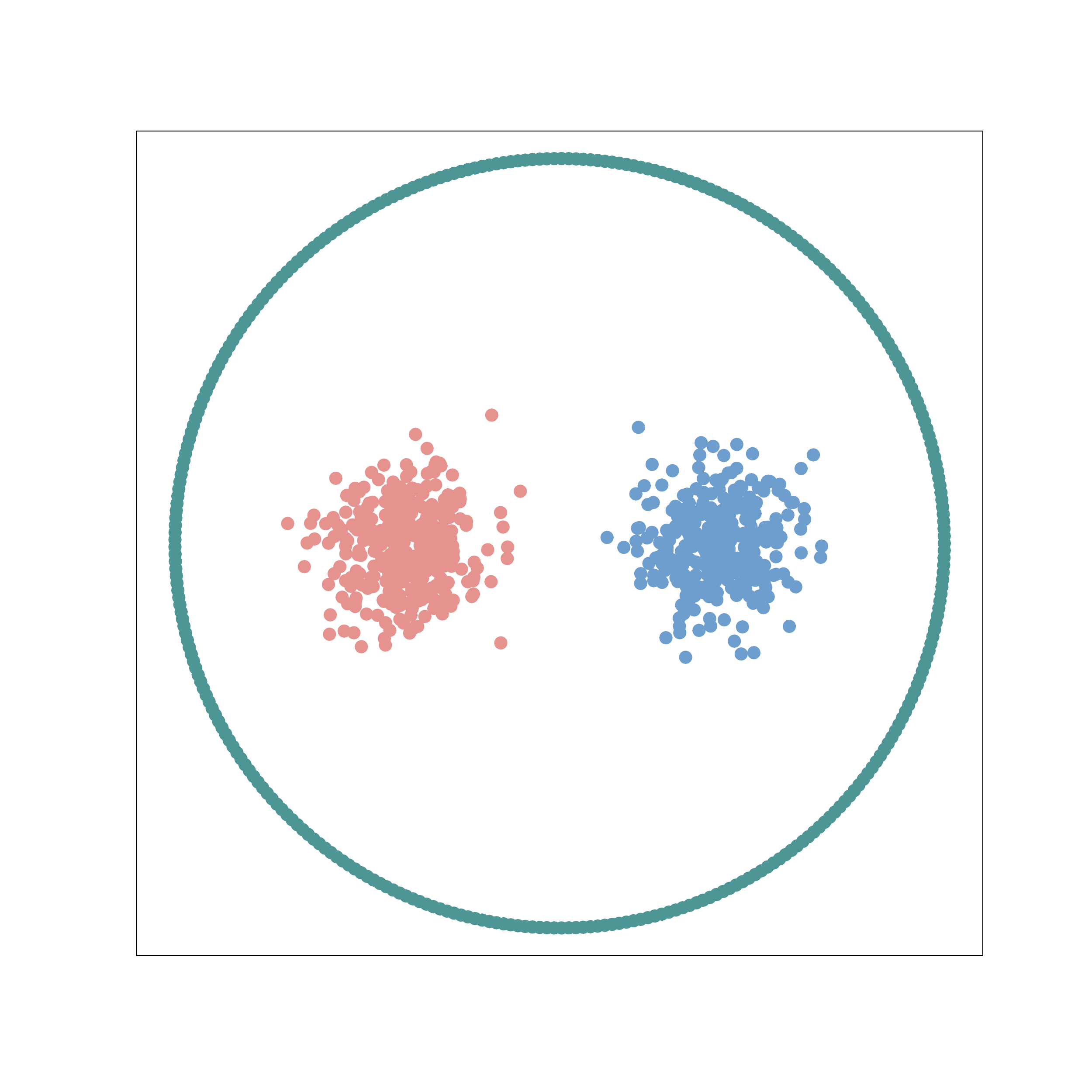}
		\label{fig: donut3 proposed}
	}
	\caption{Donut3.}
	\label{fig: donut3}
\end{figure}

\begin{figure}[htbp]\footnotesize
		\centering
		\vspace{-25pt}  
		\setlength{\abovecaptionskip}{0pt}  
		\subfigcapskip=-15pt  
		\subfigure[Ground truth]{
			\includegraphics[width=4.5cm]{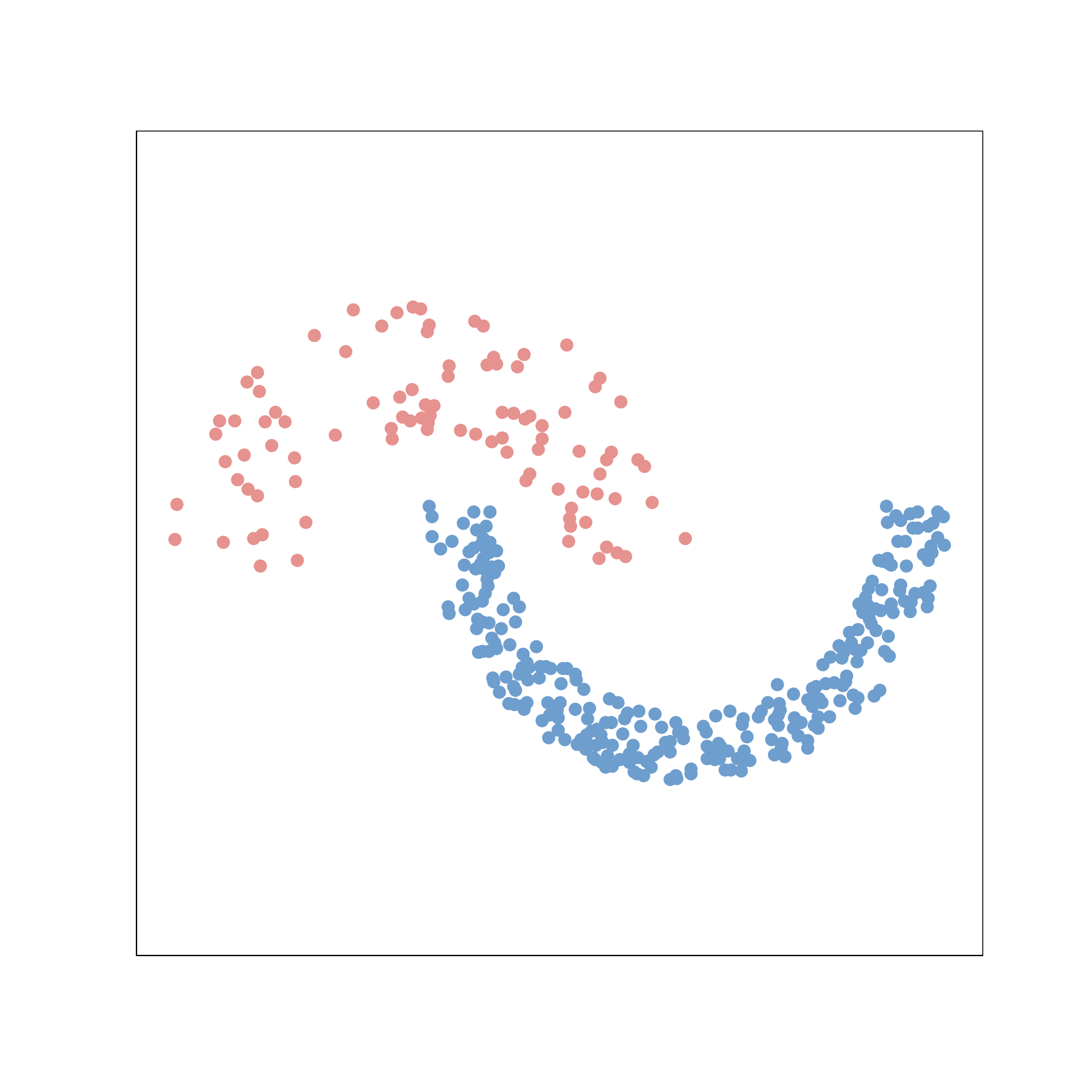}
			\label{fig: jain ground truth}
		}
		\hskip -30pt
		\subfigure[$k$-means++]{
			\includegraphics[width=4.5cm]{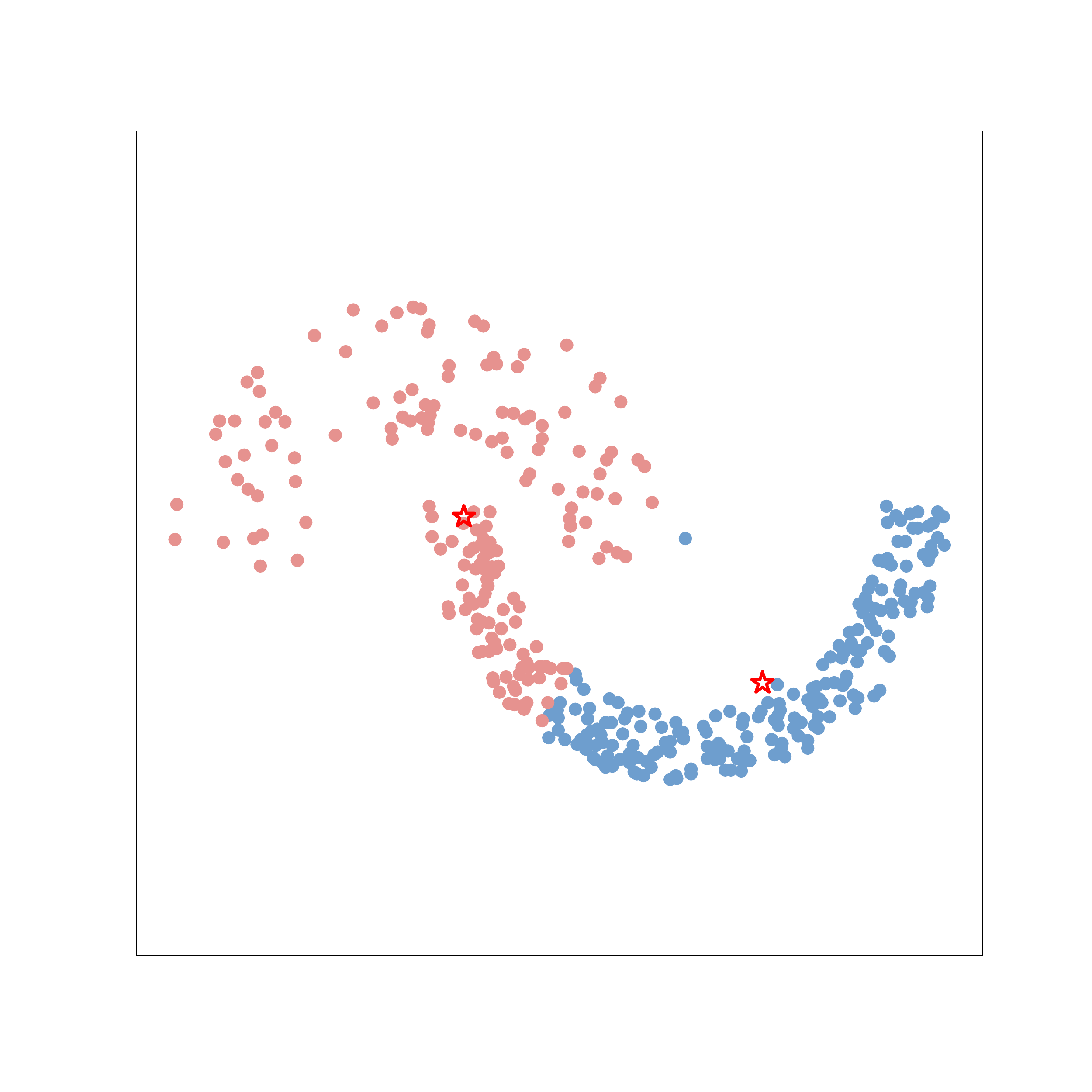}
			\label{fig: jain k_means}
		}
		\hskip -30pt
		\subfigure[SC]{
			\includegraphics[width=4.5cm]{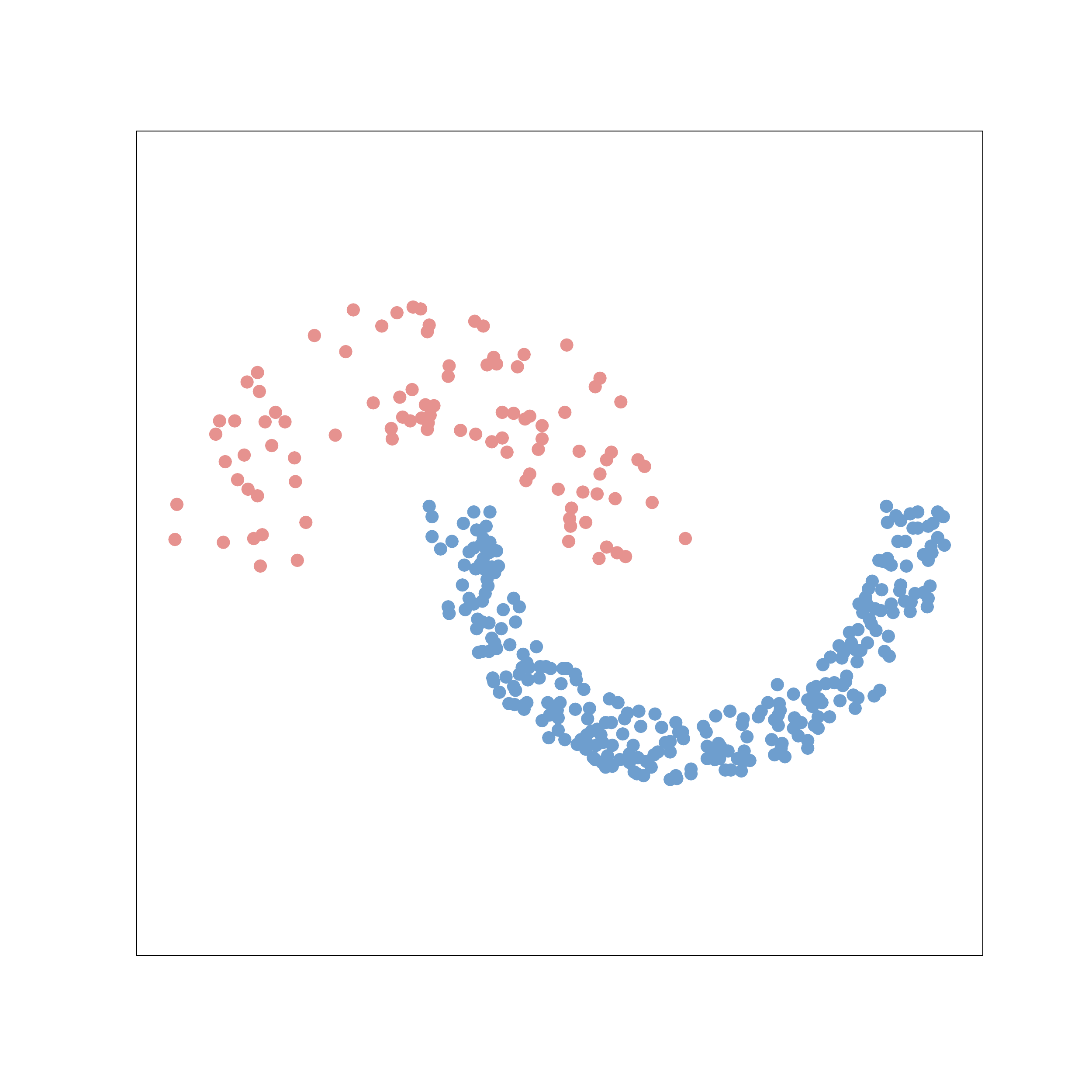}
			\label{fig: jain SC}
		}
		\hskip -30pt  
		\subfigure[DBSCAN]{
			\includegraphics[width=4.5cm]{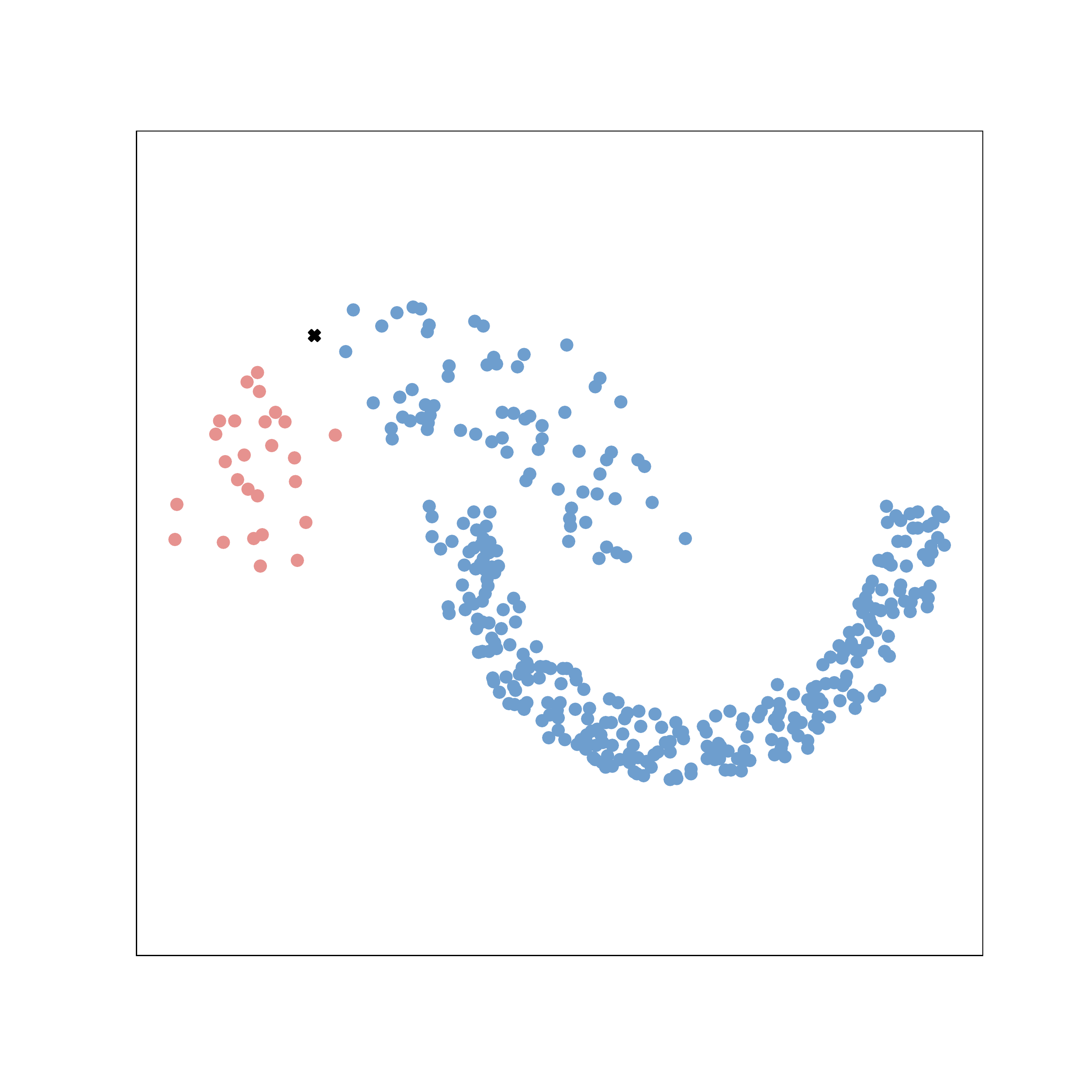}
			\label{fig: jain DBSCAN}
		}
		\vskip -20pt
		\subfigure[DPC]{
			\includegraphics[width=4.5cm]{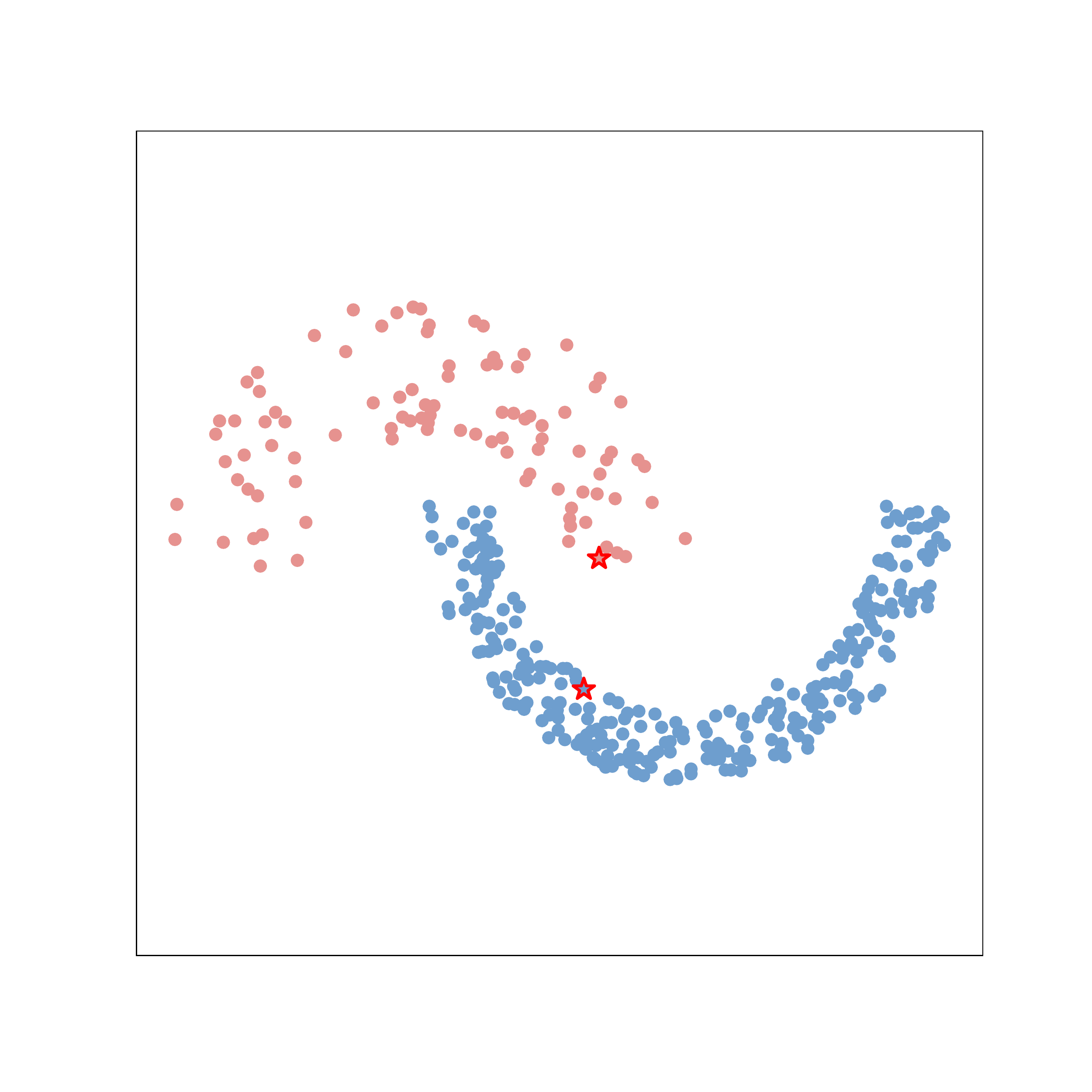}
			\label{fig: jain DPC}
		}
		\hskip -30pt
		\subfigure[DPC-KNN]{
			\includegraphics[width=4.5cm]{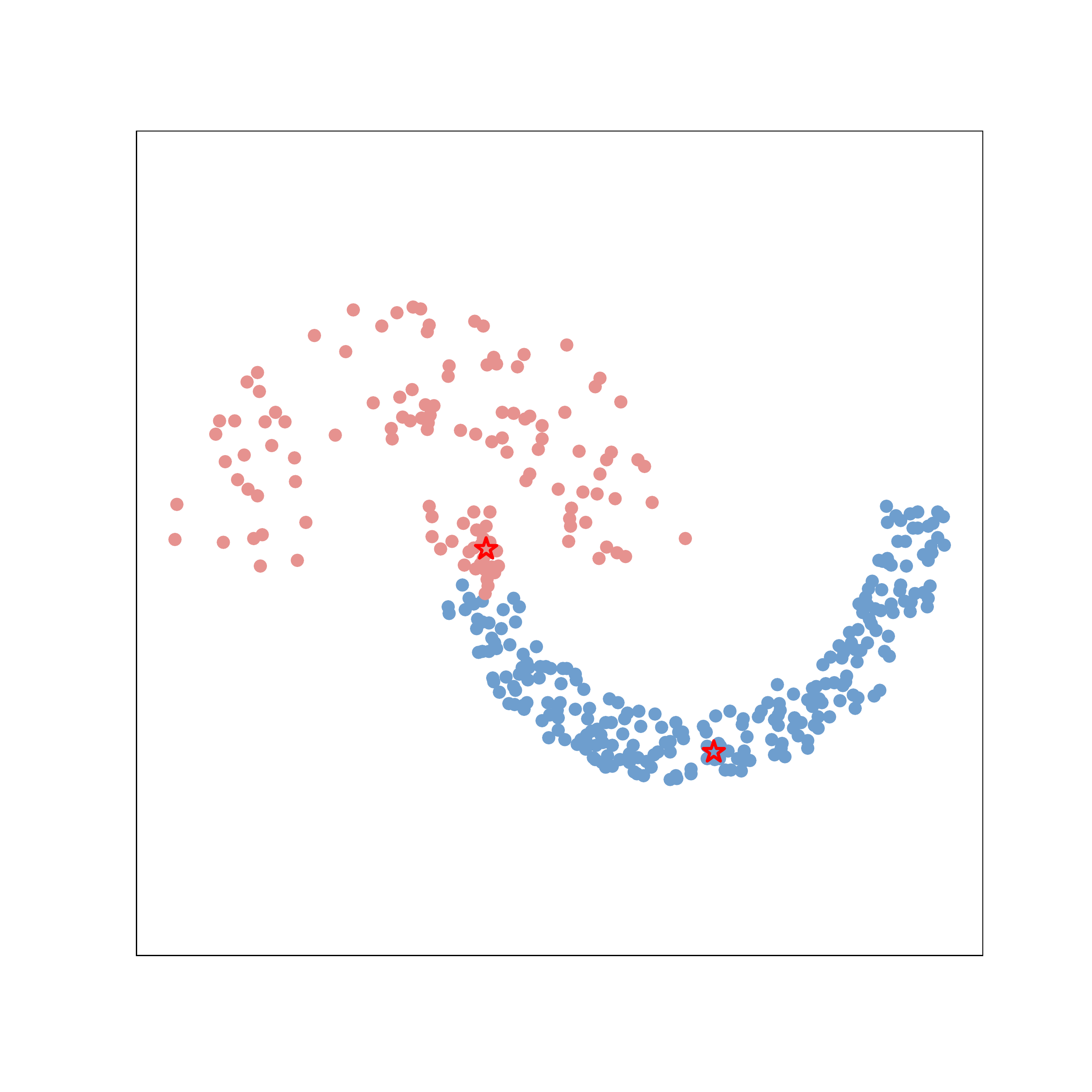}
			\label{fig: jain DPCKNN}
		}
		\hskip -30pt
		\subfigure[ECM]{
			\includegraphics[width=4.5cm]{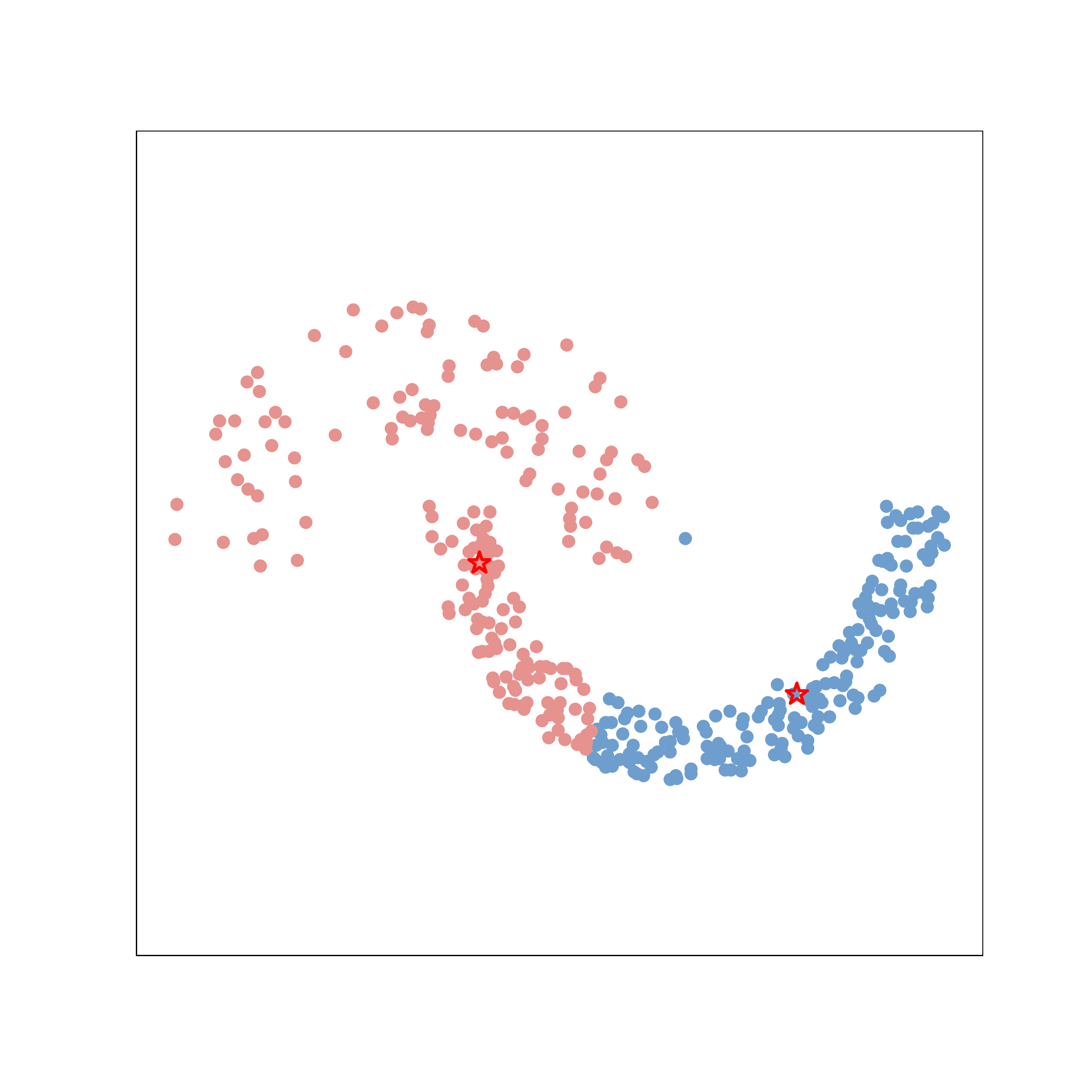}
			\label{fig: jain ECM}
		}
		\hskip -30pt
		\subfigure[GFDC]{
			\includegraphics[width=4.5cm]{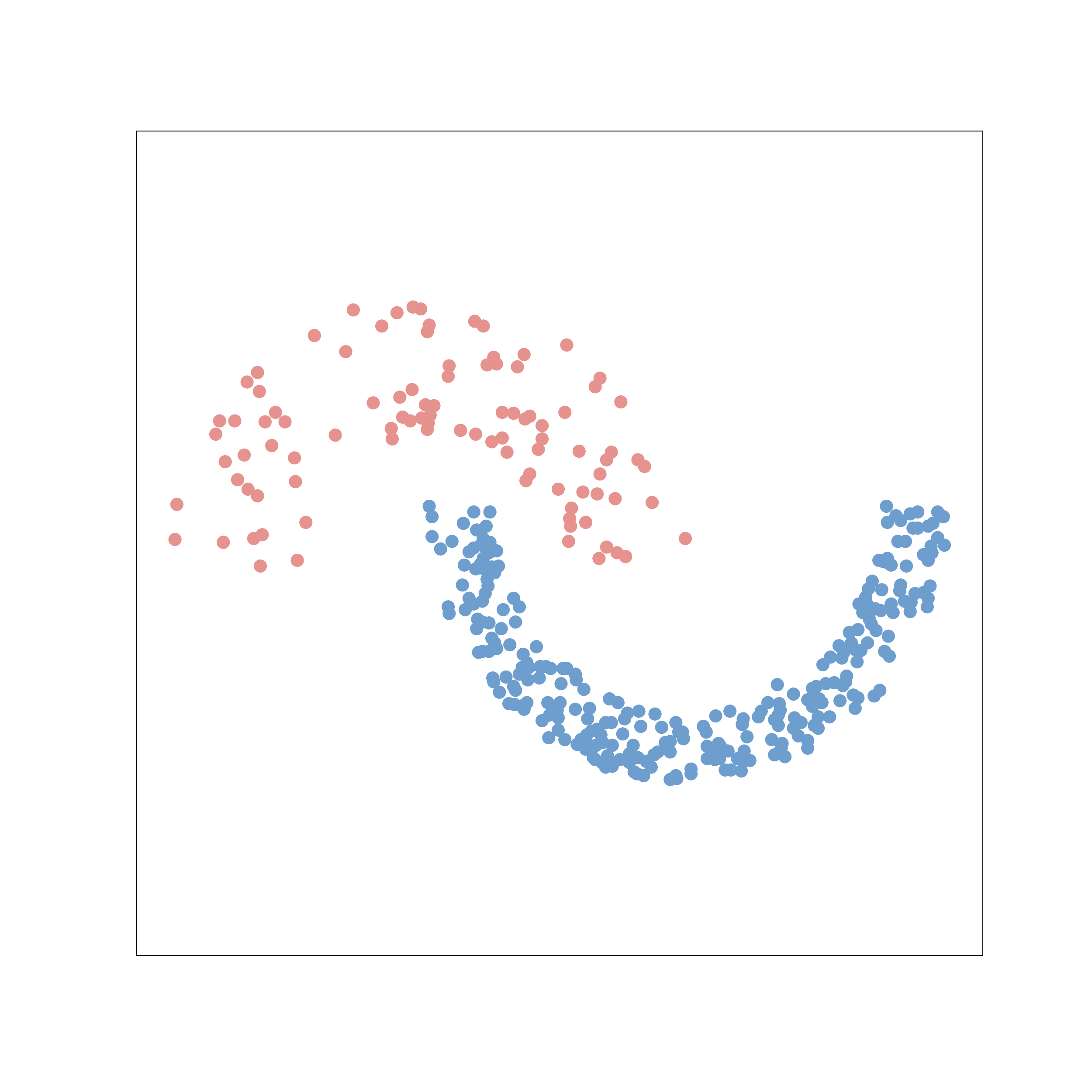}
			\label{fig: jain proposed}
		}
		\caption{Jain.}
		\label{fig: jain}
	\end{figure}

\begin{figure}[H]\footnotesize
		\centering
		\vspace{-25pt}  
		\setlength{\abovecaptionskip}{0pt}  
		\subfigcapskip=-15pt  
		\subfigure[Ground truth]{
			\includegraphics[width=4.5cm]{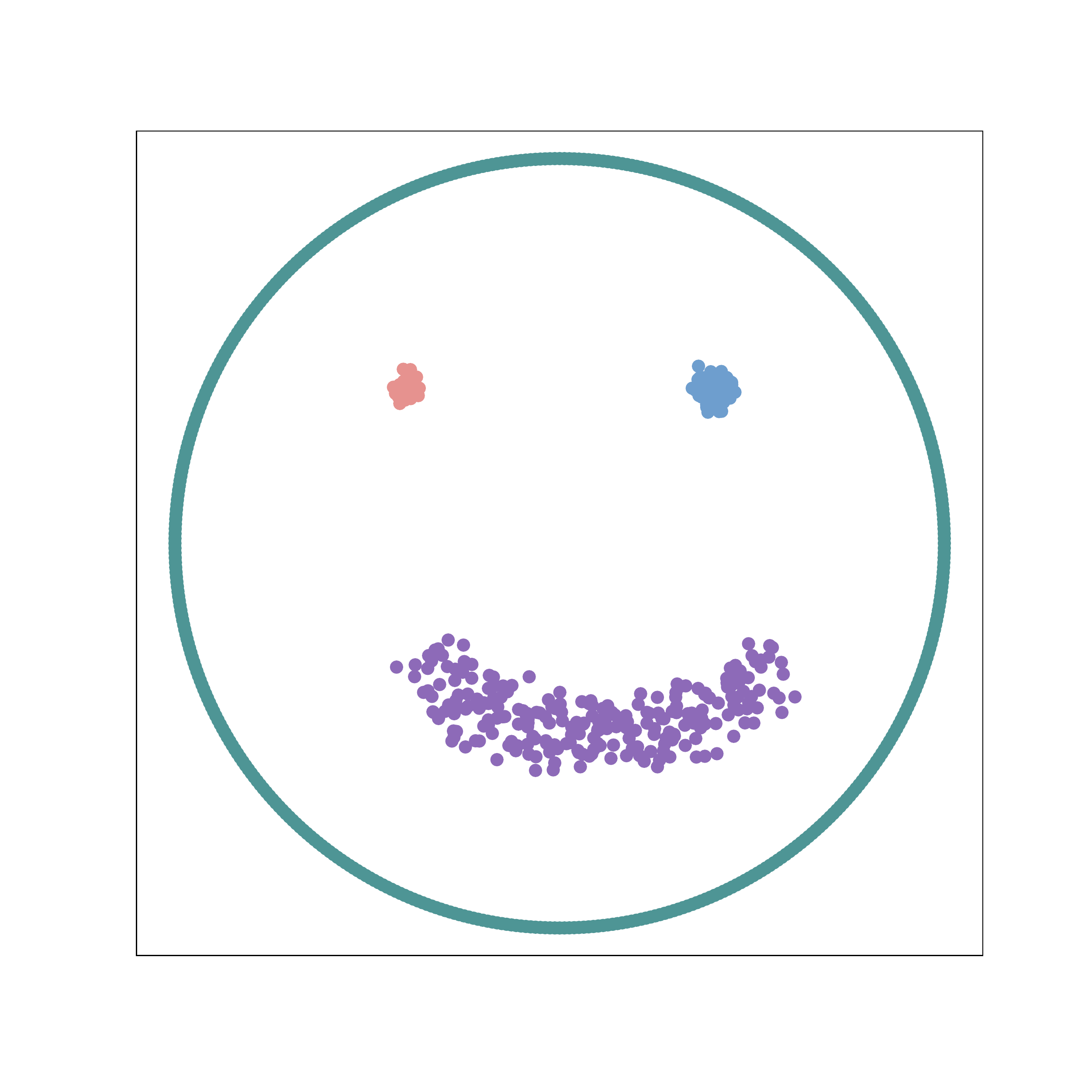}
			\label{fig: smile3 ground truth}
		}
		\hskip -30pt
		\subfigure[$k$-means++]{
			\includegraphics[width=4.5cm]{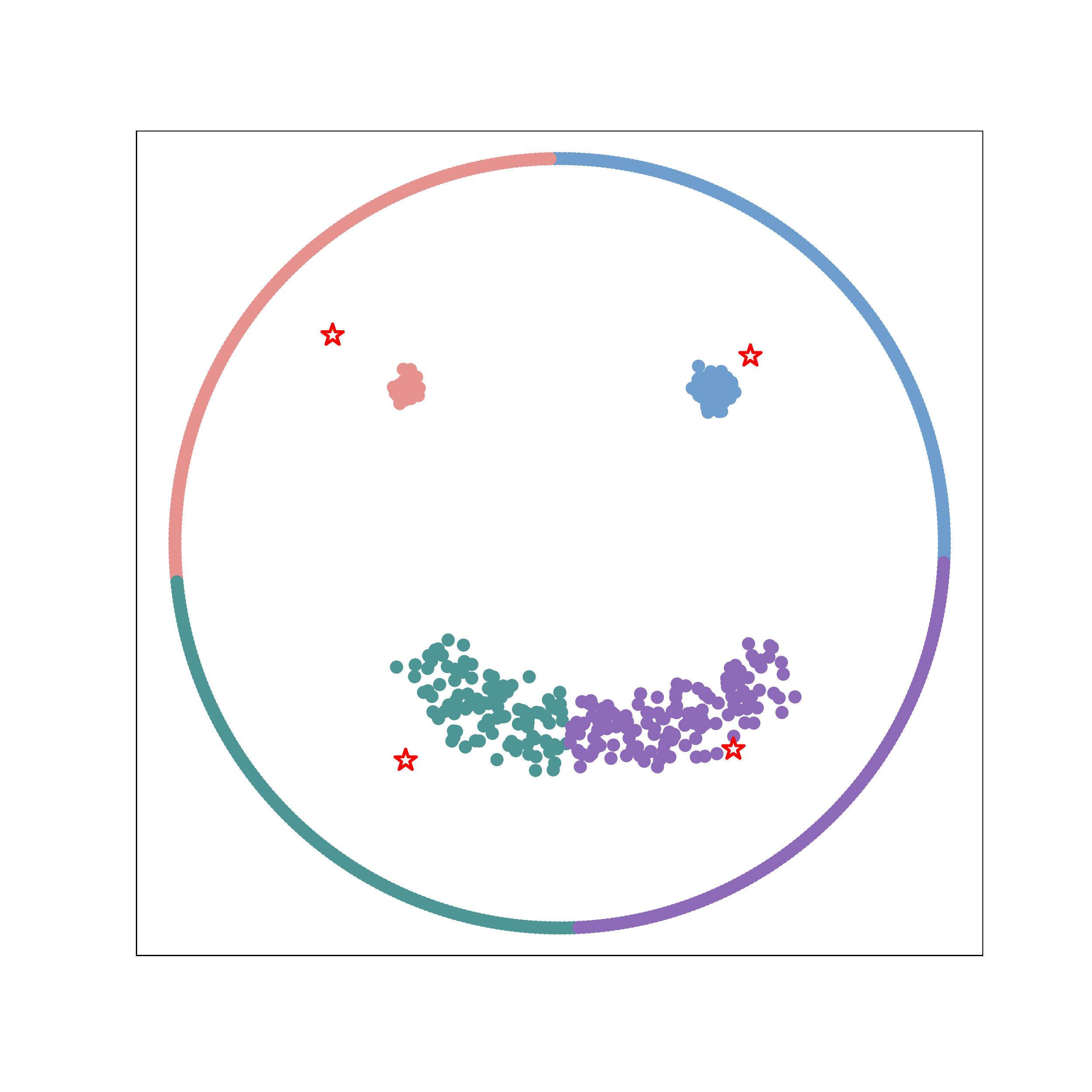}
			\label{fig: smile3 k_means}
		}
		\hskip -30pt
		\subfigure[SC]{
			\includegraphics[width=4.5cm]{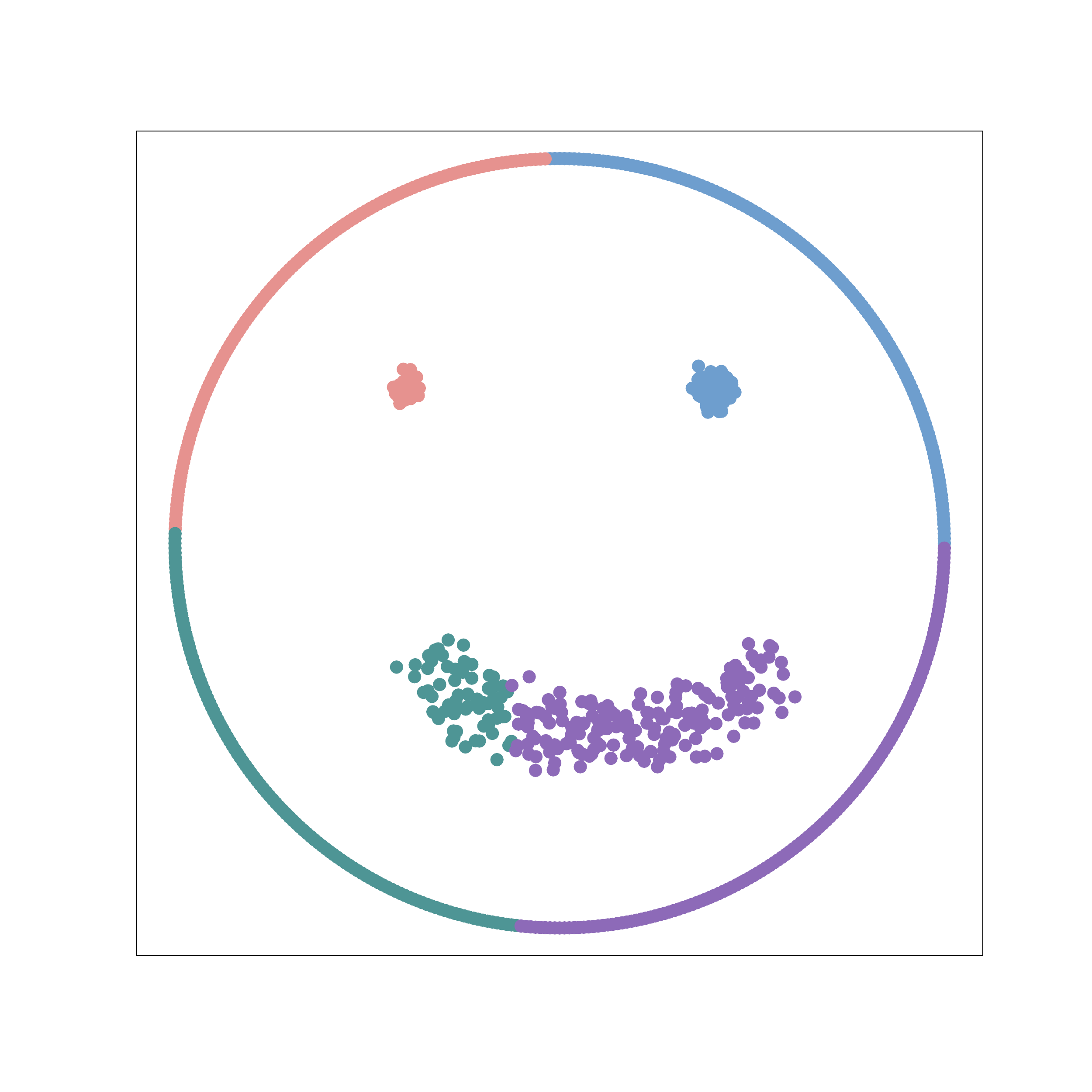}
			\label{fig: smile3 SC}
		}
		\hskip -30pt  
		\subfigure[DBSCAN]{
			\includegraphics[width=4.5cm]{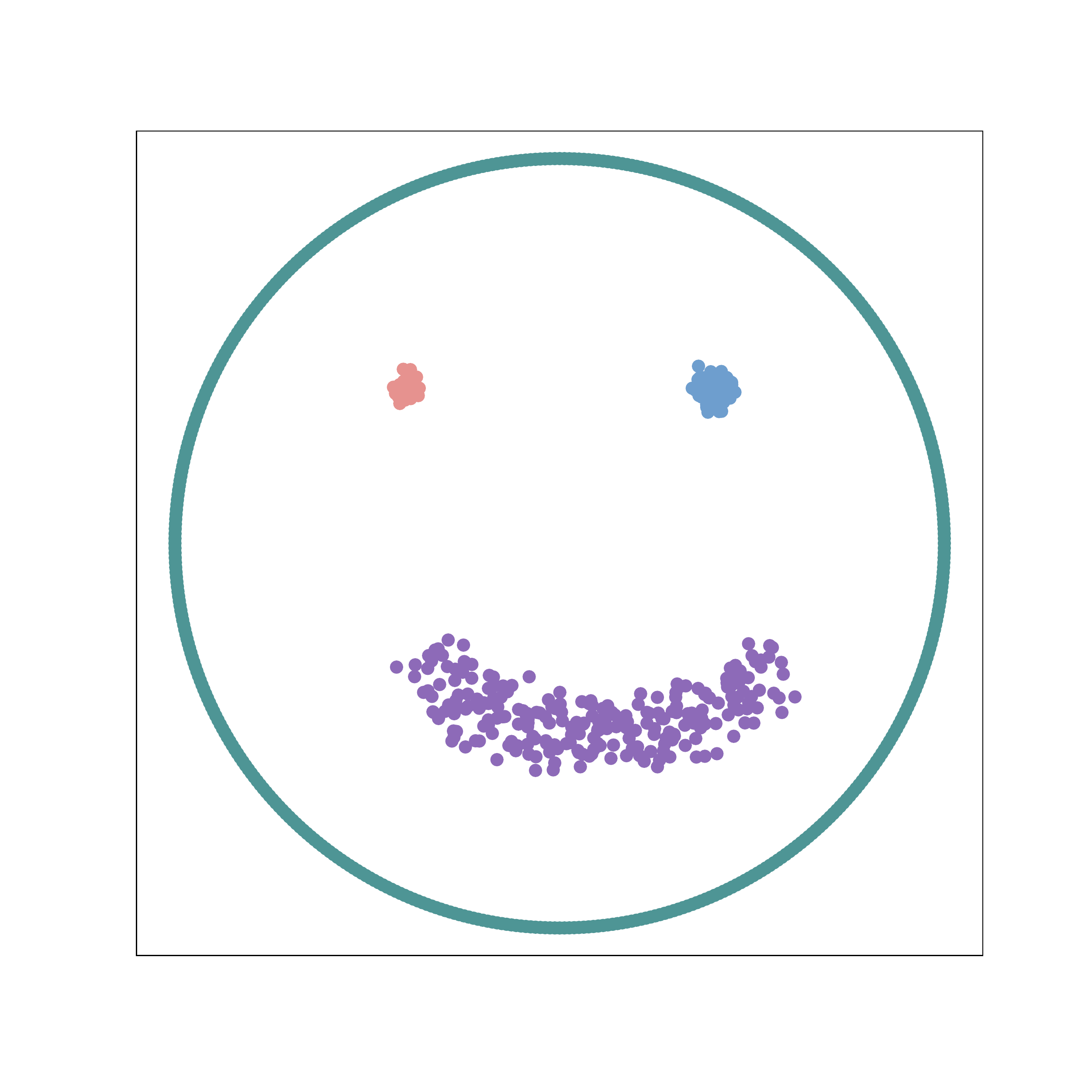}
			\label{fig: smile3 DBSCAN}
		}
		\vskip -20pt
		\subfigure[DPC]{
			\includegraphics[width=4.5cm]{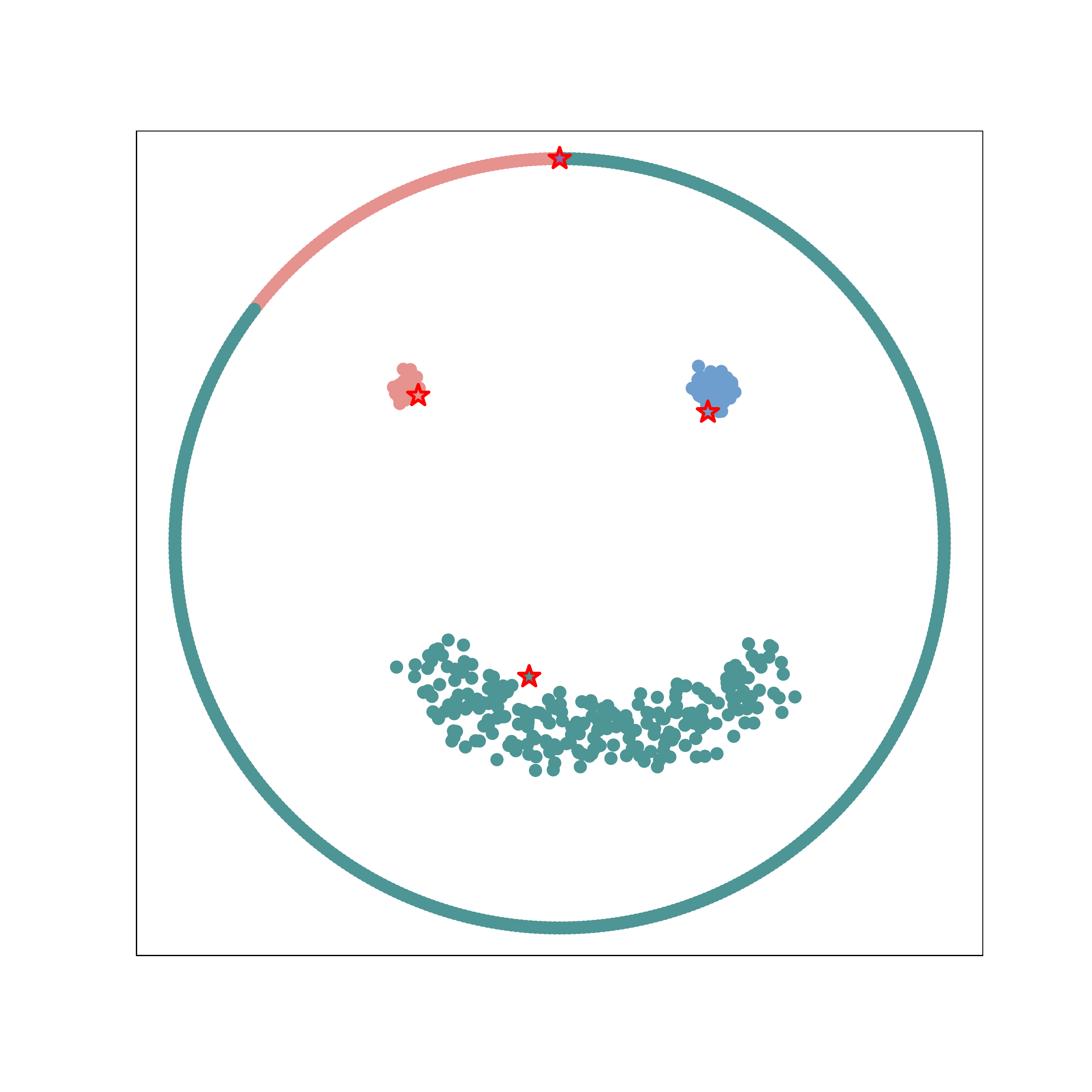}
			\label{fig: smile3 DPC}
		}
		\hskip -30pt
		\subfigure[DPC-KNN]{
			\includegraphics[width=4.5cm]{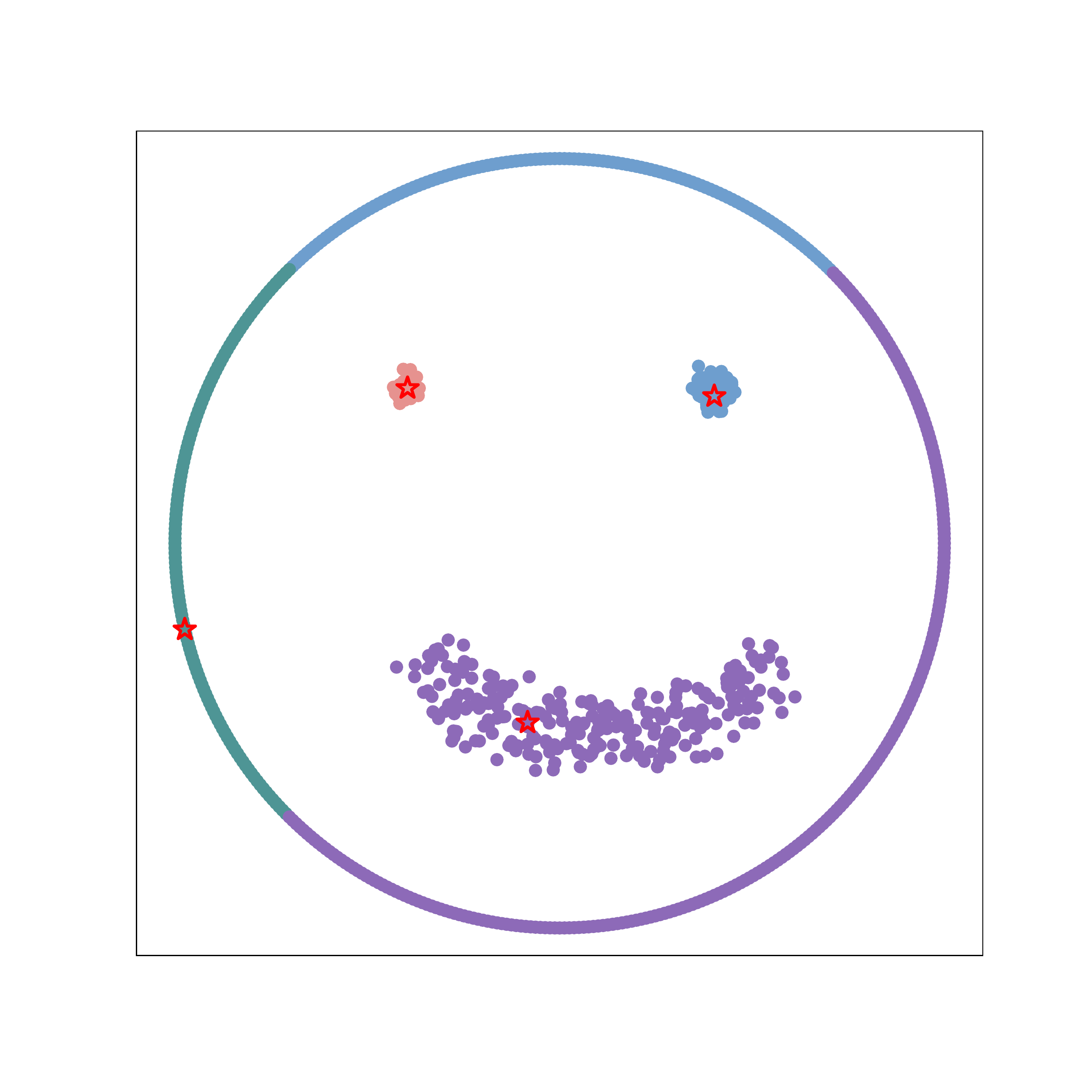}
			\label{fig: smile3 DPCKNN}
		}
		\hskip -30pt
		\subfigure[ECM]{
			\includegraphics[width=4.5cm]{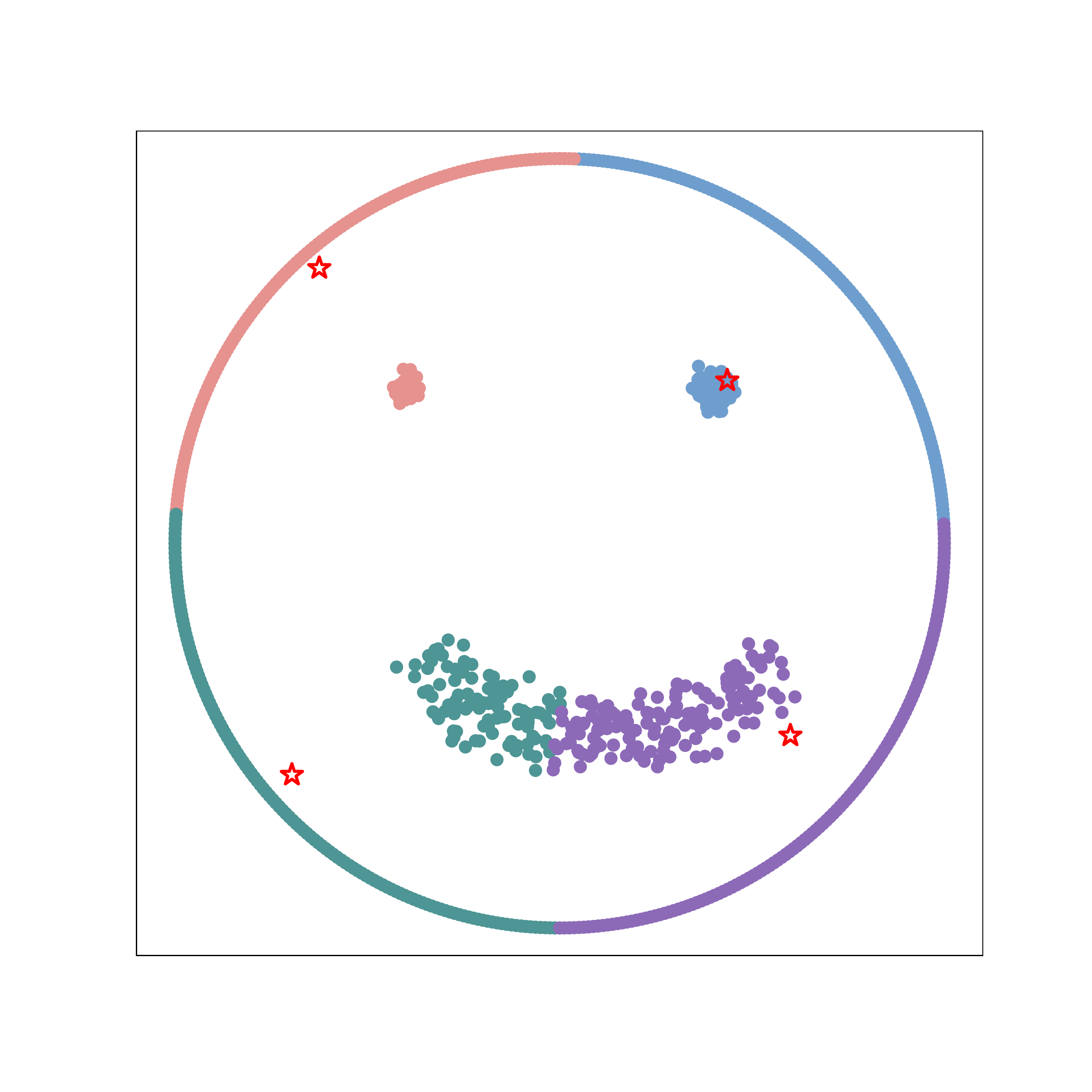}
			\label{fig: smile3 ECM}
		}
		\hskip -30pt
		\subfigure[GFDC]{
			\includegraphics[width=4.5cm]{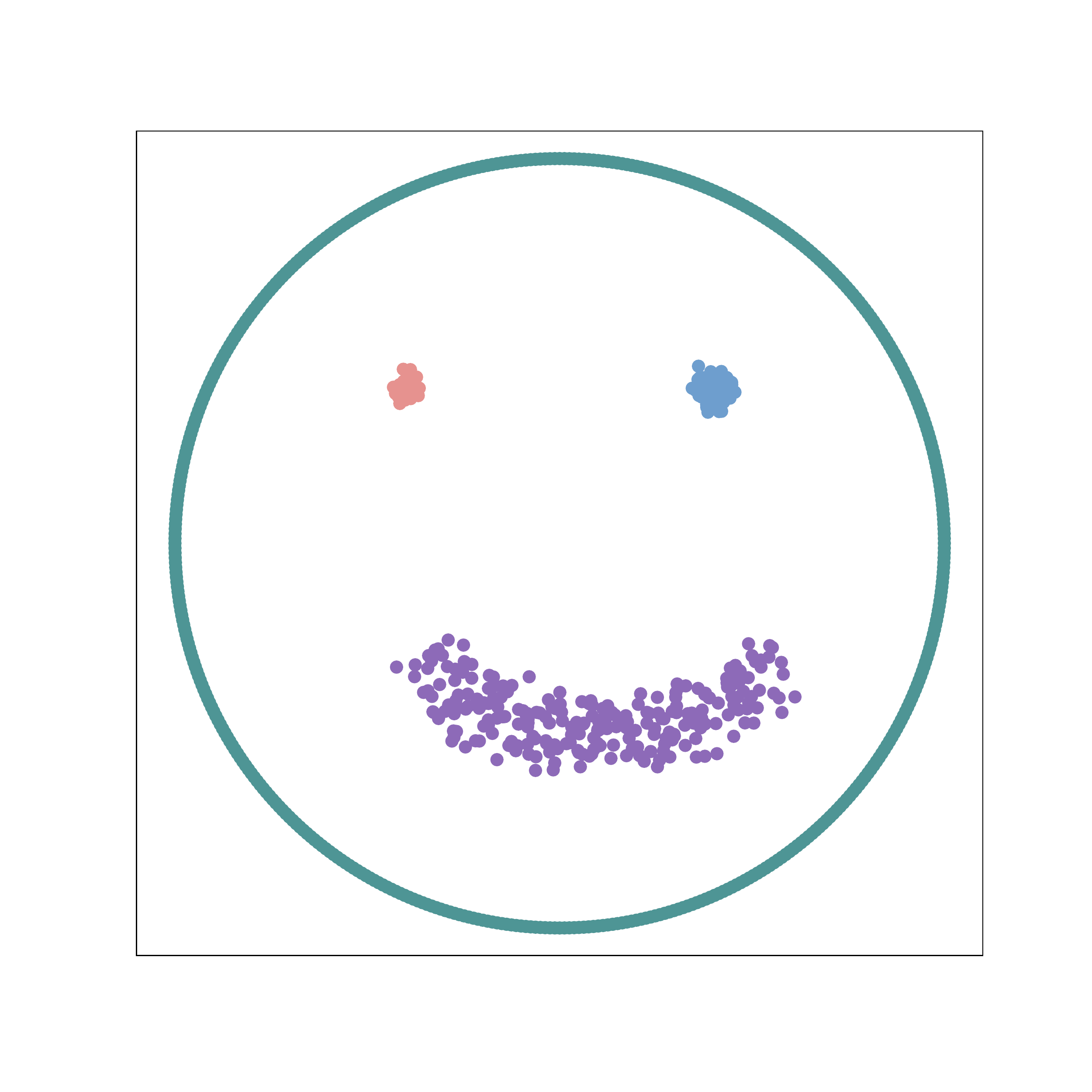}
			\label{fig: smile3 proposed}
		}
		\caption{Smile3.}
		\label{fig: smile3}
	\end{figure}

\begin{figure}[htbp]\footnotesize
		\centering
		\vspace{-25pt}  
		\setlength{\abovecaptionskip}{0pt}  
		\subfigcapskip=-15pt  
		\subfigure[Ground truth]{
			\includegraphics[width=4.5cm]{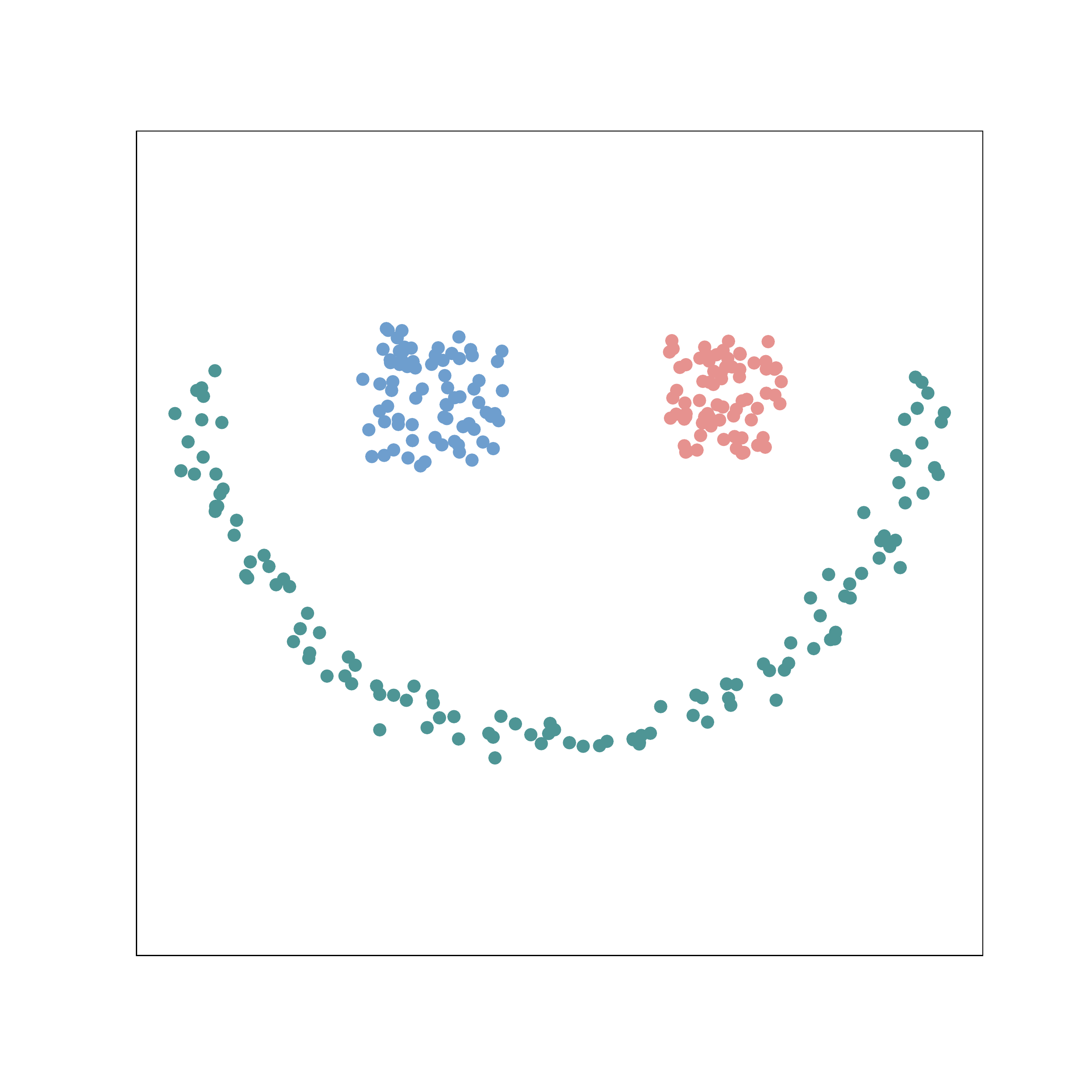}
			\label{fig: zelnik3 ground truth}
		}
		\hskip -30pt
		\subfigure[$k$-means++]{
			\includegraphics[width=4.5cm]{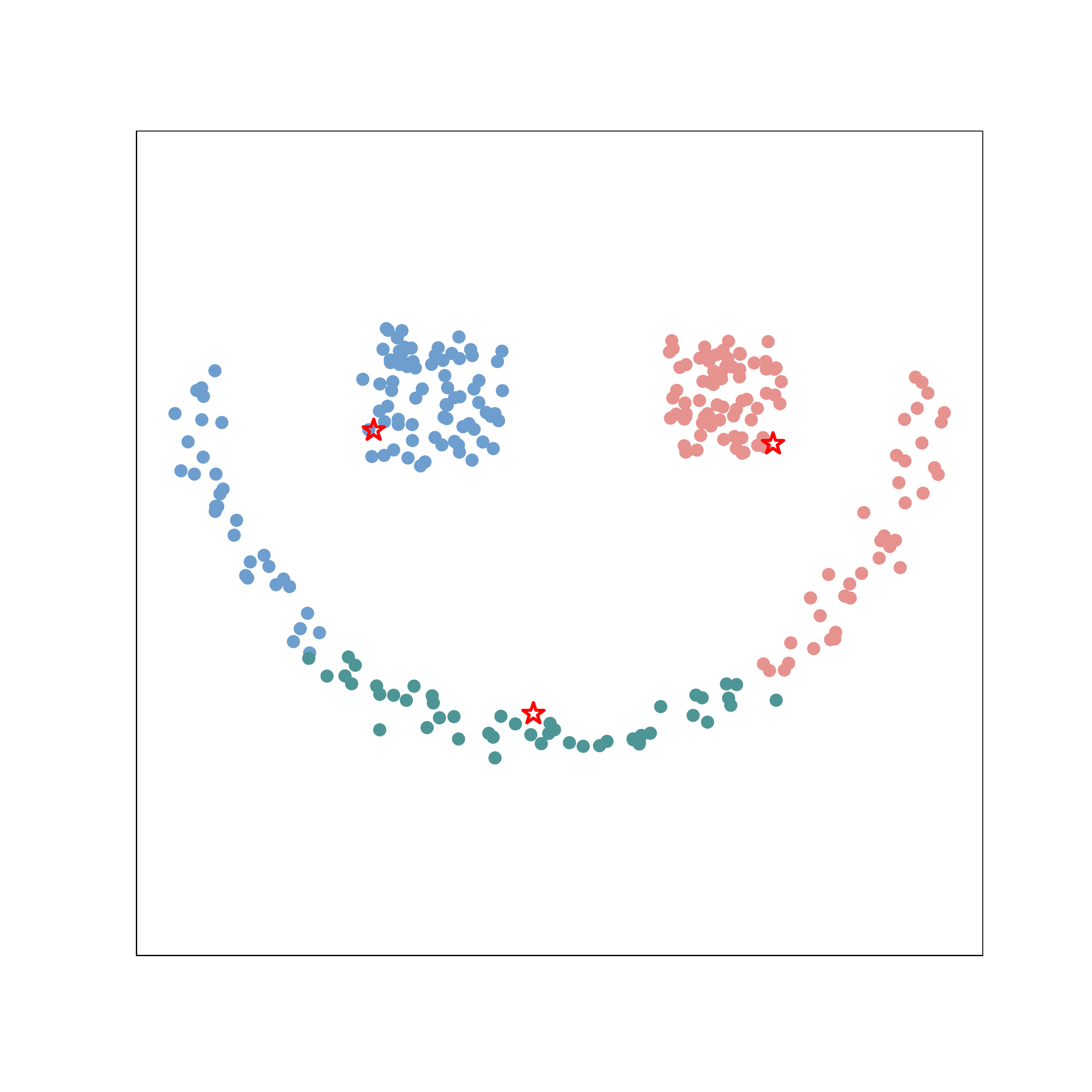}
			\label{fig: zelnik3 k_means}
		}
		\hskip -30pt
		\subfigure[SC]{
			\includegraphics[width=4.5cm]{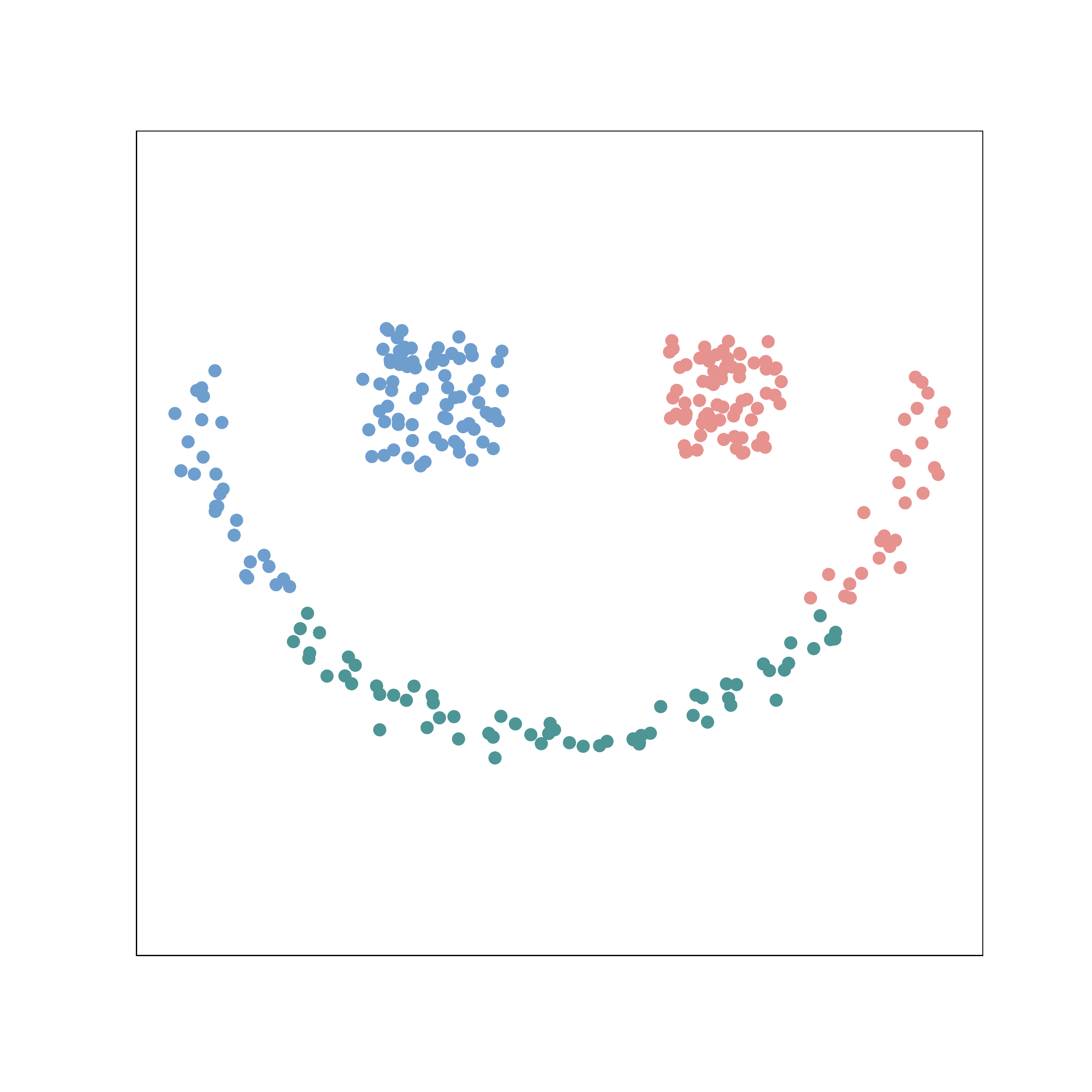}
			\label{fig: zelnik3 SC}
		}
		\hskip -30pt  
		\subfigure[DBSCAN]{
			\includegraphics[width=4.5cm]{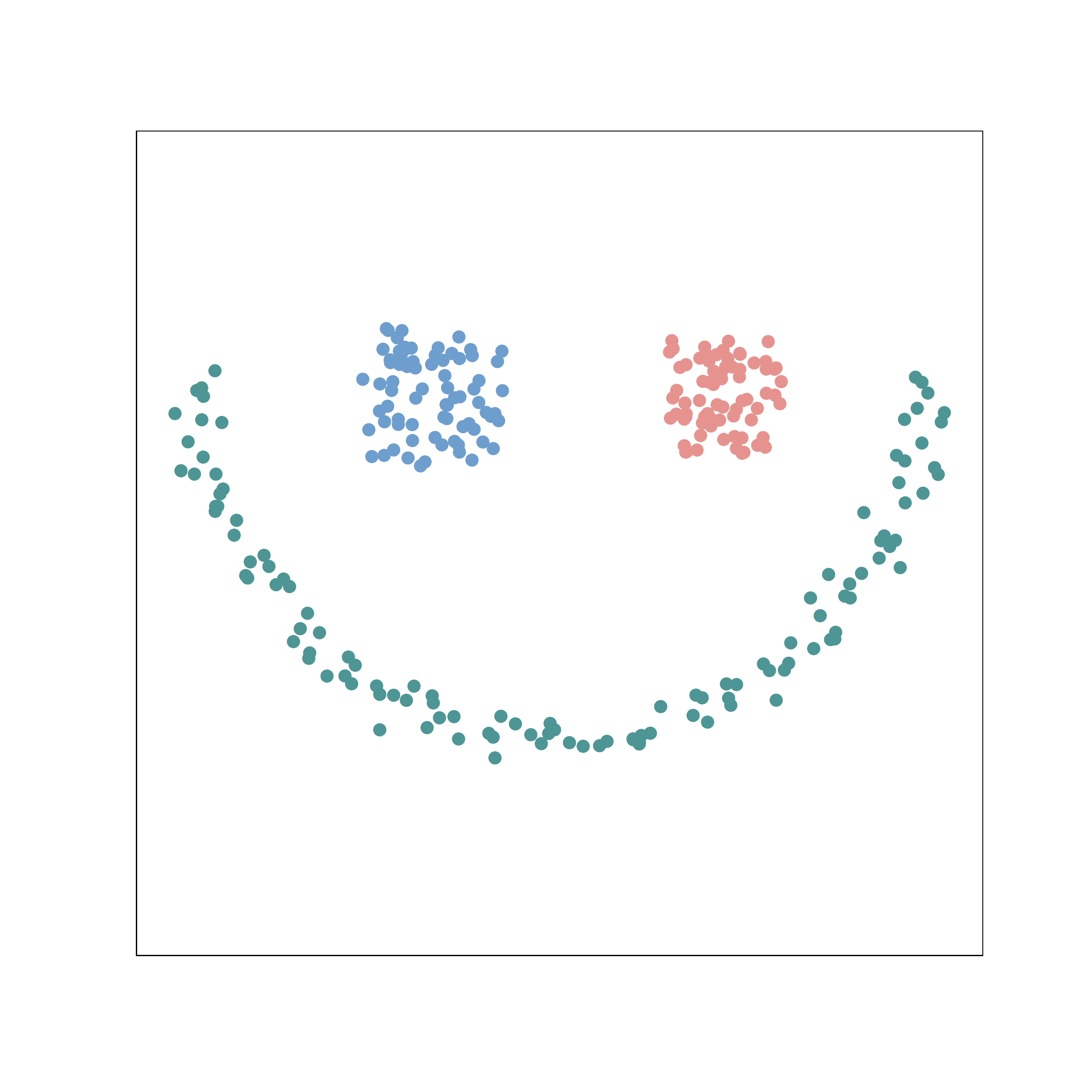}
			\label{fig: zelnik3 DBSCAN}
		}
		\vskip -20pt
		\subfigure[DPC]{
			\includegraphics[width=4.5cm]{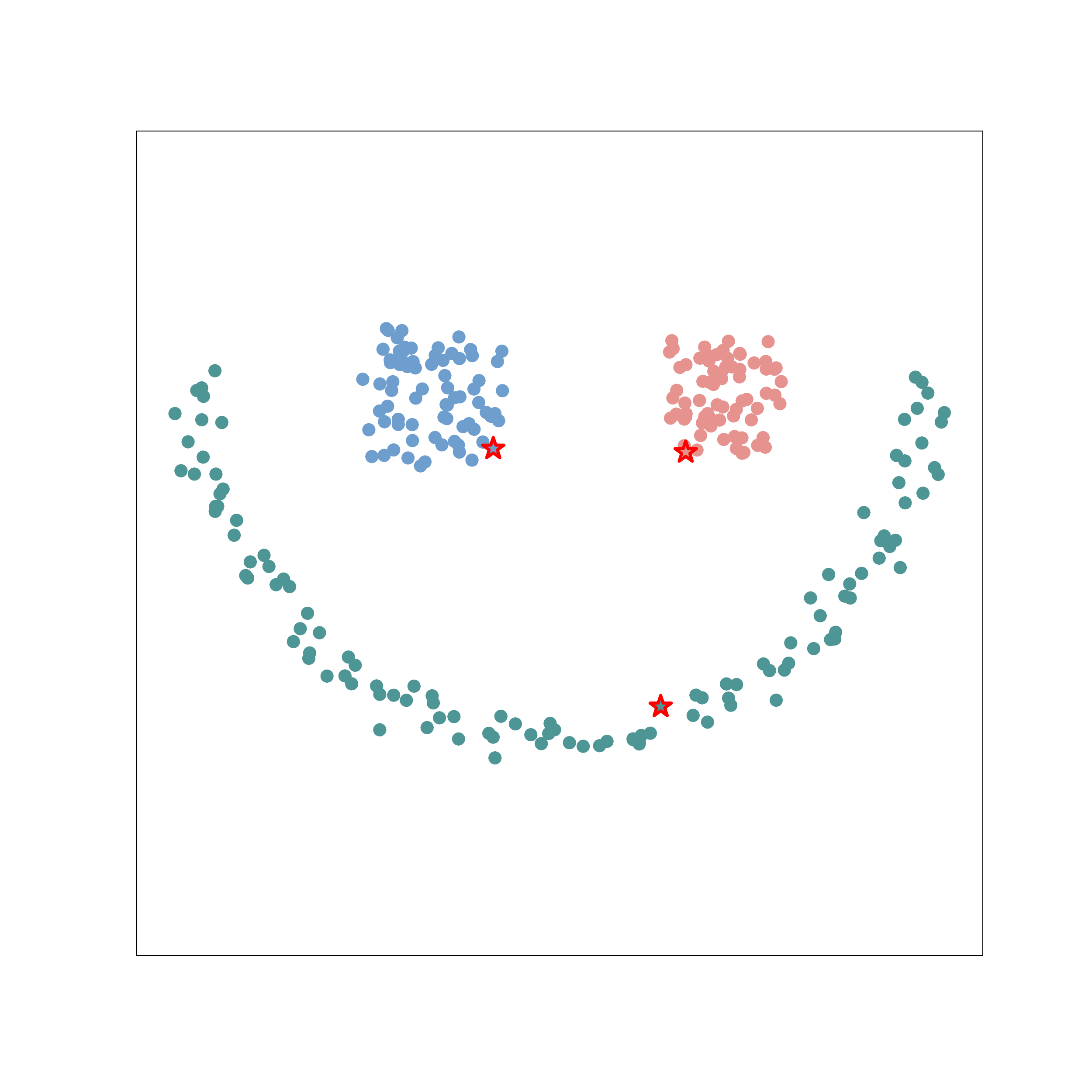}
			\label{fig: zelnik3 DPC}
		}
		\hskip -30pt
		\subfigure[DPC-KNN]{
			\includegraphics[width=4.5cm]{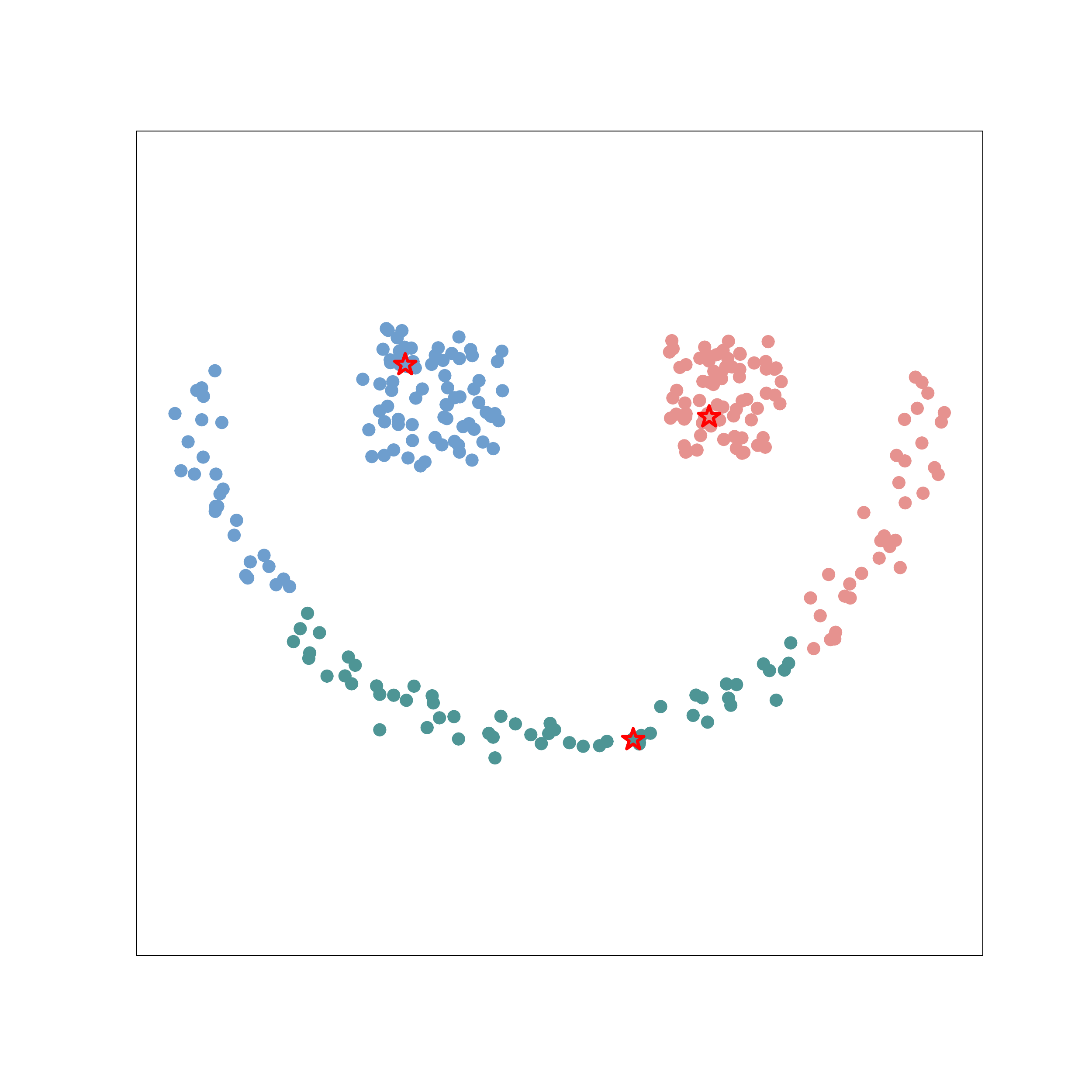}
			\label{fig: zelnik3 DPCKNN}
		}
		\hskip -30pt
		\subfigure[ECM]{
			\includegraphics[width=4.5cm]{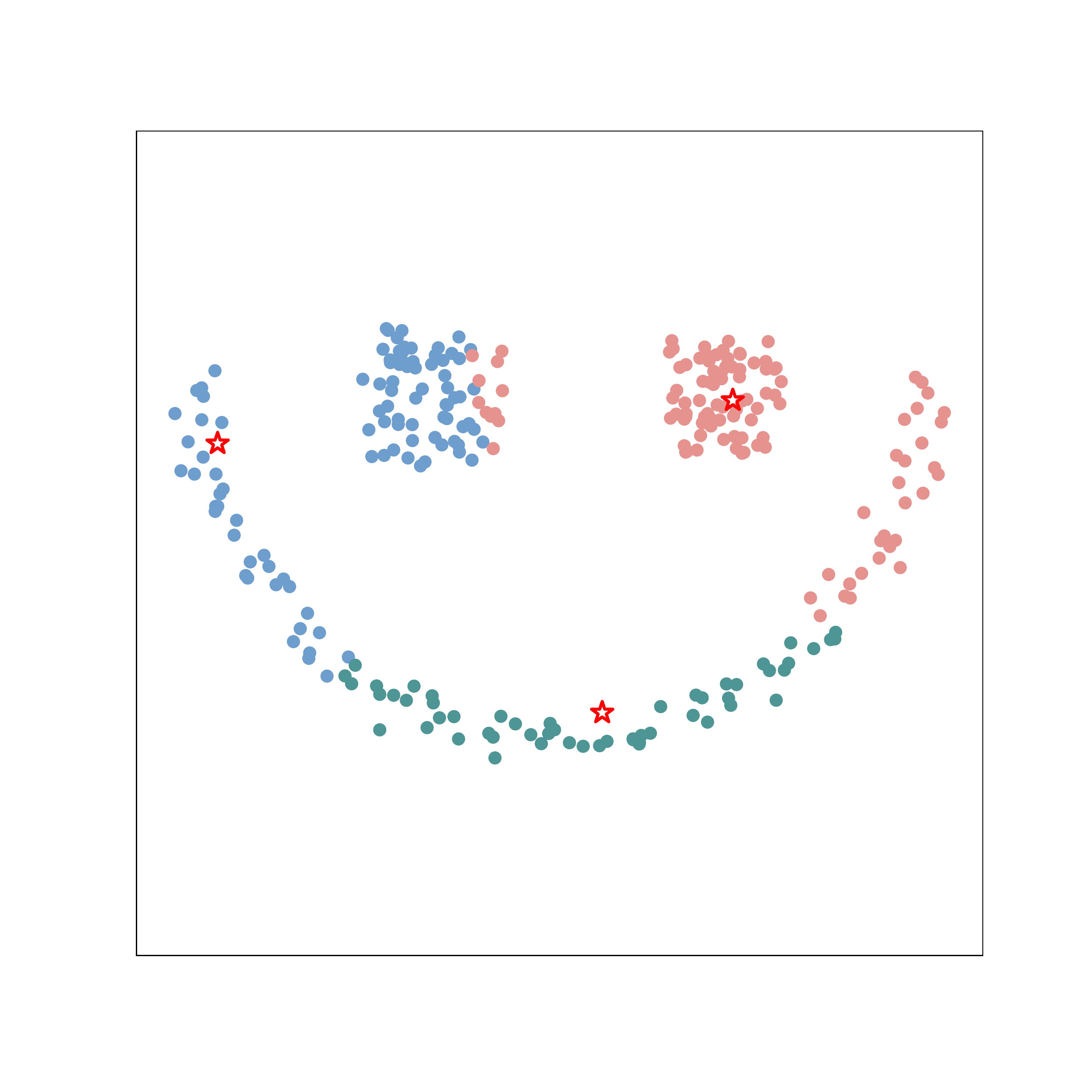}
			\label{fig: zelnik3 ECM}
		}
		\hskip -30pt
		\subfigure[GFDC]{
			\includegraphics[width=4.5cm]{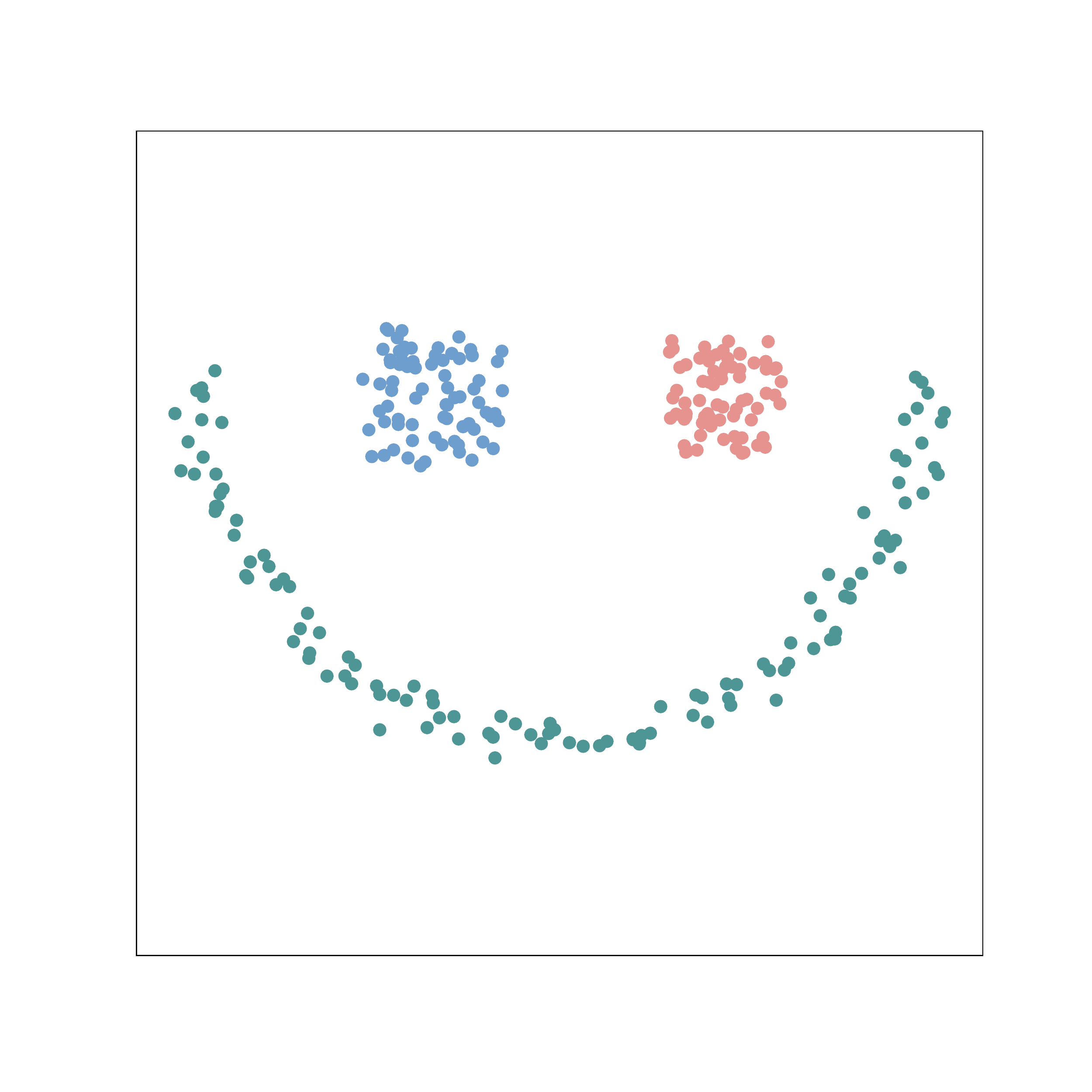}
			\label{fig: zelnik3 proposed}
		}
		\caption{Zelnik3.}
		\label{fig: zelnik3}
\end{figure}

The characteristics of all synthetic datasets used in experiments, including the number of samples, dimensions and clusters, are shown in \autoref{table: Synthetic datasets}. For the six compared clustering algorithms, the optimal parameters of each synthetic dataset are recorded in \autoref{table: parameters on synthetic datasets}, where the parameters of the number of clusters that are necessary for some compared algorithms are set to the true number of clusters. Furthermore, the number of clusters and outliers detected by DBSCAN for each dataset is also shown in \autoref{table: parameters on synthetic datasets}. The visual clustering results of seven algorithms on some datasets are shown in \autoref{fig: 2spiral} to \autoref{fig: zelnik3}, where subfigures (a) to subfigures (h) represent the ground truth, the clustering results of $k$-means++, SC, DBSCAN, DPC, DPC-KNN, ECM and GFDC respectively. It should be noted that the red star markers represent the centers of the clusters detected by $k$-means++, DPC, DPC-KNN and ECM, and the black cross markers represent the noises identified by DBSCAN. Besides, in the experiments of synthetic datasets, there are no thresholds set for identifying outliers in GFDC, because none of the datasets in the experiments contains outliers. Moreover, the evaluation of clustering results, including purity, ARI, AMI and FMI, of the six compared algorithms and the proposed algorithm on twenty datasets are listed in \autoref{table: purity comparison on synthetic datasets}, \autoref{table: ARI comparison on synthetic datasets}, \autoref{table: AMI comparison on synthetic datasets} and \autoref{table: FMI comparison on synthetic datasets}. It is important to note that in these tables, the bolded values in each row represent the best performance, and the symbol "-" indicates that the corresponding algorithm cannot produce valid clustering results. In each table, the values of $k$-means++, SC and ECM are the average results after ten experiments.

\begin{table}[H]\footnotesize
	\begin{center}
		\begin{minipage}[H]{0.90\linewidth}  
			\caption{Synthetic datasets.}
			\label{table: Synthetic datasets}
			\tabcolsep0.1in
			\begin{threeparttable}
				\begin{tabular}{cccc|cccc}
					\toprule  
					Dataset & \#sample & \#dimension & \#cluster & Dataset & \#sample & \#dimension & \#cluster\\
					\cmidrule(lr){1-4}  \cmidrule(lr){5-8}
					2d-10c	&	2990	&	2	&	9	&	DS-577	&	577	&	2	&	3	\\
					2spiral	&	1000	&	2	&	2	&	DS-850	&	850	&	2	&	5	\\
					3MC		&	400		&	2	&	3	&	Jain	&	373	&	2	&	2	\\
					Aggregation	&	788	&	2	&	7	&	Lsun	&	400	&	2	&	3	\\
					Banana-Ori	&	4811	&	2	&	2	&	Smile1	&	1000	&	2	&	4	\\
					Cassini	&	1000	&	2	&	3	&	Smile2	&	1000	&	2	&	4	\\
					Cure-t0-2000n-2D	&	2000	&	2	&	3	&	Smile3	&	1000	&	2	&	4	\\
					Dartboard1	&	1000	&	2	&	4	&	Triangle2	&	1000	&	2	&	4	\\
					Donut2	&	1000	&	2	&	2	&	Zelnik3	&	266	&	2	&	3	\\
					Donut3	&	999		&	2	&	3	&	Zelnik5	&	512	&	2	&	4	\\
					\bottomrule  
				\end{tabular}
			\end{threeparttable}
		\end{minipage}
	\end{center}
\end{table}
\vspace{-0.8cm}

\begin{table}[H]\footnotesize
	\caption{Parameters of clustering algorithms on synthetic datasets.}
	\label{table: parameters on synthetic datasets}
	\tabcolsep0.14in
	\begin{tabular}{ccccccccc}
		\toprule  
		\multirow{2}{*}{Dataset} & SC    & \multicolumn{4}{c}{DBSCAN}    & DPC		& DPC-KNN		& ECM\\
		\cmidrule(lr){2-2} \cmidrule(lr){3-6} \cmidrule(lr){7-7} \cmidrule(lr){8-8} \cmidrule(lr){9-9}
		& $gamma$ & $eps$    & $minpts$ & \begin{tabular}[c]{@{}c@{}}\#detected\\ cluster\end{tabular} & \begin{tabular}[c]{@{}c@{}}\#detected\\ noise\end{tabular} & $p$	& $p$	& $\alpha$ \\
		\midrule  
		2d-10c                   & 0.1   & 6.6684 & 5      & 8                 & 0               & 0.62 	& 0.001 & 1.5\\
		2spiral                  & 3.0   & 0.6116 & 5      & 2                 & 0               & 0.31 	& 0.002 & 3.0\\
		3MC                      & 0.5   & 1.2867 & 5      & 3                 & 0               & 0.20 	& 0.020	& 3.0\\
		Aggregation              & 0.5   & 1.8089 & 10     & 6                 & 2               & 0.45 	& 0.010	& 3.0\\
		Banana-Ori               & 5.0   & 0.0494 & 5      & 2                 & 0               & 0.16 	& 0.010	& 2.5\\
		Cassini                  & 5.0   & 0.1955 & 5      & 3                 & 0               & 0.08 	& 0.005	& 3.0\\
		Cure-t0-2000n-2D         & 4.5   & 0.1130 & 5      & 3                 & 0               & 0.56 	& 0.002	& 3.0\\
		Dartboard1               & 0.01  & 0.0476 & 5      & 4                 & 0               & 0.65 	& 0.060	& 1.0\\
		Donut2                   & 0.01  & 0.0100 & 5      & 0                 & 1000            & 0.30 	& 0.002	& 1.0\\
		Donut3                   & 0.01  & 0.0181 & 5      & 3                 & 1               & 0.33 	& 0.060	& 2.5\\
		DS-577                   & 3.5   & 0.2594 & 8      & 3                 & 93              & 0.02 	& 0.005	& 2.0\\
		DS-850                   & 4.5   & 0.3006 & 5      & 5                 & 19              & 0.03 	& 0.005	& 3.0\\
		Jain                     & 0.5   & 3.2486 & 8      & 2                 & 1               & 0.31 	& 0.020	& 1.0\\
		Lsun                     & 4.5   & 0.5134 & 5      & 3                 & 0               & 0.33 	& 0.020	& 3.0\\
		Smile1                   & 4.0   & 0.0562 & 5      & 4                 & 0               & 0.38 	& 0.060	& 3.0\\
		Smile2                   & 0.5   & 0.0582 & 5      & 4                 & 0               & 0.77 	& 0.010	& 2.5\\
		Smile3                   & 5.0   & 0.0653 & 5      & 4                 & 0               & 0.58 	& 0.005	& 1.0\\
		Triangle2                & 0.5   & 1.5225 & 8      & 4                 & 61              & 0.06 	& 0.002	& 3.0\\
		Zelnik3                  & 0.01  & 0.0338 & 5      & 3                 & 0               & 0.67 	& 0.020	& 2.5\\
		Zelnik5                  & 0.1   & 0.0434 & 5      & 4                 & 0               & 0.24 	& 0.020	& 2.0\\
		\bottomrule  
	\end{tabular}
\end{table}
\vspace{-0.5cm}

The analysis of the visual clustering results for some datasets is presented below.

(1) \emph{2spiral}: The 2spiral dataset has two spiral clusters which embrace each other. From the results of seven algorithms in \autoref{fig: 2spiral}, it can be seen that SC, DBSCAN, DPC-KNN and GFDC get correct results. However, except for DPC-KNN, the center-based clustering algorithm, $k$-means++, DPC and ECM get poor results.

(2) \emph{Aggregation}: The Aggregation dataset contains seven clusters of various shapes and there are a few indistinguishable samples at the junction of some clusters. In \autoref{fig: aggregation}, $k$-means++ incorrectly divides a cluster into three parts and detects a center in the blank area, which causes two clusters to be merged into one. DBSCAN misjudges the number of clusters and detects only six clusters. SC, DPC and GFDC misclassify two samples, where SC and GFDC have the same result, while DPC selects centers that are located within the boundaries of the clusters, which is obviously unreasonable. DPC-KNN misclassifies only one sample and its selected centers are more reasonable than DPC.

(3) \emph{Banana-Ori}: Composed of two dense U-shaped clusters, the clustering results of the Banana-Ori dataset from \autoref{fig: banana Ori} can be analyzed and found that except for DBSCAN and GFDC, all other algorithms fail in clustering. There are four algorithms, $k$-means++, SC, DPC-KNN and ECM, which recklessly split the dataset into upper and lower or left and right parts. DPC considers that the upper U-shaped cluster and part of the lower U-shaped cluster belong to the same cluster. The reason for this erroneous result is that DPC assigns non-central samples by a strategy of finding a sample closest and denser than themselves, and it is obvious that a part of misclassified samples is closer to the center of the blue cluster than the center of the pink cluster, thus leading to such an incorrect result. This result shows the limitation and drawback of the assignment strategy of DPC that it propagates errors rapidly when processing datasets with complex shapes or based on improper centers.

(4) \emph{Donut2}: It can be found from \autoref{fig: donut2} that, only GFDC detects the clusters correctly in the Donut2 dataset, which consists of two clusters including a cyclic cluster and a spherical cluster in the cyclic cluster. Firstly, $k$-means++, SC and ECM divide the dataset into upper and lower parts wrongly. Secondly, DBSCAN identifies all samples in the dataset as noises because no suitable parameters can be found within the given adjustment range. Thirdly, due to the fact that DPC considers the samples with high densities as centers of clusters, so two samples within the spherical cluster are automatically selected as centers, resulting in clustering fails. Lastly, DPC-KNN selects centers on each of the clusters, but since its assignment strategy is not applicable in dealing with cyclic clusters, it also leads to a poor clustering result.

(5) \emph{Donut3}: The Donut3 dataset, shown in \autoref{fig: donut3}, is composed of three clusters with a cyclic cluster and two spherical clusters in the cyclic cluster. Similarly, $k$-means++, SC and ECM incorrectly divide the dataset into three parts. DBSCAN clusters better, except for one sample that is mistakenly considered as noise. Similar to the Donut2 dataset, DPC does not select a center on the cyclic cluster and rigidly divides the cyclic cluster into two semicircles. DPC-KNN does select centers on the right clusters but still fails to get the right clustering result. Overall, only GFDC performs perfectly in the Donut3 dataset.

(6) \emph{Jain}: It can be seen in \autoref{fig: jain} that, the Jain dataset have two bowl-shaped clusters with different density. The clustering results of SC, DPC and GFDC are all correct, but obviously, the centers of clusters selected by DPC seem to be illogical. Additionally, $k$-means++ and ECM directly cut the dataset in half diagonally. DBSCAN detects a cluster but unfortunately misclassifies part of another cluster and wrongly identifies one sample as noise. DPC-KNN cannot even select centers on two clusters separately.

(7) \emph{Smile3}: Like a smiling face, the Smile3 dataset consists of a cyclic cluster, two dense spherical clusters and a bar-shaped cluster. In \autoref{fig: smile3}, it can be seen that DBSCAN and GFDC get the correct clustering results. Again, $k$-means++, SC and ECM divide the dataset into flour pieces in a cake-cutting way. DPC successfully selects four samples located on four different clusters as centers, but the sample selected as the center on the cyclic cluster forms a cluster alone. DPC-KNN is similar to DPC and cannot obtain the correct clustering result.

(8) \emph{Zelnik3}: Similar to the Smile3 dataset, the Zelnik3 dataset is also like a smiling face and it is composed of two square clusters and a curved cluster. As shown in \autoref{fig: zelnik3}, DBSCAN, DPC and GFDC perform well but $k$-means++, SC, DPC-KNN and ECM perform poorly. In addition, DPC once again selects centers at the boundaries of clusters, which violates the premise assumptions of the algorithm.

\begin{table}[H]\footnotesize
	\setlength{\abovecaptionskip}{0cm} 
	\setlength{\belowcaptionskip}{0cm}
	\caption{Comparison of purity for clustering algorithms on synthetic datasets.}
	\label{table: purity comparison on synthetic datasets}
	\tabcolsep0.16in
	\begin{tabular}{ccccccccc}
		\toprule  
		\multirow{2}{*}{Dataset}          & \multicolumn{7}{c}{Algorithm}              \\ \cline{2-8}
							& $k$-means++ & SC 				& DBSCAN 			& DPC 				& DPC-KNN 			& ECM				& GFDC \\ 
		\hline
		2d-10c 				& 0.9983 	& 0.8520 			& 0.8870 			& 0.9207 			& \textbf{1.0000}	& 0.8877			& 0.9997 \\
		2spiral 			& 0.5933 	& \bfseries1.0000 	& \bfseries1.0000 	& 0.8400 			& \bfseries1.0000	& 0.5910			& \bfseries1.0000 \\
		3MC 				& 0.9325 	& \bfseries1.0000 	& \bfseries1.0000 	& 0.9700 			& 0.8500			& 0.9328			& \bfseries1.0000 \\
		Aggregation 		& 0.9388 	& 0.9975 			& 0.9530 			& 0.9975 			& \textbf{0.9987}	& 0.8608			& 0.9975 \\
		Banana-Ori 			& 0.8285 	& 0.8465 			& \bfseries1.0000 	& 0.8782 			& 0.6205			& 0.8381			& \bfseries1.0000 \\
		Cassini 			& 0.8000 	& 0.8394 			& \bfseries1.0000 	& \bfseries1.0000 	& 0.9250			& 0.8000			& \bfseries1.0000 \\
		Cure-t0-2000n-2D 	& 0.8455 	& 0.9000 			& \bfseries1.0000 	& \bfseries1.0000 	& 0.9000			& 0.8489			& \bfseries1.0000 \\
		Dartboard1 			& 0.2500 	& 0.3504 			& \bfseries1.0000 	& 0.2550			& 0.3590			& 0.3025			& \bfseries1.0000 \\
		Donut2 				& 0.5391 	& 0.5380 			& - 				& 0.7150 			& 0.6950			& 0.5219			& \bfseries0.9960 \\
		Donut3 				& 0.7578 	& 0.7708 			& \bfseries1.0000	& 0.8338 			& 0.7938			& 0.6781			& \bfseries1.0000 \\
		DS-577 				& 0.9671 	& 0.9965 			& 0.9047 			& \textbf{0.9983} 	& \textbf{0.9983}	& 0.9653			& \bfseries0.9983 \\
		DS-850 				& 0.9588 	& 0.9965 			& 0.9871 			& 0.9988 			& \bfseries1.0000	& 0.9292			& 0.9976 \\
		Jain 				& 0.7853 	& \bfseries1.0000 	& 0.8123			& \bfseries1.0000 	& 0.9223			& 0.7399			& \bfseries1.0000 \\
		Lsun 				& 0.7660 	& 0.9475 			& \bfseries1.0000 	& \bfseries1.0000 	& 0.9375			& 0.6875			& \bfseries1.0000 \\
		Smile1 				& 0.7500 	& 0.7500 			& \bfseries1.0000 	& 0.7750 			& 0.8080			& 0.7433			& \bfseries1.0000 \\
		Smile2				& 0.7000 	& 0.7000 			& \bfseries1.0000 	& 0.7970 			& 0.7410			& 0.6730			& \bfseries1.0000 \\
		Smile3 				& 0.5950 	& 0.6210 			& \bfseries1.0000 	& 0.7000 			& 0.6260			& 0.6220			& \bfseries1.0000 \\
		Triangle2 			& 0.9580 	& \bfseries0.9970 	& 0.9730 			& 0.9960 			& 0.9960			& 0.9395			& \bfseries0.9970 \\
		Zelnik3 			& 0.7406 	& 0.7932 			& \bfseries1.0000 	& \bfseries1.0000 	& 0.7744			& 0.7462			& \bfseries1.0000 \\
		Zelnik5 			& 0.7197 	& 0.7496 			& \bfseries1.0000 	& \bfseries1.0000 	& 0.7383			& 0.7045			& \bfseries1.0000 \\
		\bottomrule  
	\end{tabular}
\end{table}

\begin{table}[H]\footnotesize
	\setlength{\abovecaptionskip}{0cm} 
	\setlength{\belowcaptionskip}{0cm}
	\caption{Comparison of ARI for clustering algorithms on synthetic datasets.}
	\label{table: ARI comparison on synthetic datasets}
	\tabcolsep0.16in
		\begin{tabular}{ccccccccc}
			\toprule  
			\multirow{2}{*}{Dataset}          & \multicolumn{7}{c}{Algorithm}              \\ \cline{2-8}
								& $k$-means++ & SC 				& DBSCAN 			& DPC 				& DPC-KNN 			& ECM				& GFDC \\ 
								\hline
			2d-10c 				& 0.9967 	& 0.7590 			& 0.8649 			& 0.8803 			& \textbf{1.0000}	& 0.7958			& 0.9992 \\
			2spiral 			& 0.0339 	& \bfseries1.0000 	& \bfseries1.0000 	& 0.4619 			& \bfseries1.0000	& 0.0322			& \bfseries1.0000 \\
			3MC 				& 0.8003 	& \bfseries1.0000 	& \bfseries1.0000 	& 0.9071 			& 0.6338			& 0.8026			& \bfseries1.0000 \\
			Aggregation 		& 0.7615 	& 0.9949 			& 0.9065 			& 0.9942 			& \textbf{0.9978}	& 0.6312			& 0.9949 \\
			Banana-Ori 			& 0.4316 	& 0.4800 			& \bfseries1.0000 	& 0.5720 			& 0.0578			& 0.4570			& \bfseries1.0000 \\
			Cassini 			& 0.5334 	& 0.6908 			& \bfseries1.0000 	& \bfseries1.0000 	& 0.8231			& 0.5208			& \bfseries1.0000 \\
			Cure-t0-2000n-2D 	& 0.2712 	& 0.3757 			& \bfseries1.0000 	& \bfseries1.0000 	& 0.4019			& 0.2599			& \bfseries1.0000 \\
			Dartboard1 			& -0.0030 	& 0.0520 			& \bfseries1.0000 	& -0.0014 			& 0.0353			& 0.0114			& \bfseries1.0000 \\
			Donut2 				& 0.0052 	& 0.0048 			& - 				& 0.1842 			& 0.1516			& 0.0015			& \bfseries0.9840 \\
			Donut3 				& 0.5033 	& 0.5173 			& 0.9985 			& 0.6390 			& 0.5977			& 0.3903			& \bfseries1.0000 \\
			DS-577 				& 0.9032 	& 0.9895 			& 0.7529 			& \textbf{0.9949} 	& \textbf{0.9949}	& 0.8983			& \bfseries0.9949 \\
			DS-850 				& 0.9018 	& 0.9910 			& 0.9657 			& 0.9966 			& \bfseries1.0000	& 0.8487			& 0.9944 \\
			Jain 				& 0.3235 	& \bfseries1.0000 	& 0.2654			& \bfseries1.0000 	& 0.7055			& 0.2172			& \bfseries1.0000 \\
			Lsun 				& 0.4377 	& 0.8458 			& \bfseries1.0000 	& \bfseries1.0000 	& 0.8197			& 0.3519			& \bfseries1.0000 \\
			Smile1 				& 0.5465 	& 0.5580 			& \bfseries1.0000 	& 0.6680 			& 0.6659			& 0.5336			& \bfseries1.0000 \\
			Smile2				& 0.3797 	& 0.3926 			& \bfseries1.0000 	& 0.5978 			& 0.5447			& 0.3576			& \bfseries1.0000 \\
			Smile3 				& 0.2029 	& 0.2142 			& \bfseries1.0000 	& 0.4432 			& 0.2846			& 0.2016			& \bfseries1.0000 \\
			Triangle2 			& 0.8816 	& \bfseries0.9900 	& 0.9042 			& 0.9867 			& 0.9867			& 0.8413			& \bfseries0.9900 \\
			Zelnik3 			& 0.4027 	& 0.4810 			& \bfseries1.0000 	& \bfseries1.0000 	& 0.4510			& 0.4083			& \bfseries1.0000 \\
			Zelnik5 			& 0.5031 	& 0.5143 			& \bfseries1.0000 	& \bfseries1.0000 	& 0.4604			& 0.4498			& \bfseries1.0000 \\
			\bottomrule  
		\end{tabular}
\end{table}

In general, from the illustration of some clustering results in \autoref{fig: 2spiral} to \autoref{fig: zelnik3}, it is evident that GFDC can handle those datasets with clusters of arbitrary shapes well and also can cope well with those datasets where there are large differences in density among clusters. What's more, from the comparison results in \autoref{table: purity comparison on synthetic datasets} to \autoref{table: FMI comparison on synthetic datasets}, it can be said that GFDC outperforms the other six compared algorithms on some of the datasets and achieves as good results as them on some of the datasets. In conclusion, the performance of the proposed algorithm on synthetic datasets is excellent.

\begin{table}[H]\footnotesize
	\setlength{\abovecaptionskip}{0cm} 
	\setlength{\belowcaptionskip}{0cm}
	\caption{Comparison of AMI for clustering algorithms on synthetic datasets.}
	\label{table: AMI comparison on synthetic datasets}
	\tabcolsep0.16in
		\begin{tabular}{ccccccccc}
			\toprule  
			\multirow{2}{*}{Dataset}          & \multicolumn{7}{c}{Algorithm}              \\ \cline{2-8}
								& $k$-means++ 	& SC 			& DBSCAN 			& DPC 				& DPC-KNN			& ECM				& GFDC \\ \hline
			2d-10c 				& 0.9953 	& 0.9044 			& 0.9532 			& 0.9271 			& \bfseries1.0000	& 0.8823			& 0.9989 \\
			2spiral				& 0.0246 	& \bfseries1.0000 	& \bfseries1.0000 	& 0.3652 			& \bfseries1.0000	& 0.0233			& \bfseries1.0000 \\
			3MC 				& 0.8080 	& \bfseries1.0000 	& \bfseries1.0000 	& 0.9031 			& 0.7433			& 0.7973			& \bfseries1.0000 \\
			Aggregation 		& 0.8768 	& 0.9914 			& 0.9427 			& 0.9926 			& \textbf{0.9956}	& 0.7565			& 0.9914 \\
			Banana-Ori 			& 0.3398 	& 0.3923 			& \bfseries1.0000 	& 0.5652 			& 0.0410			& 0.3655			& \bfseries1.0000 \\
			Cassini 			& 0.5454 	& 0.7569 			& \bfseries1.0000 	& \bfseries1.0000 	& 0.7881			& 0.5315			& \bfseries1.0000 \\
			Cure-t0-2000n-2D 	& 0.4268 	& 0.5903 			& \bfseries1.0000 	& \bfseries1.0000 	& 0.5982			& 0.4168			& \bfseries1.0000 \\
			Dartboard1 			& -0.0033 	& 0.1219	 		& \bfseries1.0000 	& 0.0017 			& 0.0765			& 0.0259			& \bfseries1.0000 \\
			Donut2 				& 0.0038 	& 0.0035 			& - 				& 0.1839 			& 0.2586			& 0.0011			& \bfseries0.9663 \\
			Donut3 				& 0.5613 	& 0.5729 			& 0.9969 			& 0.7341 			& 0.7177			& 0.4488			& \bfseries1.0000 \\
			DS-577 				& 0.8726 	& 0.9801 			& 0.7560 			& \bfseries0.9901 	& \textbf{0.9901}	& 0.8682			& \bfseries0.9901 \\
			DS-850 				& 0.8939 	& 0.9873 			& 0.9583 			& 0.9953 			& \bfseries1.0000	& 0.8463			& 0.9920 \\
			Jain 				& 0.3673 	& \bfseries1.0000 	& 0.2458 			& \bfseries1.0000 	& 0.6439			& 0.3078			& \bfseries1.0000 \\
			Lsun 				& 0.5395 	& 0.8508 			& \bfseries1.0000 	& \bfseries1.0000 	& 0.8327			& 0.4379			& \bfseries1.0000 \\
			Smile1 				& 0.6064 	& 0.6122 			& \bfseries1.0000 	& 0.7548 			& 0.7492			& 0.5799			& \bfseries1.0000 \\
			Smile2 				& 0.5125 	& 0.5164 			& \bfseries1.0000 	& 0.6464 			& 0.6042			& 0.4650			& \bfseries1.0000 \\
			Smile3 				& 0.3931 	& 0.4007 			& \bfseries1.0000 	& 0.6341 			& 0.5250			& 0.3770			& \bfseries1.0000 \\
			Triangle2 			& 0.8689 	& \bfseries0.9865 	& 0.8869 			& 0.9811 			& 0.9811			& 0.8325			& \bfseries0.9865 \\
			Zelnik3 			& 0.5289 	& 0.5803 			& \bfseries1.0000 	& \bfseries1.0000 	& 0.5614			& 0.5204			& \bfseries1.0000 \\
			Zelnik5 			& 0.6708 	& 0.6551 			& \bfseries1.0000 	& \bfseries1.0000 	& 0.6562			& 0.5621			& \bfseries1.0000 \\
			\bottomrule  
		\end{tabular}
\end{table}

\begin{table}[H]\footnotesize
	\setlength{\abovecaptionskip}{0cm} 
	\setlength{\belowcaptionskip}{0cm}
	\caption{Comparison of FMI for clustering algorithms on synthetic datasets.}
	\label{table: FMI comparison on synthetic datasets}
	\tabcolsep0.16in
		\begin{tabular}{ccccccccc}
			\toprule  
			\multirow{2}{*}{Dataset}          & \multicolumn{7}{c}{Algorithm}              \\ \cline{2-8}
								& $k$-means++ & SC 			& DBSCAN 			& DPC 				& DPC-KNN			& ECM				& GFDC \\ \hline
			2d-10c 				& 0.9971 & 0.8249 			& 0.8914 			& 0.8974 			& \bfseries1.0000	& 0.8273			& 0.9993 \\
			2spiral 			& 0.5165 & \bfseries1.0000 	& \bfseries1.0000 	& 0.7307 			& \bfseries1.0000	& 0.5156			& \bfseries1.0000 \\
			3MC 				& 0.8682 & \bfseries1.0000 	& \bfseries1.0000 	& 0.9389 			& 0.7600			& 0.8695			& \bfseries1.0000 \\
			Aggregation 		& 0.8152 & 0.9960 			& 0.9298 			& 0.9954 			& \textbf{0.9983}	& 0.7070			& 0.9960 \\
			Banana-Ori 			& 0.7178 & 0.7442 			& \bfseries1.0000 	& 0.7955 			& 0.5339			& 0.7315			& \bfseries1.0000 \\
			Cassini 			& 0.7029 & 0.8001 			& \bfseries1.0000 	& \bfseries1.0000 	& 0.8871			& 0.6964			& \bfseries1.0000 \\
			Cure-t0-2000n-2D 	& 0.6323 & 0.6973 			& \bfseries1.0000 	& \bfseries1.0000 	& 0.7161			& 0.6310			& \bfseries1.0000 \\
			Dartboard1 			& 0.2470 & 0.3035 			& \bfseries1.0000 	& 0.3483 			& 0.2983			& 0.2605			& \bfseries1.0000 \\
			Donut2 				& 0.5045 & 0.5044 			& - 				& 0.6246 			& 0.6491			& 0.5009			& \bfseries0.9920 \\
			Donut3 				& 0.6926 & 0.6989 			& 0.9990 			& 0.7704 			& 0.7516			& 0.6066			& \bfseries1.0000 \\
			DS-577 				& 0.9354 & 0.9930 			& 0.8313 			& \bfseries0.9966 	& \textbf{0.9966}	& 0.9321			& \bfseries0.9966 \\
			DS-850 				& 0.9218 & 0.9929 			& 0.9728 			& 0.9973 			& \bfseries1.0000	& 0.8798			& 0.9955 \\
			Jain 				& 0.7003 & \bfseries1.0000 	& 0.8047 			& \bfseries1.0000 	& 0.8779			& 0.6519			& \bfseries1.0000 \\
			Lsun 				& 0.6435 & 0.9065 			& \bfseries1.0000 	& \bfseries1.0000 	& 0.8915			& 0.5835			& \bfseries1.0000 \\
			Smile1 				& 0.6638 & 0.6722 			& \bfseries1.0000 	& 0.7715 			& 0.7619			& 0.6582			& \bfseries1.0000 \\
			Smile2 				& 0.5531 & 0.5632 			& \bfseries1.0000 	& 0.7284 			& 0.7013			& 0.5450			& \bfseries1.0000 \\
			Smile3 				& 0.4478 & 0.4592 			& \bfseries1.0000 	& 0.6902 			& 0.5447			& 0.4512			& \bfseries1.0000 \\
			Triangle2 			& 0.9164 & \bfseries0.9930 	& 0.9328 			& 0.9907 			& 0.9907			& 0.8876			& \bfseries0.9930 \\
			Zelnik3 			& 0.6158 & 0.6611 			& \bfseries1.0000 	& \bfseries1.0000 	& 0.6432			& 0.6177			& \bfseries1.0000 \\
			Zelnik5 			& 0.6466 & 0.6501 			& \bfseries1.0000 	& \bfseries1.0000 	& 0.6246			& 0.5982			& \bfseries1.0000 \\
			\bottomrule  
		\end{tabular}
\end{table}

\subsection{Experiments on real-world datasets and results analysis\label{subsection: Experiments on real-world datasets}}
The characteristics of all real-world datasets used in experiments, including the number of samples, dimensions and clusters, are shown in \autoref{table: Real-world datasets}. Before experiments, the datasets are preprocessed by first removing the missing values (the number of samples in \autoref{table: Real-world datasets} are the values after removing), and then standardizing each attribute of samples by the removal of the mean and the scaling to unit variance. For the six compared clustering algorithms, the optimal parameters of each real-world dataset are recorded in \autoref{table: parameters on Real-world datasets}, where the parameters of the number of clusters that are necessary for some compared algorithms are set to the true number of clusters. Besides, the number of clusters and outliers detected by DBSCAN for each dataset is also shown in \autoref{table: parameters on Real-world datasets}. Moreover, in the experiments of real-world datasets, there are no thresholds set for identifying outliers in GFDC, because none of the datasets in the experiments contains outliers. The evaluation of clustering results, including purity, ARI, AMI and FMI, of the six compared algorithms and the proposed algorithm on nine datasets are listed in \autoref{table: comparison on Real-world datasets}.

\begin{table}[H]\footnotesize
\setlength{\abovecaptionskip}{0cm} 
\setlength{\belowcaptionskip}{0cm}
\begin{center}
\begin{minipage}[t]{0.68\linewidth}
\caption{Real-world datasets.}
\label{table: Real-world datasets}
\tabcolsep0.21in
\begin{threeparttable} 
	\begin{tabular}{ccccccccc}
		\toprule  
	    Dataset & \#sample & \#dimension & \#cluster\\
        \midrule  
        Breast-cancer-wisconsin & 683 & 9 & 2\\
        EEG Eye State & 14980 & 14 & 2\\
        Ionosphere & 351 & 34 & 2\\
        Seeds & 210 & 7 & 3\\
        Sonar & 208 & 60 & 2\\
        SPECT-heart & 267 & 22 & 2\\
        WDBC & 569 & 30 & 2\\
        Wisc & 699 & 9 & 2\\
        Yeast & 1484 & 8 & 10\\
        \bottomrule  
	\end{tabular}
\end{threeparttable}
\end{minipage}
\end{center}
\end{table}
\vspace{-0.5cm}

\begin{table}[H]\footnotesize
\setlength{\abovecaptionskip}{0cm} 
\setlength{\belowcaptionskip}{0cm}
\caption{Parameters of clustering algorithms on real-world datasets.}
\label{table: parameters on Real-world datasets}
\tabcolsep0.12in
\begin{tabular}{ccccccccc}
\toprule  
\multirow{2}{*}{Dataset} & SC    & \multicolumn{4}{c}{DBSCAN}    & DPC  & DPC-KNN	& ECM\\
\cmidrule(lr){2-2} \cmidrule(lr){3-6} \cmidrule(lr){7-7} \cmidrule(lr){8-8} \cmidrule(lr){9-9}
                         & $gamma$ & $eps$      & $minpts$ & \begin{tabular}[c]{@{}c@{}}\#detected\\ cluster\end{tabular} & \begin{tabular}[c]{@{}c@{}}\#detected\\ noise\end{tabular} & $p$  & $p$ & $\alpha$ \\
\midrule  
Breast-cancer-wisconsin  & 1.00  & 1.5868 & 5      & 2                 & 165             & 0.39 & 0.002 & 3.0\\
EEG Eye State            & 3.50  & 0.8366 & 7      & 2                 & 12              & 0.06 & 0.005 & 1.0\\
Ionosphere               & 0.01  & 2.5203 & 6      & 2                 & 132             & 0.20 & 0.020 & 1.0\\
Seeds                    & 1.50  & 0.7586 & 8      & 3                 & 122             & 0.07 & 0.010 & 3.0\\
Sonar                    & 0.01  & 9.4047 & 5      & 2                 & 7               & 0.84 & 0.005 & 1.5\\
SPECT-heart              & 0.10  & 4.3030 & 5      & 2                 & 51              & 0.90 & 0.001 & 1.0\\
WDBC                     & 0.01  & 2.3456 & 5      & 2                 & 264             & 0.49 & 0.020 & 1.0\\
Wisc                     & 1.00  & 1.9788 & 7      & 2                 & 87              & 0.43 & 0.010 & 3.0\\
Yeast                    & 0.01  & 0.7837 & 5      & 6                 & 687             & 0.11 & 0.010 & 2.5\\
\bottomrule  
\end{tabular}
\end{table}

First of all, \autoref{table: comparison on Real-world datasets} is analyzed from a general perspective. Compared with other compared algorithms, the evaluation index purity of clustering results is considered firstly: GFDC ranks first on the SPECT-heart dataset, and ranks second on the EEG Eye State, Ionosphere, Seeds, Wisc and Yeast datasets. ARI is considered secondly: GFDC ranks first on the Ionosphere, SPECT-heart and Yeast datasets, and ranks second on the Breast-cancer-wisconsin, WDBC and Wisc datasets. AMI is considered next: GFDC ranks first on the WDBC and Yeast datasets, and ranks second on the EEG Eye State, Ionosphere, SPECT-heart and Wisc datasets. FMI is considered lastly, GFDC ranks first on the Ionosphere, Sonar and Yeast datasets, and ranks second on the EEG Eye State, WDBC and Wisc datasets. Furthermore, it is obvious from \autoref{table: comparison on Real-world datasets} that the ranks of GFDC at four evaluation indexes of clustering results for each dataset are stable, almost always ranking in the top three. Although $k$-means++ ranks first on some datasets considering some evaluation indexes, it performs poorly on other datasets and almost always ranks fifth, sixth or seventh. SC, DBSCAN, DPC, DPC-KNN and ECM are less effective than GFDC and they rank mostly fifth, sixth or seventh considering different evaluation indexes. Taking into account the average rank of each algorithm on each evaluation index, GFDC has the best rank, which indicates that the proposed algorithm has the optimal comprehensive performance on these datasets.

\begin{table}[H]\footnotesize
	\setlength{\abovecaptionskip}{0cm} 
	\setlength{\belowcaptionskip}{0cm}
	\caption{Comparison of purity, ARI, AMI and FMI for clustering algorithms on real-world datasets.}
	\label{table: comparison on Real-world datasets}
	\tabcolsep0.07in
	\begin{threeparttable} 
		\begin{tabular}{ccccccccc}
			\toprule  
			\multirow{2}{*}{Dataset}                 & \multirow{2}{*}{} & \multicolumn{7}{c}{Algorithm}              \\ \cline{3-9}
			&                   & $k$-means++ & SC & DBSCAN & DPC & DPC-KNN & ECM & GFDC \\
			\midrule  
			\multirow{4}{*}{Breast-cancer-wisconsin} & purity & \underline{\underline{0.9570}}(2) & 0.9546(4) & \textbf{0.9590}(1) & 0.9502(6) & 0.6501(7) & 0.9517(5) & 0.9546(3) \\
			& ARI & \textbf{0.8334}(1) & 0.8245(3) & 0.7595(6) & 0.8094(5) & -0.0027(7) & 0.8139(4) & \underline{\underline{0.8246}}(2) \\
			& AMI & \textbf{0.7306}(1) & \underline{\underline{0.7256}}(2) & 0.6315(6) & 0.7064(5) & -0.0019(7) & 0.7093(4) & 0.7234(3) \\
			& FMI & \textbf{0.9249}(1) & \underline{\underline{0.9216}}(2) & 0.8847(6) & 0.9121(5) & 0.7348(7) & 0.9166(4) & 0.9214(3) \\ \cline{2-9}
			\multirow{4}{*}{EEG Eye State}		& purity & 0.5513(6) & 0.5533(3) & 0.5512(7) & \textbf{0.5569}(1) & 0.5533(3) & 0.5533(3) & \underline{\underline{0.5537}}(2) \\
			& ARI & 0.0000(6) & 0.0018(4) & -0.0003(7) & \textbf{0.0129}(1) & 0.0018(3) & \underline{\underline{0.0104}}(2) & 0.0012(5) \\
			& AMI & 0.0000(7) & 0.0010(4) & 0.0010(4) & \textbf{0.0099}(1) & 0.0010(4) & 0.0048(3) & \underline{\underline{0.0057}}(2) \\
			& FMI & \textbf{0.7107}(1) & 0.6961(4) & 0.7093(3) & 0.5091(7) & 0.6960(5) & 0.5292(6) & \underline{\underline{0.7094}}(2) \\ \cline{2-9}
			\multirow{4}{*}{Ionosphere}         & purity & 0.7066(5) & 0.6838(6) & \textbf{0.8974}(1) & 0.6752(7) & 0.7293(3) & 0.7094(4) & \underline{\underline{0.8376}}(2) \\
			& ARI & 0.1679(5) & 0.1315(6) & \underline{\underline{0.4727}}(2) & 0.1200(7) & 0.2015(3) & 0.1729(4) & \textbf{0.5185}(1) \\
			& AMI & 0.1231(4) & 0.0881(6) & \textbf{0.4289}(1) & 0.0858(7) & 0.1229(5) & 0.1323(3) & \underline{\underline{0.4173}}(2) \\
			& FMI & 0.6010(5) & 0.5859(6) & \underline{\underline{0.7299}}(2) & 0.5778(7) & 0.6377(3) & 0.6025(4) & \textbf{0.7947}(1) \\ \cline{2-9}
			\multirow{4}{*}{Seeds}              & purity & 0.9190(6) & \textbf{0.9333(1)} & 0.6524(7) &\underline{\underline{ 0.9238}}(2) &0.9190(5) & \underline{\underline{0.9238}}(2) & \underline{\underline{0.9238}}(2) \\
			& ARI & 0.7733(6) & \textbf{0.8109}(1) & 0.1976(7) & \underline{\underline{0.7895}}(2) & 0.7744(5) & 0.7846(4) & 0.7877(3) \\
			& AMI & 0.7255(6) & \textbf{0.7608}(1) & 0.3904(7) & \underline{\underline{0.7486}}(2) & 0.7283(5) & 0.7367(4) & 0.7388(3) \\
			& FMI & 0.8482(6) & \textbf{0.8733}(1) & 0.4908(7) & \underline{\underline{0.8593}}(2) & 0.8489(5) & 0.8557(4) & 0.8579(3) \\ \cline{2-9}
			\multirow{4}{*}{Sonar}              & purity & 0.5337(6) & 0.5481(5) & 0.5337(6) & \textbf{0.6971}(1) & \underline{\underline{0.6587}}(2) & 0.5788(3) & 0.5721(4) \\
			& ARI & -0.0021(6) & 0.0045(5) & -0.0059(7) & \textbf{0.1513}(1) & \underline{\underline{0.0964}}(2) & 0.0209(3) & 0.0163(4) \\
			& AMI & 0.0040(7) & 0.0042(6) & 0.0332(4) & \textbf{0.1099}(1) & \underline{\underline{0.0787}}(2) & 0.0147(5) & 0.0334(3) \\
			& FMI & 0.5583(5) & 0.5028(7) & \underline{\underline{0.6611}}(2) & 0.5784(4) & 0.5899(3) & 0.5099(6) & \textbf{0.6734}(1) \\ \cline{2-9}
			\multirow{4}{*}{SPECT-heart}        & purity & 0.6296(5) & 0.6479(3) & 0.6367(4) & \underline{\underline{0.6479}}(2) & 0.5918(6) & 0.5880(7) & \textbf{0.6592}(1) \\
			& ARI & 0.0618(4) & 0.0777(3) & 0.0610(5) & \underline{\underline{0.0798}}(2) & 0.0017(7) & 0.0225(6) & \textbf{0.0825}(1) \\
			& AMI & 0.0343(5) & 0.0446(4) & \textbf{0.0482}(1) & 0.0451(3) & -0.0040(7) & 0.0173(6) & \underline{\underline{0.0481}}(2) \\
			& FMI & 0.5544(6) & 0.6013(4) & \underline{\underline{0.6058}}(2) & 0.5844(5) & \textbf{0.7066(1)} & 0.5171(7) & 0.6018(3) \\ \cline{2-9}
			\multirow{4}{*}{WDBC}               & purity & \underline{\underline{0.9098}}(2) & 0.8822(4) & 0.7118(7) & 0.8576(5) & 0.8366(6) & \textbf{0.9227}(1) & 0.8875(3) \\
			& ARI & 0.6690(3) & 0.5781(4) & 0.1764(7) & 0.5106(5) & 0.4402(6) & \textbf{0.7130}(1) & \underline{\underline{0.6931}}(2) \\
			& AMI & 0.5518(3) & 0.5094(4) & 0.1363(7) & 0.4323(6) & 0.4379(5) & \underline{\underline{0.5921}}(2)& \textbf{0.6289}(1) \\
			& FMI & 0.8490(3) & 0.8188(4) & 0.5987(7) & 0.7638(6) & 0.7768(5) & \textbf{0.8653}(1) & \underline{\underline{0.8650}}(2) \\ \cline{2-9}
			\multirow{4}{*}{Wisc}               & purity & \textbf{0.9591}(1) & 0.9528(3) & 0.7725(6) & 0.9385(5) & 0.6552(7) & 0.9499(4) & \underline{\underline{0.9542}}(2) \\
			& ARI & \textbf{0.8413}(1) & 0.8177(3) & 0.2559(6) & 0.7660(5) & -0.0027(7) & 0.8073(4)& \underline{\underline{0.8230}}(2) \\
			& AMI & \textbf{0.7381}(1) & 0.7130(3) & 0.2478(6) & 0.6483(5) & -0.0019(7) & 0.6981(4) & \underline{\underline{0.7191}}(2) \\
			& FMI & \textbf{0.9285}(1) & 0.9189(3) & 0.7459(6) & 0.8958(5) & 0.7370(7) & 0.9140(4) & \underline{\underline{0.9211}}(2) \\ \cline{2-9}
			\multirow{4}{*}{Yeast}              & purity & \textbf{0.5451}(1) & \underline{\underline{0.5160}}(2) & 0.3720(6) & 0.4353(5) & 0.4704(4) & 0.3480(7) & 0.4953(3) \\
			& ARI & 0.1635(4) & 0.1755(3) & 0.0333(7) & 0.0895(6) & \underline{\underline{0.1950}}(2) & 0.1101(5) & \textbf{0.2230}(1) \\
			& AMI & \underline{\underline{0.2843}}(2) & 0.2588(3) & 0.0617(7) & 0.1550(5) & 0.2510(4) & 0.1250(6) & \textbf{0.3079}(1) \\
			& FMI & 0.3149(7) & 0.3410(6) & 0.3494(5) & 0.3596(3) & \underline{\underline{0.4027}}(2) & 0.3590(4) & \textbf{0.4725}(1)  \\                                       
			\midrule  
			\makecell[l]{Average rank of purity} &  & 3.7778 & 3.4444 & 5.0000 & 3.7778 & 4.7778 & 4.0000 & 2.4444\\
			\makecell[l]{Average rank of ARI} &  & 4.0000 & 3.5556 & 6.0000 & 3.7778 & 4.6667 & 3.6667 & 2.3333\\
			\makecell[l]{Average rank of AMI} &  & 4.0000 & 3.6667 & 4.7778 & 3.8889 & 5.1111 & 4.1111 & 2.1111\\
			\makecell[l]{Average rank of FMI} &  & 3.8889 & 4.1111 & 4.4444 & 4.8889 & 4.2222 & 4.4444 & 2.0000\\
			\bottomrule  
		\end{tabular}
		\begin{tablenotes}  
			\item[1]{The bolded values in each row represent the best performance, and the underscored values in each row represent the second performance.}
			\item[2]{The values in brackets represent the rank of the value before the brackets in each row. If the value are same in a row, their ranks are shared equally.}
			\item[3]{The purity, ARI, AMI and FMI values of $k$-means++, SC and ECM are the average results after ten experiments.}
		\end{tablenotes}
	\end{threeparttable}
\end{table}

In the following, the performances of seven algorithms on each dataset are analyzed in detail. On the Breast-cancer-wisconsin dataset, $k$-means++ outperforms all others, and GFDC are slightly worse than $k$-means++ without distinction. Although DBSCAN has the highest value at purity, the values of other evaluation indexes are low. Since all noise identified by DBSCAN is considered as one cluster when calculating purity, it is not very meaningful to evaluate the performance of DBSCAN using purity. GFDC performs normally on the EEG Eye State dataset. Observing the algorithms that perform better on the EEG Eye State dataset. DPC outperforms all others at purity, ARI and AMI, but underperforms all others at FMI. And $k$-means++ outperforms all others at FMI, but underperforms all others at AMI and almost underperforms all others except for DBSCAN at purity and ARI. By contrast, GFDC appears to be more stable and ranks second, fifth, second and second at purity, ARI, AMI and FMI. On the Ionosphere dataset, GFDC does best at ARI and FMI, and outperforms the other five compared algorithms at purity and AMI except for DBSCAN. But DBSCAN detects two clusters and 132 noises while the Ionosphere dataset consists of two clusters with 351 samples, so it is obvious that the clustering result of DBSCAN is not good while the result of GFDC is more practical. GFDC surpasses ECM, DPC-KNN, $k$-means++ and DBSCAN on the Seeds dataset, but lags behind SC and DPC by a little. Meanwhile, GFDC, DPC and ECM are tied for second place at purity. On the Sonar dataset, DPC performs well at purity, ARI and AMI, and GFDC performs well at FMI. On the SPECT-heart dataset, GFDC gets the best result than other algorithms at purity and ARI, and a slightly worse result than DBSCAN at AMI, as well as a slightly worse result than DPC-KNN and DBSCAN at FMI. Again, DBSCAN detects two clusters and 51 noises while there are two clusters with 267 samples in the SPECT-heart dataset, so the clustering result of DBSCAN seems less reasonable. GFDC and ECM perform about the same on the WDBC dataset and are far ahead of the other compared algorithms. In addition, GFDC outperforms the other five compared algorithms on the Wisc dataset, and performs slightly worse than $k$-means++. Lastly, on the Yeast dataset, $k$-means++ and SC marginally exceed GFDC at purity, but GFDC exceeds all compared algorithms at other indexes. To sum up the above, the performance of GFDC on real-world datasets is pretty good and the experimental results are most stable.

\section{Conclusions\label{section: conclusion}}

This paper proposes a granule fusion density-based clustering with evidential reasoning, which can handle datasets where the shapes of clusters are arbitrary and there are large differences in density among clusters. The main advantages of the algorithm introduced in this paper include four aspects. First of all, the proposed novel sparse degree metric can measure both local and global densities of samples, which integrates the notion of optimal information granularity and $k$-nearest neighbors. In addition, the granulation process with sparse degrees of samples, which takes full account of different density regions in data, allows the algorithm to handle data with large density differences among clusters. Furthermore, the emerging fusion strategies break the limitation brought by the convex structure of granules, enabling the algorithm to detect clusters with arbitrary shapes. Lastly, the improved evidential assignment method based on an initial clusters structure can reduce the probability and velocity of error propagation and, if needed, identify outliers. From extensive experimental results, the effectiveness and superiority of GFDC are demonstrated on both synthetic and real-world datasets.

In the future, there are some directions for extending the proposed algorithm. The design of more fusion strategies and the extension to dynamic data are interesting research directions.

\section*{Declaration of competing interest}
	We confirm that there are no known conflicts of interest associated with this publication and our financial support will not affect the research outcome.
	
%


\begin{spacing}{1.5}

\end{spacing}

\end{document}